\documentclass{article}

\usepackage{iclr2026_conference,times}

\usepackage[utf8]{inputenc} 
\usepackage[T1]{fontenc}    
\usepackage{hyperref}       
\usepackage{url}            
\usepackage{booktabs}       
\usepackage{amsfonts}       
\usepackage{nicefrac}       
\usepackage{microtype}      
\usepackage{xcolor}         
\usepackage{xspace} 
\usepackage{amsmath} 
\usepackage{graphicx}
\usepackage[table]{xcolor}   
\usepackage{tikz}
\usepackage{tabularx}
\usepackage{pifont}    
\usepackage{makecell}  
\usepackage{colortbl}  
\usepackage{multirow}
\newcommand{\cmark}{\ding{51}}
\newcommand{\xmark}{\ding{55}}

\newcommand{\benchname}{EntityBench\xspace}
\newcommand{\sysname}{EntityMem\xspace}

\title{\benchname: Towards Entity-Consistent Long-Range Multi-Shot Video Generation}

\author{
  Ruozhen He\textsuperscript{1,3}  \quad
  Meng Wei \textsuperscript{1} \quad
  Ziyan Yang\textsuperscript{2} \quad
  Vicente Ordonez\textsuperscript{3} \\
\textsuperscript{1}ByteDance \quad
\textsuperscript{2}ByteDance Seed \quad
\textsuperscript{3}Rice University \\
\texttt{\{catherine.he, vicenteor\}@rice.edu} \\
\texttt{\{weimeng.147, ziyan.yang\}@bytedance.com} \\
}
\iclrfinalcopy
\begin{document}

\maketitle

\begin{abstract}
Multi-shot video generation extends single-shot generation to coherent visual narratives, yet maintaining consistent characters, objects, and locations across shots remains a challenge over long sequences. 
Existing evaluations typically use independently generated prompt sets with limited entity coverage and simple consistency metrics, making standardized comparison across methods difficult.
We introduce \benchname, a benchmark consisting of 140 episodes (2{,}491 shots) derived from real narrative media, with explicit per-shot entity schedules tracking characters, objects, and locations simultaneously across easy, medium, and hard difficulty tiers of up to 50 shots, 13 cross-shot characters, 8 cross-shot locations, 22 cross-shot objects, and recurrence gaps spanning up to 48 shots.
\benchname pairs the dataset with a three-pillar evaluation framework that disentangles intra-shot visual quality, prompt-following alignment, and cross-shot entity consistency. Cross-shot consistency, the central pillar, evaluates each recurring entity through both embedding similarity and LLM per-criterion judgment across entity-type-specific dimensions, with a fidelity gate that admits accurate entity appearance.
To establish baselines, we propose \sysname, a memory-augmented generation system that plans and stores verified per-entity visual references in a persistent memory bank before generation begins, enabling the video backbone to retrieve each entity's appearance across shots. 
Experiments on \benchname show that cross-shot entity consistency degrades sharply with recurrence distance in existing methods, and that explicit per-entity memory yields the highest character fidelity (Cohen's d = +2.33) and presence among methods evaluated.
\end{abstract}

\section{Introduction}
\label{sec:intro}

Recent advances in video generation have enabled high-fidelity single-shot synthesis, and a growing body of work now extends these capabilities to multi-shot video generation, where a sequence of shots forms a coherent visual narrative~\citep{guo2025long,meng2025holocine,luo2026shotstream}. This progression opens new possibilities for automated storytelling, previsualization, and long-form content creation~\citep{wang2025multishotmaster, guo2025long}.

Single-shot models focus on generating visually appealing clips with coherent motion and prompt adherence within a single scene~\citep{gao2025longvie, wan2025wan}.
Multi-shot generation, however, introduces an additional requirement that entities must maintain their visual identity across shots, not just within a single shot.
This \emph{entity consistency} requires awareness of how the same entity was rendered in previous shots, as even small appearance variations may accumulate over long sequences.
The complexity grows further as realistic narratives involve multiple entity types simultaneously. Common entities include characters, objects, and locations, each with different consistency challenges and reappearance patterns across shots.
Current methods address it implicitly through shared attention~\citep{meng2025holocine}, reference conditioning~\citep{zhang2025storymem}, or autoregressive context~\citep{guo2025long}, but how well they actually preserve entity identity over long sequences remains difficult to assess without a standardized evaluation framework.

As shown in Table~\ref{tab:benchmark_comparison}, existing benchmarks focus on single-shot quality, or provide limited multi-shot coverage with few episodes, short shot sequences, restricted entity types, no transition annotations, and narrow evaluation dimensions for intra- and inter-shot quality.
This makes it difficult to systematically diagnose where and why entity consistency breaks down over long sequences.

We introduce \benchname, a benchmark consisting of 140 episodes (2{,}491 shots) derived from real narrative media and enriched through
LLM-based refinement. It spans easy, medium, and hard difficulty tiers with up to 50 shots organized into 1{,}146 scenes with explicit cut and continuation transitions, and recurrence gaps of up to 48 shots per episode.
Each shot is annotated with an explicit entity schedule specifying which characters, objects, and locations should appear.
For comprehensive analysis, we propose a three-pillar evaluation framework comprising 51 metrics: 6 intra-shot quality metrics, 24 prompt-following metrics, and 21 cross-shot consistency metrics combining embedding similarity with LLM judgments. A fidelity gate is used to ensure cross-shot consistency and is measured only on correctly rendered entities.

Using \benchname, we explore entity-level memory management as a path to improving entity fidelity and cross-shot consistency.
We propose \sysname, a memory-augmented generation system that maintains a persistent per-entity memory bank, populated by VLM-based agents that generate, select, and verify entity visual and textual references before video generation begins.
\sysname enables the video backbone to retrieve entity information across shots while reducing error accumulation
that arises from extracting references from generated outputs.

Our contributions are summarized as follows.
\begin{itemize}
\item We propose \benchname, a multi-shot video generation benchmark with explicit per-shot entity schedules, simultaneous multi-entity tracking across characters, objects, and locations.
\item We design a three-pillar evaluation framework that measures intra-shot quality, prompt-following alignment, and cross-shot entity consistency comprehensively.
\item Through \sysname, we show that entity memory management with quality-gated verification can help cross-shot fidelity and consistency.
\end{itemize}

\begin{table*}[t]
\centering
\small
\setlength{\tabcolsep}{4pt}
\caption{Comparison with existing video generation benchmarks.
Char/Obj/Loc: number of annotated characters, objects, and locations. 
Entity Sched: per-shot entity-level schedule annotations. 
Transition: explicit cut/continuation labels.
Intra/Inter-Shot: number of evaluation metrics for within-shot and cross-shot assessment. 
\benchname provides large-scale multi-shot episodes with simultaneous tracking of 3 entity types, per-shot entity schedules, and a comprehensive evaluation suite.}
\label{tab:benchmark_comparison}
\scalebox{0.75}{
\begin{tabular}{l c c c c c c c c c c c c c}
\toprule
\textbf{Benchmark} & \textbf{\makecell{Media-\\Source}} & \textbf{Episodes} & \textbf{\makecell{Max\\shots}} & \textbf{\makecell{Total\\shots}} & \textbf{\makecell{Multi-\\Shot}} & \textbf{\makecell{Entity\\Sched.}} & \textbf{Char} & \textbf{Obj} & \textbf{Loc} & \textbf{Transition} & \textbf{\makecell{Intra-\\Shot}} & \textbf{\makecell{Inter-\\Shot}} \\
\midrule
VBench~\citep{huang2024vbench}                  & \xmark & 946 & - & 946 & \xmark & \xmark & - & 79 & 86 & \xmark & 16 & 0 \\
  OpenS2V-Nexus~\citep{yuan2025opens2v}            & \xmark & 240 & - & 240 & \xmark & \xmark & 3 & 64 & 14 & \xmark & 6 & 0 \\
  LongVGenBench~\citep{gao2025longvie}       & \xmark & 100 & - & 100 & \xmark & \xmark & - & - & - & \xmark & 7 & 0 \\
  MovieBench~\citep{wu2025moviebench}              & \cmark & 6 & 656 & 2{,}875 & \cmark & \xmark & 94 & - & 846 & \xmark & 0 & 1 \\
  VideoMemory~\citep{zhou2026videomemory}          & \cmark & 54 & 12 & 648 & \cmark & \xmark & 54 & 54 & 54 & \xmark & 0 & 3 \\
  MSVBench~\citep{shi2026msvbench}                    & \cmark & 20 & ~14 & 280 & \cmark & \xmark & 72 & - & 104 & \xmark & 10 & 10 \\
  NarrLV~\citep{feng2025narrlv}                    & \xmark & 14 & 6 & 280 & \cmark & \cmark & - & 570 & 347 & \xmark & 0 & 3 \\
  ST-Bench~\citep{zhang2025storymem} & \cmark & 30 & 12 & 300 & \cmark & \cmark & 30 & - & 95 & \cmark & 3 & 3 \\
\midrule
\rowcolor{gray!10}
\textbf{\benchname (ours)} & \textbf{\cmark} & \textbf{140} & \textbf{50} & \textbf{2{,}491} & \textbf{\cmark} & \textbf{\cmark} & \textbf{987} & \textbf{2{,}077} & \textbf{654} & \textbf{\cmark} & \textbf{30} & \textbf{21} \\
\bottomrule
\end{tabular}
}
\vspace{-0.1in}
\end{table*}

\vspace{-0.1in}
\section{Related Work}
\label{sec:related_work}

\noindent{\bf Benchmarks for Video Generation.}
Single-shot video generation quality has been extensively benchmarked.
VBench~\citep{huang2024vbench} established the de facto standard with 16 evaluation dimensions, later extended to I2V and trustworthiness in VBench++~\citep{huang2025vbench++} and to intrinsic faithfulness in VBench-2.0~\citep{zheng2025vbench}.
Other single-shot benchmarks evaluate human-aligned multi-aspect quality~\citep{liu2024evalcrafter,han2025video}, fine-grained text-video alignment~\citep{liu2023fetv}, compositional generation~\citep{sun2025t2v}, and video dynamics~\citep{liao2024evaluation}.
For identity and subject consistency, the IPVG Challenge~\citep{wang2025identity} released VIP-200K with 200K unique identities and OpenS2V-Nexus~\citep{yuan2025opens2v} provides a million-scale subject-to-video benchmark, but both evaluate only single-subject preservation within individual shots.
LongVGenBench~\citep{gao2025longvie} evaluates controllability and consistency for minute-long single-scene videos but does not address multi-shot narratives.

\noindent{\bf Multi-shot Video Generation Evaluation.}
For multi-shot evaluation, MovieBench~\citep{wu2025moviebench} provides a hierarchical movie-level dataset with character banks, shot-level annotations, and evaluation tasks including character ID consistency measured via face recognition.
VideoMemory~\citep{zhou2026videomemory} introduces a 54-case benchmark structured as 3 entity subclasses $\times$ 3 shot lengths $\times$ 6 samples, evaluated at $K \in \{4, 8, 12\}$ shots with only 6 samples per condition.
Each case isolates a single persistent entity type (character, property, or background) while deliberately varying the other two, which is an ablation protocol rather than a realistic narrative setting where multiple entity types must remain consistent simultaneously.
Other multi-shot works~\citep{luo2026shotstream,wang2025multishotmaster,meng2025holocine,wu2025cinetrans} each construct ${\sim}$100 ad-hoc LLM-generated prompts for their own comparisons.
These efforts lack explicit per-shot entity schedules specifying which characters, objects, and locations should appear in each shot, and do not evaluate simultaneous multi-entity consistency at scale over 12 shots per episode.
Our benchmark fills this gap with 140 curated episodes of up to 50 shots spanning easy/medium/hard tiers, explicit entity schedules tracking all entity types per shot, and a dual evaluation framework combining automated metrics with VLM-based holistic judgment for intra-shot quality and inter-shot consistency.

\vspace{0.02in}
\noindent{\bf Multi-Shot Video Generation.}
While text-to-video models~\citep{wan2025wan,yang2024cogvideox,zheng2024open} now produce high-fidelity single-shot clips, real-world narratives demand \emph{multi-shot} sequences with consistent characters and scenes across shot boundaries~\citep{guo2025long,meng2025holocine,luo2026shotstream}.
A comprehensive survey of these methods is beyond the scope of this work but there is already a rich body of work where existing approaches roughly fall into three broad categories: (1) Two-stage keyframe-then-animate methods that first generate consistent keyframes and then animate each with an image-to-video (I2V) model~\citep{zhou2024storydiffusion,huang2024context,meng2025holocine,xiao2025captain,yang2026shotverse,zhang2025storymem,zhou2026videomemory}, 
(2) Holistic multi-shot methods that jointly process all shots in a single denoising pass, learning cross-shot consistency directly from data~\citep{guo2025long,meng2025holocine,wu2025cinetrans,wang2025multishotmaster,qi2025mask,wang2025echoshot,kara2025shotadapter,cai2025mixture,jia2025moga}, 
and (3) Autoregressive multi-shot methods that reformulate the task as sequential next-shot prediction~\citep{luo2026shotstream,yin2025slow,huang2025self,liu2025rolling,yang2025longlive,yesiltepe2025infinity}.
Across all three paradigms, entity consistency emerges implicitly from architectural design rather than being an explicit objective.
Our work addresses this gap with both a benchmark that directly measures per-entity consistency across shots and a multi-agent system with explicit per-entity visual memory management.

\begin{table}[t]
\centering
\caption{Data curation funnel from raw clips to the final benchmark. Each stage shows the input count, output count, and retention rate.}
\label{tab:curation_pipeline}
\small
\begin{tabular}{lccc}
\toprule
\textbf{Stage} & \textbf{Before} & \textbf{After} & \textbf{Retained} \\
\midrule
Quality filtering (clips) & 100{,}000 & 45{,}589 & 46\% \\
Content filtering (episodes) & 831 & 606 & 73\% \\
Window selection (shots) & 55{,}142 & 2{,}491 & 5\% \\
\bottomrule
\end{tabular}
\vspace{-0.25in}
\end{table}

\begin{table}[t]
\centering
\caption{Benchmark statistics across difficulty tiers. Cross-shot counts report entities appearing in 2+ shots. Recurrence gap measures the number of intervening shots between consecutive appearances of the same entity. Memory-test shots contain recurring entities without first-appearance descriptions.}
\label{tab:benchmark_stats}
\small
\begin{tabular}{lcccc}
\toprule
& \textbf{Easy} & \textbf{Medium} & \textbf{Hard} & \textbf{All} \\
\midrule
Episodes & 80 & 40 & 20 & 140 \\
Shots & 873 & 618 & 1{,}000 & 2{,}491 \\
\midrule
Cross-shot characters & 5.1\tiny{$\pm$1.4} & 6.0\tiny{$\pm$1.8} & 8.9\tiny{$\pm$2.1} & 5.9\tiny{$\pm$2.1} \\
Cross-shot locations & 2.4\tiny{$\pm$0.8} & 2.5\tiny{$\pm$0.7} & 4.9\tiny{$\pm$1.7} & 2.8\tiny{$\pm$1.3} \\
Cross-shot objects & 3.9\tiny{$\pm$2.1} & 6.1\tiny{$\pm$2.5} & 13.3\tiny{$\pm$4.5} & 5.9\tiny{$\pm$4.2} \\
\midrule
Mean recurrence gap & 2.1\tiny{$\pm$1.8} & 2.2\tiny{$\pm$2.0} & 3.4\tiny{$\pm$4.8} & 2.7\tiny{$\pm$3.6} \\
Max recurrence gap & 8.0\tiny{$\pm$2.0} & 9.9\tiny{$\pm$2.2} & 33.5\tiny{$\pm$7.8} & 12.2\tiny{$\pm$9.4} \\
\bottomrule
\end{tabular}
\vspace{-0.15in}
\end{table}

\section{\benchname: Cross-Shot Entity Consistency Benchmark}
\label{sec:benchmark}
 
As multi-shot video generation methods advance, there is a need for standardized evaluation of entity consistency across shots.
Existing works typically evaluate on ad-hoc sets of
LLM-generated prompts~\citep{wu2025cinetrans,
wang2025multishotmaster, meng2025holocine}, 
or on small controlled benchmarks that isolate individual entity types~\citep{zhou2026videomemory} rather than evaluating simultaneous multi-entity consistency.
\benchname provides a curated benchmark of 140 episodes totaling approximately 2{,}491 shots across easy, medium, and hard difficulty tiers, with explicit entity schedules that specify which characters, objects, and locations should appear in each shot, and a standardized evaluation framework for intra-shot quality and inter-shot entity consistency.

\subsection{Data Construction}
\label{sec:data_construction}

\vspace{0.02in}
\noindent{\bf Source data.}
Constructing multi-shot video scripts with natural entity dynamics is difficult through LLM prompting alone: scheduling entities across shots with reasonable occurrence patterns, diverse interactions, and coherent scene structures remains an open challenge.
\benchname instead derives its scripts from existing narrative media, filtered by visual clarity, aesthetics, and motion quality.
This provides a foundation of natural character interactions, scene transitions, and entity reappearance patterns that reflect how characters, objects, and locations actually co-occur in real media. 
Yet, the source material serves only as a seed. The final scripts are generated through LLM-based enrichment, allowing for story adaptation and tolerance for deviation from the original narrative.
 
\vspace{0.02in}
\noindent{\bf Entity extraction and linking.}
From the source material, we extract shots and identify recurring entities through a multi-stage annotation pipeline. Characters are first detected per frame using an object detector~\citep{wang2024yolov10} and a face detector with embedding extraction~\citep{deng2019arcface}, then tracked into per-shot tracklets via IoU-based assignment~\citep{zhang2022bytetrack}.
To establish cross-shot character identities, tracklet embeddings (face and body features~\citep{oquab2023dinov2}) are clustered within each episode using hierarchical agglomerative clustering,
with a co-occurrence constraint that rejects merges between tracklets overlapping temporally within the same shot.
However, embedding-based clustering alone produces fragmented identities with limited recurrence, especially when appearances are far apart in the source material.
We address this with an LLM-based~\citep{comanici2025gemini} deduplication stage that consolidates character clusters across distant shots, while preserving the constraint that characters co-occurring in the same shot or sharing adjacent tracklet IDs remain distinct.
Objects and locations are harder to cluster from visual features alone due to the entanglement of foreground and background regions.
We instead use an LLM~\citep{comanici2025gemini} to first propose local registries within temporal chunks, then merge them into episode-level identities, and finally verify their appearance against the episode script and videos for potential contradictions.
 
\vspace{0.02in}
\noindent{\bf Script refinement and enrichment.}
With entity identities established, the raw annotations undergo a multi-stage enrichment pipeline to produce generation-ready video prompts.
Character descriptions are polished to focus on detailed facial features and demographics while removing actions, camera directions, and transient states.
Object descriptions are refined to distinguish visual properties from functional context, and location descriptions are expanded with spatial and atmospheric detail.
Per-shot action text is then enriched to avoid static actions and encourage interactions between characters, guided by the resolved entity schedules, the global story context, and a temporal window of neighboring shots.
 
\vspace{0.02in}
\noindent{\bf Verification.}
We verify that every entity in a shot's entity schedule is actually mentioned in the action text, repairing mismatches through targeted LLM calls that decide whether it is logical to add the missing entity or remove it from the schedule.
We adopt multi-pass refinement and validation in this stage.
A final validation stage detects and repairs contradictions between action text and entity descriptions, as well as physically impossible actions.
\benchname provides a structured story script per episode containing scene boundaries, shot descriptions with entity descriptions, and an explicit entity schedule mapping each shot to its scheduled characters, objects, and locations.

\vspace{0.02in}
\noindent{\bf Benchmark.}
Table~\ref{tab:curation_pipeline} summarizes the curation funnel.
Starting from 100K production clips, quality filtering retains 46\% based on visual clarity and aesthetics.
After entity annotation, content filtering removes 27\% of episodes with high subtitle density or documentary-style content.
Finally, we select the best contiguous shot windows using a sliding-window approach that scores windows by cross-shot entity recurrence, interaction density, and scene transition frequency, retaining 5\% of all shots.
The final benchmark contains 140 episodes totaling 2{,}491 shots across 1{,}146 scenes, all passing comprehensive verification.
As shown in Table~\ref{tab:benchmark_stats}, easy episodes contain 8--12 shots, already matching or exceeding existing benchmarks in scale, while hard episodes average 8.9 cross-shot characters with a maximum recurrence gap of 33.5 shots.
62\% of hard-tier shots are recurrence-only (no first-appearance description), serving as direct tests of entity memory.

\begin{figure}[t]
    \centering
    \includegraphics[width=\linewidth]{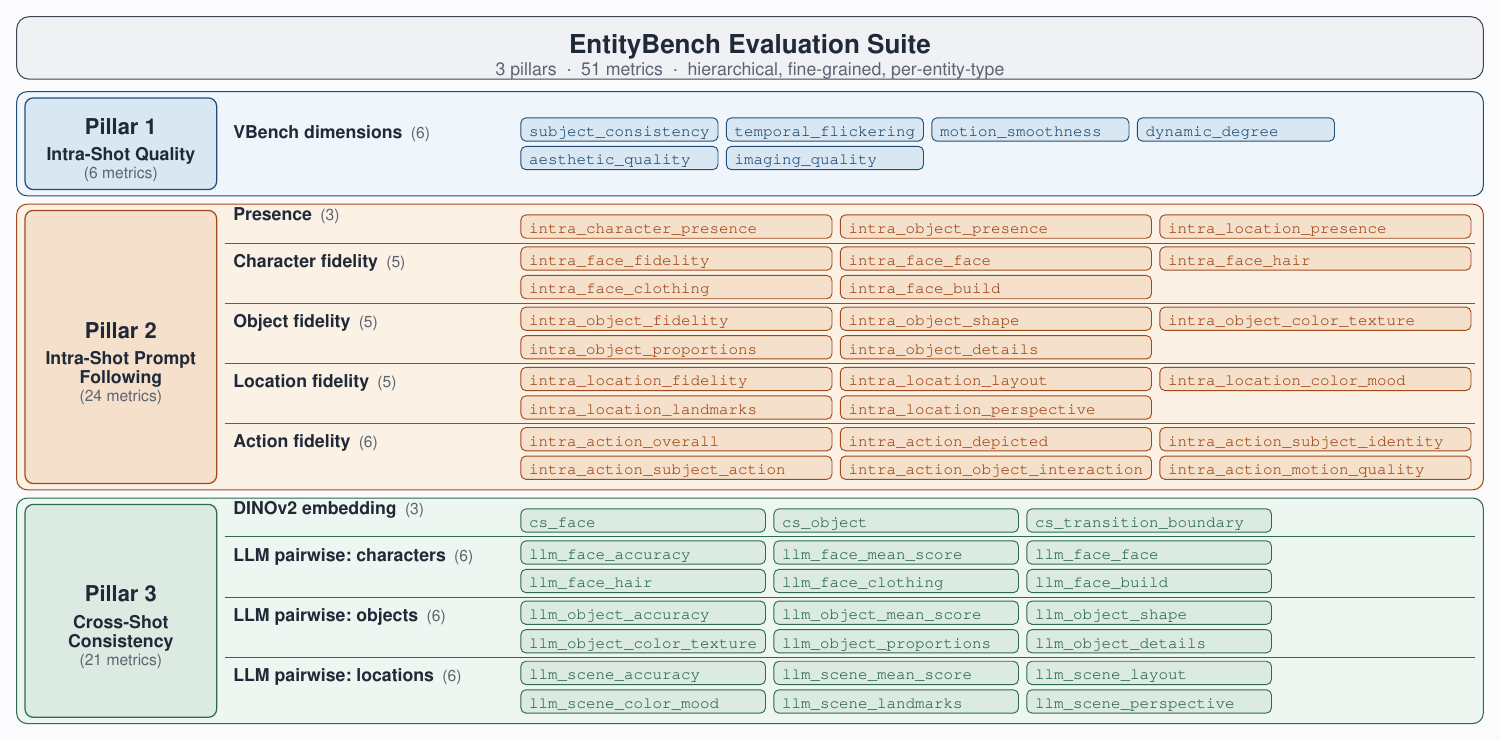}
    \caption{Overview of the \benchname evaluation suite. Three pillars progressively assess whether each shot is well-formed (Pillar~1), whether it follows its prompt (Pillar~2), and whether
entities remain consistent across shots (Pillar~3). Pillar~2's per-entity fidelity scores gate admission into Pillar~3's cross-shot pool. 51 metrics total across 3 pillars.}
    \label{fig:placeholder}
    \vspace{-0.15in}
\end{figure}

\subsection{Evaluation Framework}
\label{sec:evaluation}

\benchname evaluates generated multi-shot videos through three pillars that ask questions progressively : (i) is each shot well-formed in isolation, (ii) does each shot match its prompt, and (iii) do shots agree with one another. Pillars build on each other. For example, Pillar~2's per-shot fidelity scores filter the cross-shot pool used in Pillar~3, and the same canonical entity crops are shared across pillars, so the audit chain is cohesive.

\vspace{0.02in}
\noindent{\bf Pillar 1: Intra-shot quality.}
Inspired by~\citet{huang2024vbench}, we adopt six intra-shot quality dimensions, including subject consistency, temporal flickering, motion smoothness, dynamic degree, aesthetic quality, and imaging quality. The first pillar measures each shot's quality independently.

\vspace{0.02in}
\noindent{\bf Pillar 2: Intra-shot prompt-following alignment.}
For each shot, we evaluate three aspects of prompt-following through a unified grounding pass.
GroundingDINO~\citep{liu2024grounding} localizes each scheduled entity using the entity registry description as the query, yielding per-entity crops with a tri-valued status (present/weak/absent) gated on a CLIP~\citep{radford2021learning} text-image similarity threshold. 
We then measure:
(i)~\textbf{presence}: the fraction of scheduled entities that achieve status \emph{present} in the shot, computed separately for characters, objects, and locations;
(ii)~\textbf{per-entity fidelity}: a multimodal LLM~\citep{comanici2025gemini} scores each canonical crop against its registry description on type-specific criteria. It considers face, hair, clothing, build for characters; shape, color/texture, proportions, details for objects; layout, color mood, landmarks, perspective for locations;
(iii)~\textbf{action fidelity}: a labeled multi-frame grid is constructed by tiling six bounding-box-annotated frames into an image, and the LLM judges whether the prompted action is depicted correctly across six sub-criteria. 

\vspace{0.02in}
\noindent{\bf Pillar 3: Cross-shot consistency.}
For each entity that recurs across multiple shots, we measure whether its visual appearance remains stable. The pillar uses two signals computed on the canonical crops from Pillar~2:
(i)~\textbf{Embedding similarity}~\citep{oquab2023dinov2} to a per-entity centroid, computing cross-shot consistency for characters and objects. A cross-shot transitioning boundary metric measures continuity at the scene-internal cuts. (ii)~\textbf{LLM pairwise judging}: each non-anchor appearance is compared to a centroid-representative anchor on the same type-specific criteria as Pillar~2, for accuracy and per-criterion similarity scores. Locations use full frames with a camera-invariant prompt that explicitly handles different angles and partial views of
the same place. Centroid-anchored similarity is adopted rather than
first-anchor, because the centroid is invariant to shot ordering, and more robust to outliers. 

\paragraph{Cross-shot fidelity gate.}
\label{sec:gate}
A naive cross-shot metric may mistakenly reward methods that produce nearly static yet incorrect renderings. They are similar to one another, so their consistency scores are high despite a lack of entity fidelity.
We prevent this by gating the cross-shot pool on Pillar~2's per-shot fidelity. Only (shot,~entity) pairs with intra-shot fidelity above a threshold are admitted into Pillar~3 cross-shot computation. This ensures cross-shot consistency is measured on appearances where the entity was rendered correctly in the first place.
Following this principle, we report all per-entity metrics as fidelity-gate-corrected means: an instance-weighted mean over all eligible (shot,~entity) instances, with gate-skipped instances counting as zero contributions. This convention jointly captures rendering fidelity (the gate pass-rate) and consistency (the score on passed instances), preventing methods from inflating their scores by failing the gate on harder cases.

\section{\sysname: Entity-Aware Context Management}
\label{sec:method}
Multi-shot video narratives require characters, objects, and locations to maintain consistent visual identities across shots.
Existing multi-shot approaches have shown conditioning each shot on whole-frame keyframes from earlier generations can improve overall consistency~\citep{zhang2025storymem, zhou2026videomemory}.
\sysname explores whether it can further help the entity consistency by maintaining a persistent \emph{entity memory bank} that stores isolated, per-entity visual and textual references rather than whole-frame keyframes.
As a starting point, references are generated and verified before any video generation begins, so that each entity's visual identity is established once and reused consistently throughout the sequence.
At generation time, the video backbone retrieves each entity's appearance independently of the scene in which it previously appeared, disentangling entity identity from scene context. \textit{The full design is provided in Appendix~\ref{app:method}.}

The pipeline operates in three stages, each managed by specialized LLM agents~\citep{comanici2025gemini} that make planning, selection, and verification decisions while delegating deterministic execution to tools, such as text-to-image generator~\citep{flux2024} and segmentation model~\citep{ravi2024sam}.

\paragraph{Stage 1: Entity reference generation.}
\textit{A Classification Agent} first determines which entities require standalone visual references: characters always receive portraits, locations receive panoramic backgrounds, and objects are evaluated individually.
For each entity that requires a reference, a \textit{Portrait Agent} gathers the entity's description and first-appearance context, infers the visual style from the story overview, and writes a generation prompt.
A text-to-image model produces $N$ candidates on a chroma-key background, a segmentation model extracts the foreground of each, and the Portrait Agent selects the best result from a composite grid.
A \textit{Verification Agent} then inspects the selected portrait for incorrect characteristics or segmentation failures. If verification fails, the pipeline retries with an alternative background color to improve segmentation contrast.
For locations, a panoramic image is generated and cropped into angle variants (left, center, right) for camera-aware keyframe composition.
The bank also stores a textual description of each entity at its first appearance for prompt injection in later shots.

\paragraph{Stage 2: Keyframe composition.}
A \textit{Layout Agent} translates each shot's narrative action into one or more keyframe layouts.
Given the action text, entity schedule, and (for continuation shots) the previous shot's layout, it determines character positions, camera angle, and the number of keyframes needed.
When the action changes the spatial arrangement mid-shot, the agent produces multiple keyframes capturing the progression.
For continuation shots, the agent reasons about camera panning direction, shifting retained characters accordingly and selecting the matching location angle variant.
A compositor then places height-normalized portraits at planned positions alongside scheduled objects.

\paragraph{Stage 3: Memory-augmented generation.}
The memory bank for each shot is assembled as an ordered sequence: per-character labeled portraits, followed by keyframe composites.
The video backbone receives this alongside a text prompt that includes entity descriptions and shot actions. For recurring entities, stored descriptions are injected into the prompt automatically.
For continuation shots, the last frame of the previous shot serves as a first-frame input for temporal continuity but is excluded from the memory bank to prevent it from overriding curated entity references.

\section{Experiments}
\label{sec:experiments}
 
\subsection{Experimental Setup}
\label{sec:setup}
 
We evaluate three representative open-sourced SotA methods on \benchname.
For the holistic paradigm, we evaluate HoloCine~\citep{meng2025holocine}, which jointly processes all shots in a single denoising pass with window cross-attention and sparse inter-shot self-attention, and CineTrans~\citep{wu2025cinetrans}, which uses mask-based transition control for cinematic shot boundaries. For the two-stage keyframe-then-animate paradigm, we evaluate StoryMem~\citep{zhang2025storymem}, which introduces a persistent memory module for cross-shot keyframe retrieval. We additionally evaluate \sysname, which extends StoryMem with per-entity memory management without additional training.
To ensure fair comparison, we convert \benchname's structured story scripts into each method's native prompt format (e.g., character names or abstract entity IDs), resolving entity schedule annotations into the input representation each method expects.

All experiments are conducted on two nodes with 8 NVIDIA L20 GPUs.
Given the scale of \benchname (2{,}491 shots), each full benchmark run requires substantial compute.
We report all 51 metrics across the three pillars of the \benchname evaluation framework (\S\ref{sec:evaluation}), with the fidelity gate (\S\ref{sec:gate}) and corresponding fidelity-gate-corrected aggregation applied throughout.

\begin{table}[t]
\centering
\caption{Main results on \benchname.
Reported as fidelity-gate-corrected means (\S\ref{sec:gate}).
\textbf{Bold} marks the best score per row; $^\dagger$ marks wins by EntityMem with Cohen's $d > 0.5$ vs the next-best baseline (Table~\ref{tab:cohens_d}).
Pillars~1 and 2 evaluate within-shot quality and prompt alignment; Pillar~3 evaluates cross-shot consistency.
For VBench (Pillar~1), \texttt{imaging\_quality} is on $[0, 100]$; all other metrics are on $[0, 1]$.
Full 51-metric results in Appendix~\ref{appendix:results:full}.}
\label{tab:main}
\small
\setlength{\tabcolsep}{2pt}
\scalebox{0.8}{
\begin{tabular}{l@{\hskip 6pt}cccc@{\hskip 18pt}l@{\hskip 6pt}cccc}
\toprule
\multicolumn{5}{c}{\textbf{Pillars 1 \& 2: Intra-shot}} & \multicolumn{5}{c}{\textbf{Pillar 3: Cross-shot}} \\
\cmidrule(r{15pt}){1-5} \cmidrule(l){6-10}
Metric & Ours & StoryMem & HoloCine & CineTrans & Metric & Ours & StoryMem & HoloCine & CineTrans \\
\midrule
\multicolumn{5}{l}{\textit{P1: Quality}} & \multicolumn{5}{l}{\textit{P3: DINOv2 similarity}} \\
imaging\_quality   & 66.00          & 56.41 & 49.97 & \textbf{68.57}  &  cs\_face                 & 0.737           & \textbf{0.792} & 0.751 & 0.772 \\
aesthetic\_quality & 0.593          & 0.475 & 0.518 & \textbf{0.596}  &  cs\_object               & 0.798           & \textbf{0.839} & 0.803 & 0.794 \\
motion\_smoothness & 0.988          & 0.849 & 0.964 & \textbf{0.990}  &  cs\_transition\_boundary & \textbf{0.738}  & 0.663          & 0.498 & 0.508 \\
\midrule
\multicolumn{5}{l}{\textit{P2: Presence}} & \multicolumn{5}{l}{\textit{P3: LLM characters}} \\
char\_presence     & \textbf{0.967}$^\dagger$  & 0.849          & 0.882 & 0.796  &  llm\_face\_accuracy   & \textbf{0.406}$^\dagger$  & 0.226 & 0.228 & 0.091 \\
obj\_presence      & 0.888                     & \textbf{0.893} & 0.723 & 0.776  &  llm\_face\_mean\_score & \textbf{0.426}$^\dagger$  & 0.234 & 0.242 & 0.145 \\
loc\_presence      & \textbf{0.687}            & 0.681          & 0.624 & 0.651  &  llm\_face\_face       & \textbf{0.381}$^\dagger$  & 0.216 & 0.223 & 0.145 \\
\midrule
\multicolumn{5}{l}{\textit{P2: Fidelity (overall)}} & \multicolumn{5}{l}{\textit{P3: LLM objects}} \\
face\_fidelity     & \textbf{0.740}$^\dagger$  & 0.452          & 0.349 & 0.327  &  llm\_object\_accuracy   & 0.164 & \textbf{0.203} & 0.088 & 0.092 \\
object\_fidelity   & 0.601                     & \textbf{0.618} & 0.267 & 0.384  &  llm\_object\_mean\_score & 0.202 & \textbf{0.222} & 0.094 & 0.145 \\
location\_fidelity & \textbf{0.555}            & 0.504          & 0.306 & 0.428  &  llm\_object\_shape       & 0.232 & \textbf{0.239} & 0.104 & 0.180 \\
\midrule
\multicolumn{5}{l}{\textit{P2: Action}} & \multicolumn{5}{l}{\textit{P3: LLM scenes}} \\
action\_overall    & \textbf{0.618}$^\dagger$  & 0.547          & 0.569 & 0.273  &  llm\_scene\_accuracy    & 0.309          & \textbf{0.398} & 0.304 & 0.119 \\
action\_subject    & \textbf{0.706}$^\dagger$  & 0.595          & 0.606 & 0.478  &  llm\_scene\_mean\_score  & 0.659          & \textbf{0.671} & 0.616 & 0.432 \\
action\_interaction & \textbf{0.781}$^\dagger$ & 0.712          & 0.616 & 0.346  &  llm\_scene\_layout       & \textbf{0.697} & 0.684          & 0.641 & 0.449 \\
\bottomrule
\end{tabular}
}
\end{table}

\subsection{\benchname Evaluation}
\label{sec:results:main}

Table~\ref{tab:main} reports fidelity-gate-corrected means across representative metrics from each pillar. The full 51-metric breakdown appears in Appendix~\ref{appendix:results:full}, and Cohen's~$d$ effect sizes for the head-to-head against the strongest baseline are reported in Table~\ref{tab:cohens_d}.

\vspace{0.02in}
\noindent{\bf \sysname dominates entity-centric prompt-following (Pillar~2).}
Across all five Pillar~2 sub-categories, \sysname produces the most prompt-aligned characters and scenes. Character-related metrics show the largest gains: \texttt{face\_fidelity} reaches 0.740 vs.\ 0.452 for the next-best baseline (StoryMem), with all four sub-criteria (face, hair, clothing, build) won by margins of 0.18--0.30 (Appendix~\ref{appendix:results:full}). \sysname also achieves the highest character presence (0.967, vs.\ 0.882 for HoloCine), demonstrating that scheduled characters consistently appear in their intended shots. On action correctness, \sysname's overall score (0.618) leads the next baseline by 0.05, with the largest gaps on \texttt{subject\_identity} (+0.11) and \texttt{object\_interaction} (+0.07). The per-entity memory bank not only renders characters correctly, but also keeps them recognizable while they execute the prompted action. Location fidelity follows the same pattern, with \sysname winning all five sub-criteria. The single Pillar~2 sub-category where \sysname does not lead is object fidelity, where StoryMem holds a small margin (0.618 vs.\ 0.601); we discuss this trade-off in \S\ref{app:results:tradeoffs}.

\vspace{0.02in}
\noindent{\bf Cross-shot consistency: identity vs.\ embedding similarity (Pillar~3).}
Pillar~3 reveals a structural disagreement between embedding-based metrics and LLM identity judgment. On DINOv2 cosine similarity, StoryMem leads on \texttt{cs\_face} (0.792 vs.\ 0.737) and \texttt{cs\_object} (0.839 vs.\ 0.798). However, on the LLM-judged identity metrics that ask whether the same character is recognizably the same character across shots, \sysname dominates. \texttt{llm\_face\_accuracy} reaches 0.406 vs.\ 0.226 for StoryMem (a 1.8$\times$ improvement), and \sysname wins all six LLM character cross-shot metrics. This disagreement reflects a different concentration embedding-similarity metrics on consistency, where high embedding similarity may not relate to correct identities preserving similar details. \sysname also wins \texttt{cs\_transition\_boundary} (0.738 vs.\ 0.663), capturing continuity at scene-internal cuts, and ties with StoryMem on the new camera-invariant scene metric (\texttt{llm\_scene\_layout} 0.697 vs.\ 0.684; \texttt{llm\_scene\_perspective} 0.727 vs.\ 0.696, both leading).

\vspace{0.02in}
\noindent{\bf Visual quality vs. entity consistency are distinct.}
On Pillar~1 VBench dimensions, CineTrans wins three of three highlighted dimensions (\texttt{imaging\_quality}, \texttt{aesthetic\_quality}, \texttt{motion\_smoothness}); HoloCine wins \texttt{dynamic\_degree} and \texttt{temporal\_flickering} on the full VBench (Appendix~\ref{appendix:results:full}). Both are holistic multi-shot methods that produce all shots in a single denoising pass, which favors per-frame polish but does not, by itself, specifically enforce entity-level consistency across shots. \sysname is competitive on visual quality (second on imaging quality, second on aesthetic quality), but its contribution lies in a complementary direction that produces the most identifiable and prompt-aligned entities across long multi-shot sequences. The contrast is most visible on \texttt{character\_presence}, where CineTrans drops to 0.796 despite winning the quality dimensions, and on \texttt{face\_fidelity}, where CineTrans renders characters at less than half of \sysname's quality (0.327 vs.\ 0.740).

 \begin{table}[t]
\centering
\caption{Paired effect sizes of \sysname vs.\ StoryMem on \benchname, by metric category. Cohen's $d$ is reported with pooled variance (positive favors \sysname). $n_{\mathrm{paired}}$ is the number of episodes where both methods produced an evaluable score (averaged across metrics in the category). Per-metric values are in Appendix~\ref{appendix:results:cohens_d_full}.}
\label{tab:cohens_d}
\small
\begin{tabular}{lcrr}
\toprule
\textbf{Category} & \textbf{\#\,metrics} & \textbf{Avg.\ $d$} & \textbf{$n_{\mathrm{paired}}$} \\
\midrule
\multicolumn{4}{l}{\textit{Where \sysname helps most: character-centric metrics}} \\
\quad Character fidelity (intra-shot)               & 5 & $\mathbf{+1.71}$ & 139 \\
\quad Character presence                            & 1 & $\mathbf{+1.23}$ & 139 \\
\quad Action overall \& sub-criteria                & 6 & $\mathbf{+0.25}$ & 138 \\
\quad Location fidelity (intra-shot)                & 5 & $\mathbf{+0.17}$ & 140 \\
\quad LLM character (cross-shot)                    & 6 & $-0.07$\,$^*$    & 129 \\
\midrule
\multicolumn{4}{l}{\textit{Where \sysname trails: object-centric and embedding-similarity metrics}} \\
\quad Object presence                               & 1 & $-0.24$           & 138 \\
\quad Object fidelity (intra-shot)                  & 5 & $-0.33$           & 138 \\
\quad DINOv2 cross-shot (face / object / boundary)  & 3 & $-0.50$\,$^\dagger$ & 124 \\
\quad LLM object (cross-shot)                       & 6 & $-0.60$           & 121 \\
\quad LLM scene (cross-shot)                        & 6 & $-0.14$           & 140 \\
\midrule
\multicolumn{4}{l}{\textit{Single-Shot Metrics}} \\
\quad VBench intra-shot quality (Pillar 1)          & 6 & $+0.13$           & 140 \\
\bottomrule
\end{tabular}
\end{table}
 
\subsection{Where \sysname Helps Most}
\label{sec:results:effect_size}

\sysname builds and manages a per-entity memory bank which influences the 
rendering of 
recurring characters. Table~\ref{tab:cohens_d} summarizes paired effect sizes across metric categories.
The largest single effect is intra-shot character fidelity, with the broader character-fidelity category (face, hair, clothing, build) averaging $d = +1.71$. Character presence moves substantially as well ($d = +1.23$). \sysname renders the scheduled character in 96.7\% of shots vs.\ 84.9\% for StoryMem, meaning roughly one of every eight scheduled character appearances is missing from StoryMem outputs. Both effects trace to the same architectural choice: each character is regenerated against its own dedicated memory bank rather than being averaged into a shared per-shot context. When a shot needs to depict a character, the model conditions on a tight, per-entity description that survives across shots without being diluted by other entities or scene-level conditioning.

The picture inverts on objects ($d = -0.33$ intra-shot fidelity, $d = -0.60$ in pairwise cross-shot LLM scoring) and on DINOv2 cross-shot embeddings ($d = -0.50$). The DINOv2 deficit, however, is not a cross-shot consistency loss but an embedding-similarity limitation: on the LLM-judged cross-shot character metrics that evaluate identity rather than embedding distance, the comparison is essentially tied (Appendix~\ref{appendix:results:coverage}). The object regression is real: StoryMem's scene-level prompt expansion appears to retain object identity better when objects are scene-bound props rather than character-attached items. It may cause by the condition and entity incompatibility with the keyframe-finetuned storymem weight. The base model lacks knowledge of integrating objects with the video from independent object conditions.

\subsection{Qualitative Comparison}
\label{sec:results:qualitative}

\begin{figure}[t]
\centering
\includegraphics[width=\textwidth]{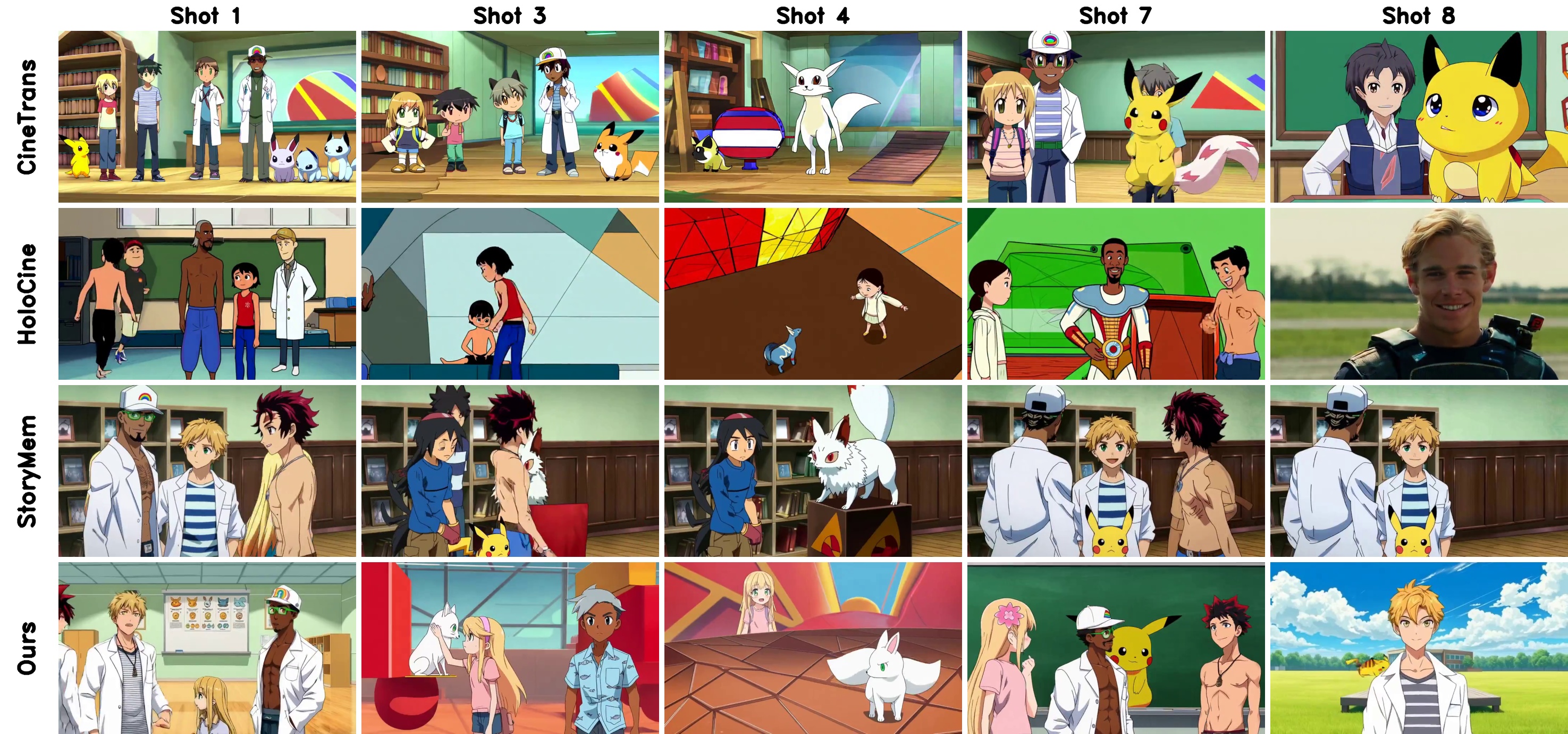}
\caption{
\textbf{Qualitative comparison on a representative episode.}
Multiple characters recur in shots 1, 3, 4, 7, 8.
\sysname preserves all four characters identity, while changing locations according to the prompt.
}
\vspace{-0.1in}
\label{fig:qualitative}
\end{figure}

Figure~\ref{fig:qualitative} grounds the quantitative results in visual evidence. Across the four methods, identity stability and prompt alignment scales directly to per-entity context. The holistic generators (CineTrans, HoloCine) lose character identity gradually despite producing high-quality individual frames, while the persistent-memory baseline (StoryMem) preserves \emph{some} characters but inserts entities not scheduled in the script and fails at generating corresponding locations. \sysname's per-entity memory bank, preserves all four recurring characters and the recurring Pokémon across all eight shots while transitioning to different locations. 

\section{Conclusion}
\label{sec:conclusion}

We introduced \benchname, a comprehensive benchmark for evaluating entity consistency in multi-shot video generation, comprising 140 episodes (2{,}491 shots) derived from real narrative media with explicit per-shot entity schedules across three difficulty tiers. 
The accompanying three-pillar evaluation suite provides 51 metrics spanning intra-shot quality, prompt-following alignment, and cross-shot entity consistency, enabling fine-grained diagnosis of where and why current methods fail to maintain entity identity over
long sequences. 
Using \benchname, we showed that cross-shot consistency degrades with recurrence distance. Through \sysname, per-entity visual and textual memory management system, we show that entity condition for shot generation improves the quality and consistency on 29 dimensions.

\bibliography{custom}
\bibliographystyle{iclr2026_conference}

\appendix

\vspace{0.25in}

In this appendix, we provide benchmark statistics in Section~\ref{appendix:stats},
details on evaluation metrics in Section~\ref{appendix:metrics}, \sysname pipeline details in Section~\ref{app:method},
data examples in Section~\ref{app:entitybench_examples},
prompts used for \sysname in Section~\ref{app:prompts},
supplementary experimental results in Section~\ref{appendix:results},
additional related work in Section~\ref{sec:app_related_work},
and broader impact in Section~\ref{sec:broader_impact}.

\section{Benchmark Statistics}
\label{appendix:stats}

This appendix provides comprehensive descriptive statistics for \benchname, organized around the four properties that distinguish it from prior multi-shot benchmarks. We report (i) episode-level scale and taxonomy (Section~\ref{appendix:stats:scale}); (ii) per-shot multi-entity composition that probes simultaneous tracking of characters, objects, and locations (Section~\ref{appendix:stats:composition}); (iii) long-range structural properties that constitute the cross-shot memory test signal (Section~\ref{appendix:stats:structure}); and (iv) the prompt-level linguistic profile (Section~\ref{appendix:stats:linguistic}). Section~\ref{appendix:stats:tiers} closes with distributions that extends Table~\ref{tab:benchmark_stats} of the main paper. 

\subsection{Scale and Taxonomy}
\label{appendix:stats:scale}

\benchname comprises 140 episodes spanning 1{,}136 scenes and 2{,}491 shots. Across the benchmark, episode registries collectively declare $3{,}718$ unique entities, each described once in the registry block of its episode. Per-shot schedules then reference registry entries by name, yielding $11{,}445$ entity-slot appearances aggregated over all 2{,}491 shots (each appearance is one entity scheduled into one shot). Table~\ref{tab:appendix:stats:scale} reports both quantities with the per-type breakdown.

Characters and locations are scheduled most densely. Each character is referenced by $5.05$ shots on average and each location by $3.72$ shots, while objects skew toward the long tail of single-shot props ($1.94$ references per object). This per-type density gap motivates type-specific evaluation criteria (Sections~\ref{appendix:metrics:pillar2:fid} and~\ref{appendix:metrics:pillar3:llm}): each character contributes roughly $2.6\times$ more to the cross-shot evaluation pool than each object.

\begin{table}[t]
\centering
\caption{Top-level scale and entity statistics for \benchname. \emph{Registry} counts the unique entities declared once per episode in the entity-description block. \emph{Total appearances} is the cumulative number of entity-slots across all per-shot schedules; one entity scheduled into one shot counts as one appearance.}
\label{tab:appendix:stats:scale}
\small
\begin{tabular}{lrr}
\toprule
\textbf{Quantity} & \textbf{Total} & \textbf{Mean / episode} \\
\midrule
Episodes              & 140      & --- \\
Scenes                & 1{,}136  & 8.1\tiny{$\pm$5.3} \\
Shots                 & 2{,}491  & 17.8\tiny{$\pm$13.3} \\
\midrule
\multicolumn{3}{l}{\textit{Entity registry (unique entities)}} \\
\quad Characters             & 987     & 7.05 \\
\quad Locations              & 654     & 4.67 \\
\quad Objects                & 2{,}077 & 14.84 \\
\quad \textbf{Total registry} & \textbf{3{,}718} & \textbf{26.56} \\
\midrule
\multicolumn{3}{l}{\textit{Total scheduled appearances (sum over per-shot schedules)}} \\
\quad Characters             & 4{,}989  & 35.64 \\
\quad Locations              & 2{,}436  & 17.40 \\
\quad Objects                & 4{,}020  & 28.71 \\
\quad \textbf{Total appearances} & \textbf{11{,}445} & \textbf{81.75} \\
\bottomrule
\end{tabular}
\end{table}

\paragraph{Episode size.} Easy and medium episodes range from 10 to 22 shots (median 12, mean 12.4 across these two tiers), drawn from real screenplay structure. The hard tier fixes episode length at 50 shots to provide a controlled stress test of long-range consistency without confounding episode length with content variation. \benchname covers in-distribution and slightly challenging lengths for existing multi-shot video generation models~\citep{meng2025holocine, wu2025cinetrans, zhang2025storymem} plus a fixed-length stress test, measuring both typical-case behavior (easy/medium) and worst-case scaling (hard) within a tractable compute budget.

\paragraph{Scene structure.} Each episode contains a median of 6 scenes; easy and medium episodes span 2 to 13 scenes, while hard-tier episodes extend up to 38 scenes per episode under the 50-shot constraint. The median shots-per-scene ratio is 2.1, reflecting short-form storytelling pacing and ensuring every episode contains multiple scene transitions, which we use to stratify cross-shot evaluation by cut type (Section~\ref{appendix:metrics:pillar3:dino}).

\begin{figure}[t]
\centering
\includegraphics[width=\linewidth]{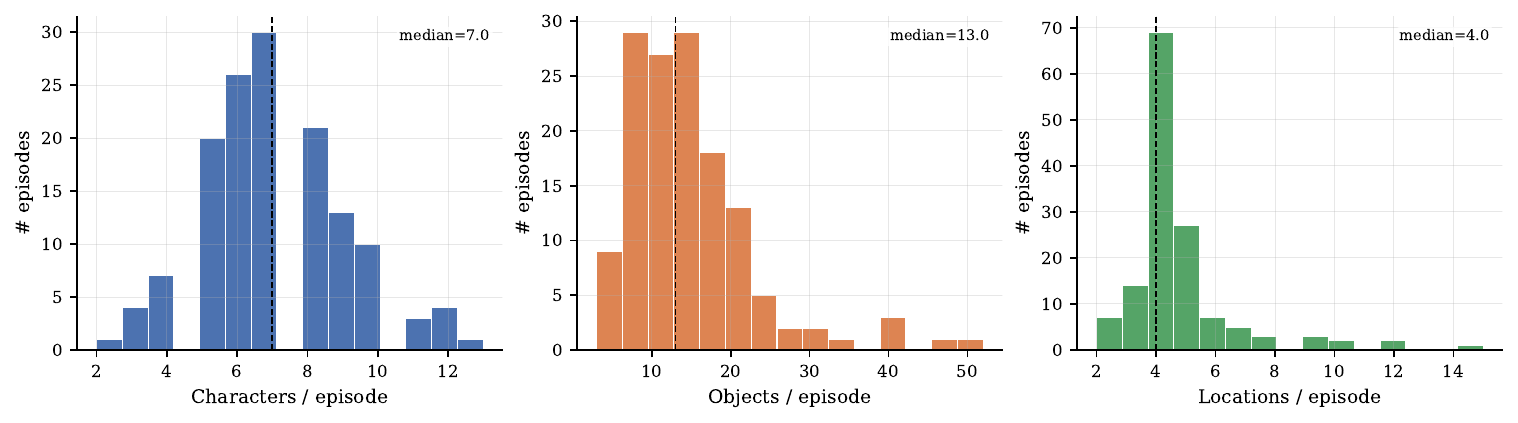}
\caption{Per-episode entity counts (declared in the registry), broken down by entity type.}
\label{fig:appendix:stats:per_episode_entities}
\end{figure}

\paragraph{Entity counts per episode.} An episode declares on average $7$ characters, $5$ locations, and $15$ objects, with the largest episodes declaring up to $13$ characters and $52$ objects. Figure~\ref{fig:appendix:stats:per_episode_entities} shows the per-type histograms. The object distribution has a long right tail: the top 10\% of episodes declare more than $25$ distinct objects, driven by hard-tier episodes that span multiple sub-environments (kitchen, study, garden, etc.) each contributing their own object inventories.

\subsection{Per-Shot Multi-Entity Composition}
\label{appendix:stats:composition}
\begin{figure}[t]
\centering
\includegraphics[width=\linewidth]{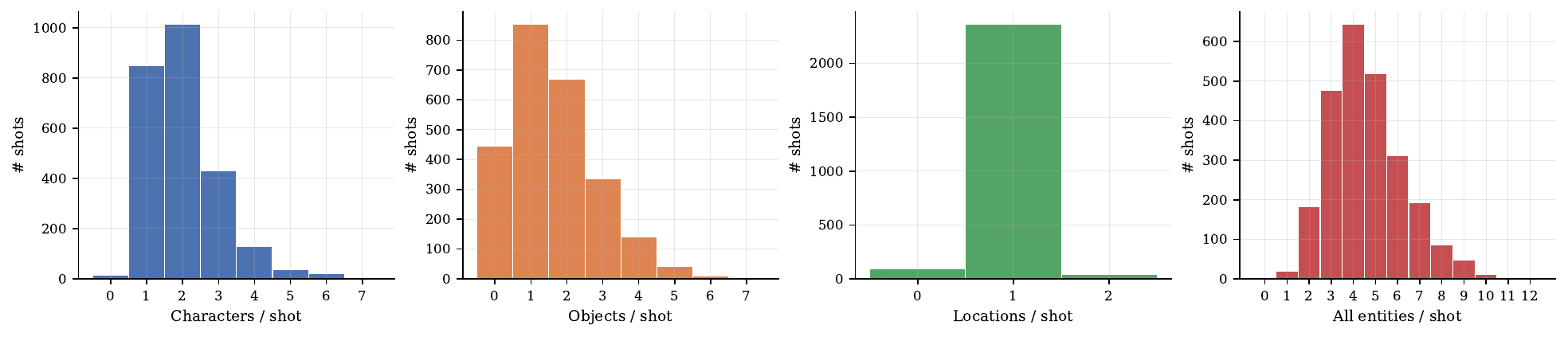}
\caption{Per-shot entity-load distributions, broken down by type. Location counts cluster tightly at $1$ (almost every shot has a single scheduled location), while character and object counts spread across a wide range, with characters concentrated at $1$--$3$ and objects exhibiting a heavier right tail.}
\label{fig:appendix:stats:per_shot_entities}
\end{figure}

A central property of \benchname is that each shot is annotated with a multi-type entity schedule, enabling joint evaluation of character, object, and location consistency rather than evaluation in isolation. Table~\ref{tab:appendix:stats:composition} shows the resulting per-shot composition: the left sub-table reports the mean entity load by type, and the right sub-table reports the fraction of shots satisfying each compositional condition. The mean shot contains $2.0$ characters, $1.6$ objects, and effectively $1$ location, for a mean total entity load of $4.6$ scheduled entities. Beyond raw counts, the compositional breakdown highlights the simultaneous multi-entity test signal that distinguishes \benchname from prior benchmarks: $79.1\%$ of shots schedule at least one entity of \emph{each} of the three types (character, object, location), and $54.3\%$ schedule at least two characters together with at least one object. Figure~\ref{fig:appendix:stats:per_shot_entities} displays the per-type underlying distributions.

\begin{table}[t]
\centering
\caption{Per-shot composition of \benchname. \textbf{Left:} mean entity load by type. \textbf{Right:} fraction of shots satisfying each compositional condition; ``2c+1o'' denotes ``$\geq 2$ characters and $\geq 1$ object,'' and ``tri-type'' denotes simultaneous presence of $\geq 1$ character, $\geq 1$ object, and $\geq 1$ location. Single-entity-type evaluation protocols can only audit a small subset of these compositions.}
\label{tab:appendix:stats:composition}
\small
\begin{minipage}{0.46\linewidth}
\centering
\begin{tabular}{lr}
\toprule
\textbf{Per-shot entity load} & \textbf{Mean} \\
\midrule
Characters / shot      & 2.00 \\
Objects / shot         & 1.61 \\
Locations / shot       & 0.98 \\
\textbf{Total entities / shot} & \textbf{4.59} \\
\bottomrule
\end{tabular}
\end{minipage}\hfill
\begin{minipage}{0.52\linewidth}
\centering
\begin{tabular}{lr}
\toprule
\textbf{Shot-composition fractions} & \textbf{\% shots} \\
\midrule
0 characters                & 0.6\%  \\
Exactly 1 character         & 34.1\% \\
Exactly 2 characters        & 40.6\% \\
$\geq$ 3 characters         & 24.7\% \\
2c+1o (multi-character + object) & 54.3\% \\
\textbf{Tri-type (character + object + location)} & \textbf{79.1\%} \\
\bottomrule
\end{tabular}
\end{minipage}
\end{table}

\subsection{Long-Range Entity Structure}
\label{appendix:stats:structure}

The cross-shot memory signal in \benchname is determined by how entities recur across shot boundaries. We summarize four complementary structural quantities: (i) recurrence rates, (ii) reappearance gap distributions, (iii) the cut/continuation pattern, and (iv) the registry-vs-memory test signal.

\paragraph{Recurrence and cross-scene reappearance.} Of the $3{,}593$ entities scheduled into at least one shot, $2{,}026$ ($56.4\%$) recur in two or more shots, and $1{,}445$ ($40.2\%$) recur across two or more scenes. This places \benchname firmly in the cross-shot regime: the majority of registry entries cannot be evaluated within a single isolated shot, but only by tracking identity across shots.

\paragraph{Reappearance gap.} For each recurring entity we compute the \emph{maximum reappearance gap}: across all consecutive pairs of shots in which the entity appears, the largest number of \emph{intervening} shots that the entity is absent. A gap of $0$ means the entity reappeared in immediately consecutive shots; a gap of $g$ means $g$ intervening shots separate the two closest re-appearances. Figure~\ref{fig:appendix:stats:gap_ccdf} plots the complementary cumulative distribution (CCDF) of this quantity, stratified by tier. The benchmark contains $36.1\%$ of recurring entities with max gap $\geq 5$, $12.4\%$ with max gap $\geq 10$, and $3.5\%$ with max gap $\geq 20$; the global maximum is 48 intervening shots, observed in the hard tier.

\begin{figure}[t]
\centering
\includegraphics[width=0.65\linewidth]{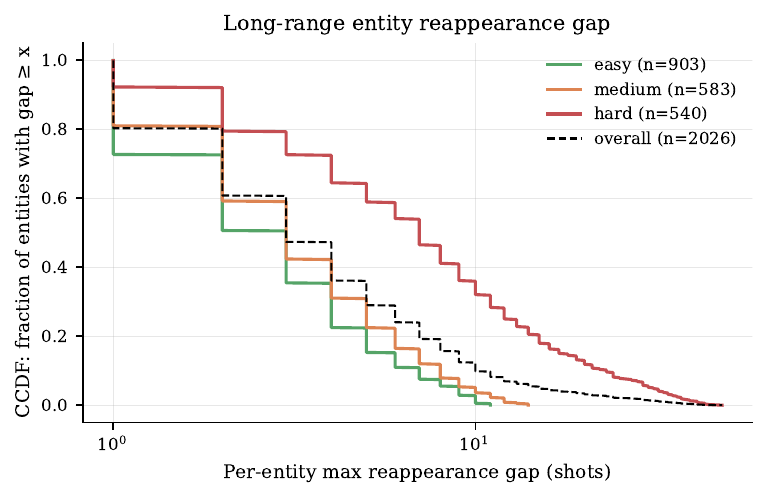}
\caption{Complementary CDF of per-entity maximum reappearance gap, stratified by tier. The hard-tier curve dominates the easy and medium curves at every gap threshold and carries a heavy tail well past 30 intervening shots, providing a long-range stress test that is absent from prior benchmarks. Counts in the legend ($n$) are the numbers of recurring entities in each tier; entities that appear in only one shot are excluded by construction.}
\label{fig:appendix:stats:gap_ccdf}
\end{figure}

\paragraph{Cut and continuation structure.} Each shot is annotated with a binary cut flag, yielding a global cut rate of $45.6\%$ ($1{,}136$ cuts across $2{,}491$ shots). Equivalently, the benchmark partitions into $1{,}136$ \emph{continuation chains}, which represent maximal runs of consecutive shots not separated by a hard cut, with mean length $2.19$ and a maximum chain of $36$ consecutive non-cut shots. The distribution is heavily right-skewed (Figure~\ref{fig:appendix:stats:continuation}): roughly $62\%$ of chains are length-$1$ isolated shots, while the remaining $38\%$ form multi-shot continuation runs that tests the ability of transitioning from and continuing the previous content. A chain of length $k>1$ requires $k-1$ smooth cross-shot transitions in addition to per-shot quality, so the right tail of this distribution (\textbf{e.g.}, chains of $5$ shots and beyond) is the regime that most directly probes transition fidelity at scale.

\begin{figure}[t]
\centering
\includegraphics[width=0.55\linewidth]{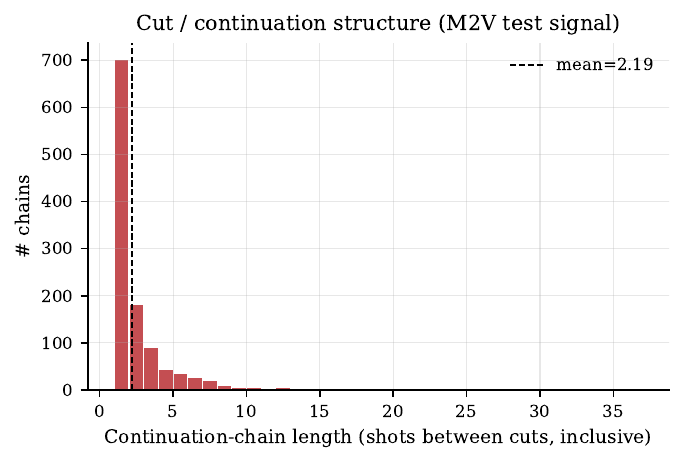}
\caption{Continuation-chain length distribution (number of consecutive shots between two cuts). The bulk of mass at length $1$ corresponds to isolated single-shot scenes; the right tail of multi-shot chains, extending to $36$ shots, examine transitioning ability at scale.}
\label{fig:appendix:stats:continuation}
\end{figure}

\paragraph{Cross-cut entity carry-over.} Of the $996$ within-episode cuts, $555$ ($55.7\%$) preserve at least one entity across the boundary, where a character or object that was present in the last shot before the cut reappears in the first shot after it. Carry-over cuts are particularly difficult: the model must maintain identity across an explicit visual discontinuity, with no continuation context. Pure scene-change cuts (no carry-over, $44.3\%$) could be relatively easier in the consistency sense but force the model to handle an entirely new entity configuration without warm-up.

\paragraph{The memory test signal: re-appearance rate at the entity level.} The cross-shot identity test in \benchname is measured at the level of \emph{entity-slot appearances}. Each (shot, entity) pair is either a \emph{first appearance} in which case the entity's description block is supplied in the shot's prompt header, or a \emph{re-appearance}in which case the entity is referenced by name only and must be rendered from prior context. By construction, the global re-appearance count equals total scheduled appearances minus the number of unique scheduled entities. Across the benchmark, $7{,}852$ of $11{,}445$ entity-slot appearances ($\mathbf{68.6\%}$) are re-appearances and constitute the memory test signal. A shot-level view, where a shot is counted as ``memory-only'' iff \emph{all} of its scheduled entities are re-appearances, understates this, because a shot scheduling one new entity alongside two recurring entities still exercises memory on the two recurring entities even though the shot ships with a registry block. Table~\ref{tab:appendix:stats:memory} reports the breakdown.

\begin{table}[t]
\centering
\caption{Memory test signal of \benchname, measured at the entity-slot level. Each row counts (shot, entity) pairs across the entire benchmark. \emph{First-appearance} pairs ship with a registry description block in the prompt; \emph{re-appearance} pairs reference the entity by name only and must be rendered from episode-level memory of prior appearances.}
\label{tab:appendix:stats:memory}
\small
\begin{tabular}{lrrrr}
\toprule
& \textbf{Characters} & \textbf{Locations} & \textbf{Objects} & \textbf{All entities} \\
\midrule
Total entity-slot appearances & 4{,}989 & 2{,}436 & 4{,}020 & 11{,}445 \\
First appearances (registry block in prompt) & 984 & 648 & 1{,}892 & 3{,}593 \\
Re-appearances (memory test)                 & 4{,}005 & 1{,}788 & 2{,}128 & 7{,}852 \\
\midrule
\textbf{Re-appearance rate} & \textbf{80.3\%} & \textbf{73.4\%} & \textbf{52.9\%} & \textbf{68.6\%} \\
\bottomrule
\end{tabular}
\end{table}

Characters are tested most aggressively. $80.3\%$ of every character slot in the benchmark must be rendered from memory rather than from a prompt-level description. Locations follow at $73.4\%$, while objects, dominated by single-shot props, exhibit the lowest rate at $52.9\%$. The hard tier is even more demanding: $80.7\%$ of all entity-slot appearances in hard episodes are re-appearances (versus $57.7\%$ easy and $64.5\%$ medium), with the per-tier character rate climbing further still.\footnote{Per-tier per-type unique-entity counts are summarized in Section~\ref{appendix:stats:tiers}.}

\begin{figure}[t]
\centering
\includegraphics[width=\linewidth]{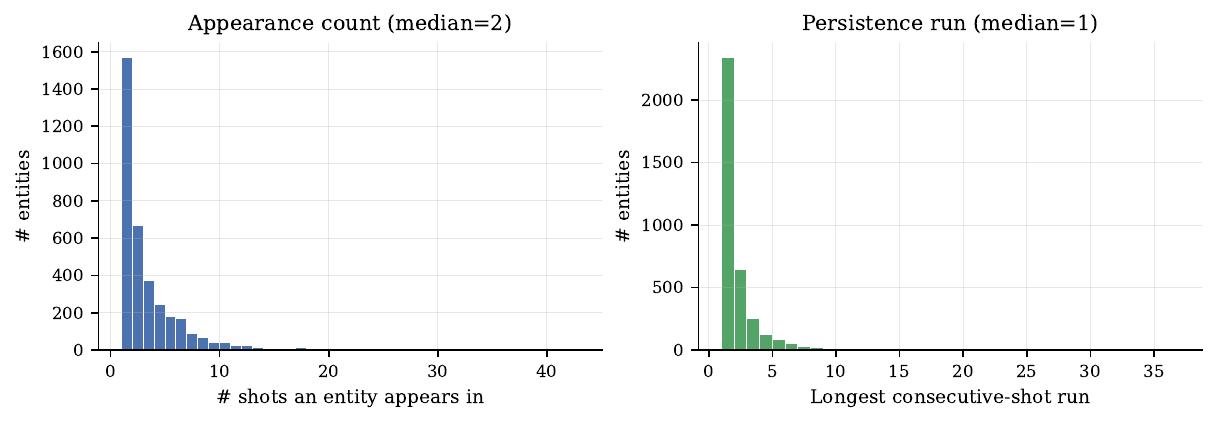}
\caption{Per-entity persistence statistics. \textbf{Left}: number of shots an entity appears in (median $2$; right tail extends past $25$ appearances). \textbf{Right}: longest consecutive-shot run an entity sustains (median $1$; the right tail corresponds to anchor entities across multi-shot continuation segments).}
\label{fig:appendix:stats:persistence}
\end{figure}

\paragraph{Persistence and appearance counts.} Beyond gaps, we also examine entity \emph{persistence}: the longest run of consecutive shots in which an entity appears. The median entity appears in $2$ shots (left panel of Figure~\ref{fig:appendix:stats:persistence}), and roughly two-thirds of entities have a persistence run of $1$ -- they appear, disappear, and possibly recur later, never anchoring a multi-shot continuation. The right tail of the persistence distribution corresponds to anchor entities that drive the narrative across consecutive shots, with persistence runs extending up to $9$ shots.

\begin{figure}[t]
\centering
\includegraphics[width=0.7\linewidth]{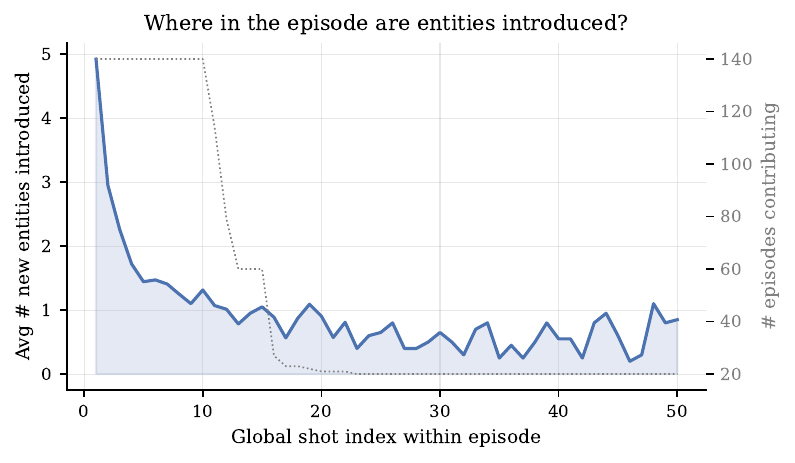}
\caption{Average number of new entities introduced at each shot index (left axis, blue), with the number of episodes contributing at each index (right axis, gray dashed). New entity introductions are heavily front-loaded but never fully stop: the tail beyond shot $\sim 22$ reflects the 20 hard-tier episodes, which continue to introduce entities at a steady $\sim 0.5$--$1.0$ rate throughout their 50-shot length.}
\label{fig:appendix:stats:new_entities_curve}
\end{figure}

\paragraph{Where in an episode are entities introduced?} Figure~\ref{fig:appendix:stats:new_entities_curve} plots the average number of new entities introduced at each shot index, averaged across all episodes that contain at least that many shots. The first shot of an episode introduces, on average, $\sim 5$ new entities (the opening establishes the cast and setting), and roughly $70\%$ of an episode's entity inventory is introduced within the first $10$ shots. The curve then plateaus at $\sim 0.5$--$1.0$ new entities per shot for the remainder of the episode---hard-tier episodes (the only ones that contribute to shot indices beyond $\sim 22$) continue to introduce entities at a steady drip well past the midpoint. This shape implies that \benchname does not partition cleanly into an ``introduction phase'' followed by a ``recall phase''; instead, models must handle both regimes simultaneously throughout long episodes, with the recall burden growing monotonically while introductions never fully cease.

\subsection{Linguistic Profile and Scene-Design Tags}
\label{appendix:stats:linguistic}

\benchname prompts are derived from natural narrative scripts rather than synthesized from a fixed template, which is reflected in their linguistic statistics. Across all $2{,}491$ action descriptions the vocabulary contains $5{,}230$ distinct word forms over $93{,}123$ total tokens, yielding a type/token ratio of $0.056$. Action descriptions average $37.4$ words (median $37$); full prompts including the registry header average longer due to the prepended entity descriptions. From a curated set of approximately $460$ inflected English action-verb forms we identify $326$ distinct verbs in use; Table~\ref{tab:appendix:stats:topverbs} lists the top-18 action-verb lemmas after merging inflections (stand/stands/standing count as one) and excluding state descriptors (wear, light, glow) and ambiguous noun forms (face, head, hand). The resulting inventory mixes \emph{posture} (stand, sit, lean; $33\%$ of the top-18 mass), \emph{perception and gaze} (look, watch, gaze, stare, observe, glance; $26\%$), \emph{dialogue} (speak, talk, listen; $19\%$), and \emph{motion} (walk, turn, hold; $15\%$). The relative weight of posture/perception/dialogue ($78\%$) over motion ($15\%$) is a property worth noting because most prior video-generation benchmarks favor high-motion prompts~\citep{huang2024vbench,huang2025vbench++} with a dedicated Dynamic Degree dimension that explicitly penalizes static videos, and dedicated motion benchmarks~\citep{ling2025vmbench,liu2024evalcrafter} structure their entire prompt suite around motion patterns. \benchname is complementary: 
with motion held subtle, the visual evaluation budget shifts to entity-level identity preservation, which is the consistency property our benchmark targets.

\begin{table}[t]
\centering
\caption{Top-18 most frequent action-verb lemmas across all $2{,}491$ action descriptions, with raw counts. Inflections are merged under their lemma; state descriptors (\emph{wear/wearing}, \emph{light/lit/illuminated}, \emph{glow/glowing}) and ambiguous noun-dominant forms (\emph{face, head, hand, hands}) are excluded. Verb extraction uses a curated $\sim$460-form English verb list rather than a POS tagger, so rarer or domain-specific verbs may be undercounted, but the relative ordering is informative.}
\label{tab:appendix:stats:topverbs}
\small
\begin{tabular}{lr@{\hspace{2em}}lr@{\hspace{2em}}lr}
\toprule
\textbf{Verb} & \textbf{\#} & \textbf{Verb} & \textbf{\#} & \textbf{Verb} & \textbf{\#} \\
\midrule
stand & 1{,}023 & listen & 292 & gaze    & 128 \\
look  & 711     & smile  & 237 & nod     & 80  \\
sit   & 539     & hold   & 234 & stare   & 67  \\
speak & 530     & turn   & 231 & observe & 54  \\
watch & 369     & talk   & 194 & glance  & 47  \\
walk  & 315     & lean   & 143 & show    & 44  \\
\bottomrule
\end{tabular}
\end{table}

\paragraph{Scene-design tag rates.} Lexical-tag coverage on action descriptions is uneven: shot type is named in $52.5\%$ of descriptions (close-ups and extreme close-ups together account for $61\%$ of those), indoor/outdoor in $24.5\%$, time of day in $19.2\%$ ($2.7\times$ more night than day), and explicit visual-style tags in only $2.2\%$. We do not use these tags as inputs to any evaluation metric, as our concentration is on entity consistency. They are reported here as a profile of the prompt corpus.

\subsection{Tier-Stratified Comparison}
\label{appendix:stats:tiers}

Table~\ref{tab:appendix:stats:tiers} extends Table~\ref{tab:benchmark_stats} of the main paper with fine-grained per-tier statistics. It shows that \benchname's difficulty axis isolates long-range memory burden specifically. Per-shot composition is essentially constant across tiers, such as, mean characters per shot, multi-character rate, and tri-type rate all vary by less than two percentage points from easy to hard. However, the long-range memory load scales sharply. From easy to hard, the mean per-entity max gap triples ($3.2 \to 9.7$ shots), the global maximum gap quadruples ($11 \to 48$), and the entity-slot re-appearance rate climbs from $\sim 58\%$ to $\sim 81\%$. This separation is by design: it evaluate methods that target long-range identity preservation against tier-level scaling without confounding from increased intra-shot complexity, which would be a separate and orthogonal failure mode.

\begin{table}[t]
\centering
\caption{Tier-stratified statistics. Rows that also appear in the main paper's Table~\ref{tab:benchmark_stats} (scale, shots/episode, recurrence gap) are not duplicated here; this table reports the additional dimensions made available by the released annotations. ``$\geq 3$ chars'' is the fraction of shots scheduling at least three characters; ``2c+1o'' is the fraction with $\geq 2$ characters and $\geq 1$ object. ``Re-appearance rate'' is the fraction of entity-slot appearances that are re-appearances (Section~\ref{appendix:stats:structure}); ``memory-only rate'' is the stricter shot-level analog (a shot is memory-only iff \emph{all} of its scheduled entities are re-appearances).}
 \label{tab:appendix:stats:tiers}
 \small
 \begin{tabular}{lrrrr}
 \toprule
 & \textbf{Easy} & \textbf{Medium} & \textbf{Hard} & \textbf{All} \\
 \midrule
 Episodes & 80 & 40 & 20 & 140 \\
 Shots    & 873 & 618 & 1{,}000 & 2{,}491 \\
 Scheduled entities (unique) & 1{,}694 & 1{,}021 & 878 & 3{,}593 \\
 \midrule
 \multicolumn{5}{l}{\textit{Per-entity reappearance (recurring entities only)}} \\
 Recurring-entity rate           & 53.3\% & 57.1\% & 61.5\% & 56.4\% \\
 Mean per-entity max gap         & 3.24   & 3.85   & 9.72   & 5.14 \\
 Median per-entity max gap       & 3.0    & 3.0    & 7.0    & 3.0 \\
 Global max gap                  & 11     & 14     & \textbf{48} & 48 \\
 \midrule
 \multicolumn{5}{l}{\textit{Per-shot composition}} \\
 Mean characters / shot          & 1.99   & 2.06   & 1.98   & 2.00 \\
 Max characters / shot           & 6      & 7      & 6      & 7 \\
 Frac.\ shots $\geq$ 3 chars     & 23.5\% & 27.0\% & 24.4\% & 24.7\% \\
 Frac.\ shots 2c+1o              & 55.4\% & 55.3\% & 52.7\% & 54.3\% \\
 \midrule
 \multicolumn{5}{l}{\textit{Memory test signal}} \\
+Entity-slot re-appearance rate  & $\sim$57.7\% & $\sim$64.5\% & \textbf{$\sim$80.7\%} & 68.6\% \\
 Cut rate                        & 54.3\% & 39.5\% & 41.8\% & 45.6\% \\
 Registry-shot rate              & 69.5\% & 62.0\% & \textbf{37.3\%} & 54.7\% \\
 Memory-only rate (shot-level)   & 30.5\% & 38.0\% & \textbf{62.7\%} & 45.3\% \\
 \bottomrule
 \end{tabular}
 \end{table}

\section{Evaluation Metrics}
\label{appendix:metrics}

This section specifies every metric in our evaluation suite formally. We begin with notation in Section~\ref{appendix:metrics:notation}, then detail each pillar in turn: intra-shot quality (Section~\ref{appendix:metrics:pillar1}), cross-shot consistency (Section~\ref{appendix:metrics:pillar2}), and intra-shot prompt-following alignment (Section~\ref{appendix:metrics:pillar3}). Section~\ref{appendix:metrics:strict} describes the strict-mode reproducibility contract that governs all aggregations.

\subsection{Notation}
\label{appendix:metrics:notation}

\paragraph{Episodes, scenes, and shots.} An episode $E = (S_1, S_2, \ldots, S_K)$ is a sequence of $K$ shots, each a video clip $S_k$ of $F_k$ frames at fixed resolution. Each shot belongs to a scene $\mathcal{S}$, and a scene cut at shot $k$ is indicated by an attribute $\mathrm{cut}(k) \in \{\mathsf{True}, \mathsf{False}\}$, with $\mathrm{cut}(k) = \mathsf{True}$ when shot $k$ begins a new scene and $\mathrm{cut}(k) = \mathsf{False}$ when shot $k$ continues the previous shot. Frames of shot $k$ are denoted $f_{k,1}, \ldots, f_{k,F_k}$.

\paragraph{Entity registry and schedule.} Each episode is equipped with an entity registry $\mathcal{E} = \mathcal{E}^{\mathrm{char}} \cup \mathcal{E}^{\mathrm{obj}} \cup \mathcal{E}^{\mathrm{loc}}$ partitioned into characters, objects, and locations. Each entity $e \in \mathcal{E}$ has a textual description $\mathrm{desc}(e)$. The script associates each shot $k$ with a scheduled entity set $\mathcal{E}_k \subseteq \mathcal{E}$ (the entities expected to appear in shot $k$) and an action description $a_k$ (free-form text describing what happens in the shot).

\paragraph{Visual encoders.} We use three frozen pretrained encoders throughout. Let $\phi_{\mathrm{DINO}} : \mathbb{R}^{H \times W \times 3} \to \mathbb{S}^{767}$ denote DINOv2-base~\citep{oquab2023dinov2} CLS embeddings (unit-normalized, 768-dim sphere); $\phi_{\mathrm{CLIP}}^{\mathrm{img}}$ and $\phi_{\mathrm{CLIP}}^{\mathrm{txt}}$ denote CLIP ViT-B/32~\citep{radford2021learning} image and text embeddings respectively (jointly trained, 512-dim, unit-normalized). For an image $x$ and text $t$, the CLIP text-image similarity is
\begin{equation}
    \mathrm{CLIPsim}(x, t) = \phi_{\mathrm{CLIP}}^{\mathrm{img}}(x)^\top \phi_{\mathrm{CLIP}}^{\mathrm{txt}}(t) \in [-1, 1].
\end{equation}

\paragraph{Grounding.} Let $G$ denote the GroundingDINO\citep{liu2024grounding} detector with text encoder \texttt{bert-base-uncased}. For frame $f$ and query $q$, $G(f, q)$ returns a (possibly empty) set of detections $\{(b_i, p_i)\}_i$ where $b_i \subset f$ is a bounding box (xyxy pixel coordinates) and $p_i \in [0, 1]$ is the model's confidence. We threshold detections at $\tau_{\mathrm{box}} = 0.25$ and the per-token text alignment at $\tau_{\mathrm{text}} = 0.20$. The crop operator $\mathrm{Crop}(f, b)$ extracts the pixel region inside $b$ with a $10\%$ padding margin, then resizes to $224 \times 224$ for embedding.

\paragraph{LLM judgement.} Let $M_{\mathrm{LLM}}$ denote the multimodal LLM \texttt{gemini-2.5-pro}~\citep{comanici2025gemini}, which we treat as an oracle returning structured JSON conditioned on a list of images and a textual prompt: $M_{\mathrm{LLM}}(\{x_1, \ldots, x_n\}, t) \to J$ where $J$ is a parsed dictionary. Per-criterion scores returned on a 1--10 scale are normalized to $[0, 1]$ via $s \mapsto s/10$. 

\paragraph{Aggregation conventions.} For a list of values $V = (v_1, \ldots, v_n)$, we write $\mathrm{mean}(V)$ for the sample mean, $\mathrm{median}(V)$ for the median, and $\|V\|$ for the cardinality $n$. A value of $\mathsf{None}$ is excluded from any aggregation; if all values are $\mathsf{None}$, the aggregate is also $\mathsf{None}$ (never substituted with $0$). See Section~\ref{appendix:metrics:strict} for the formal contract.

\paragraph{Human validation of LLM judgement.} Because every per-entity fidelity score (Pillar 2) and cross-shot identity score (Pillar 3) ultimately depends on LLM, we conducted a human-agreement study to verify that the LLM judge produces decisions consistent with human raters. We sampled (shot, entity) pairs uniformly across the four evaluated methods and across all three difficulty tiers, stratified to include equal numbers of character, object, and location instances, and balanced between gate-passing and gate-failing cases. 
For each sampled instance, 3 independent human annotators who are research scientists in generative AI, were shown the same canonical crop and registry description used by LLM, and asked to provide both the binary present/absent verdict and the per-criterion fidelity scores on the same 1--10 scale. For cross-shot identity, annotators received the same anchor-vs-each pairwise format described in Section~\ref{appendix:metrics:pillar3}, with the binary same/different verdict as the primary outcome. We report agreement using Cohen's $\kappa$ for the binary verdicts and Pearson's $r$ for the continuous scores, computed both between LLM and the human majority vote and between individual human raters as an upper bound on achievable agreement. Across the 200 samples, LLM achieved $\kappa$ = 0.93 on intra-shot presence, $\kappa$ = 0.94 on cross-shot identity verdicts, falling within the inter-human range of $\kappa$ = [0.8, 0.96]. Disagreement cases were concentrated in (i) half face distortion and (ii) blurry features with dim lighting. We treat these as inherent to the task rather than judge-specific failures. The agreement levels support the use of LLM as the operational judge throughout the benchmark, with the caveat that all reported metrics inherit a residual uncertainty bounded by the LLM--human gap.

\subsection{Pillar 1: Intra-Shot Quality}
\label{appendix:metrics:pillar1}

Inspired by~\citet{huang2024vbench}, we adopt six standard intra-shot quality metrics that capture whether each shot is technically well-formed in isolation. We drop the \texttt{background\_consistency} metric from the original VBench suite because it measures within-shot CLIP cosine on consecutive frames, which is confounded by intentional camera motion: a pan or zoom of a stable background is incorrectly penalized as inconsistency. For our long-range multi-shot benchmark where camera motion is common, this metric may be inaccurate. The remaining six are computed per shot and averaged across the episode.

For shot $S_k$ with frames $f_{k,1}, \ldots, f_{k,F_k}$:

\paragraph{Subject consistency} (range $[0, 1]$):
\begin{equation}
    \mathrm{SC}(S_k) = \frac{1}{F_k - 1} \sum_{i=1}^{F_k - 1} \phi_{\mathrm{DINO}}(f_{k,i})^\top \phi_{\mathrm{DINO}}(f_{k,i+1}).
\end{equation}
Measures stability of the dominant subject within the shot.

\paragraph{Temporal flickering} (range $[0, 1]$):
\begin{equation}
    \mathrm{TF}(S_k) = 1 - \frac{1}{F_k - 1} \sum_{i=1}^{F_k - 1} \mathrm{MAE}(f_{k,i}, f_{k,i+1}),
\end{equation}
where $\mathrm{MAE}$ is the mean absolute pixel difference normalized to $[0, 1]$. Penalizes high-frequency flicker.

\paragraph{Motion smoothness} (range $[0, 1]$): RAFT~\citep{teed2020raft} optical flow is used to interpolate intermediate frames; $\mathrm{MS}(S_k)$ is the mean reconstruction quality of the interpolation, with higher values indicating smoother apparent motion. Implementation follows~\citet{huang2024vbench}.

\paragraph{Dynamic degree} (range $[0, 1]$): the fraction of inter-frame pairs whose RAFT optical flow magnitude exceeds a threshold; penalizes static slideshow-like outputs.

\paragraph{Aesthetic quality} (range $[0, 1]$):
\begin{equation}
    \mathrm{AQ}(S_k) = \frac{1}{F_k} \sum_{i=1}^{F_k} \mathrm{MLP}_{\mathrm{LAION}}\!\left(\phi_{\mathrm{CLIP}}^{\mathrm{img}}(f_{k,i})\right),
\end{equation}
where $\mathrm{MLP}_{\mathrm{LAION}}$ is the LAION aesthetic predictor head trained on human aesthetic ratings.

\paragraph{Imaging quality} (range $[0, 100]$):
\begin{equation}
    \mathrm{IQ}(S_k) = \frac{1}{F_k} \sum_{i=1}^{F_k} \mathrm{MUSIQ}(f_{k,i}),
\end{equation}
where $\mathrm{MUSIQ}$ is the no-reference image quality predictor of~\citet{ke2021musiq}. We report this on its canonical $[0, 100]$ scale rather than normalizing to $[0, 1]$, to maintain comparability with the broader literature.

\paragraph{Episode-level aggregation.} For each Pillar 1 metric $m$, the episode-level value is the mean over admissible shots:
\begin{equation}
    m(E) = \frac{1}{K} \sum_{k=1}^{K} m(S_k).
\end{equation}

\subsection{Pillar 2: Intra-Shot Prompt-Following Alignment}
\label{appendix:metrics:pillar2}
 
Pillar 2 measures, for each shot in isolation, three aspects of prompt-following. (i) entity presence: do the scheduled entities actually appear? (ii) per-entity fidelity: when an entity does appear, does it match its registry description? (iii) action fidelity: does the shot depict the action described in the script? All three sub-evaluations are built on a unified grounding pass, described next, that is also reused by Pillar 3 (Section~\ref{appendix:metrics:pillar3}). The same canonical crop saved per (shot, entity) pair is the exact image used for fidelity judging \emph{and} for cross-shot comparison. This ensures the audit chain from headline metric to underlying pixels is consistent, representing the review process of drilling into a cross-shot score for a specific entity sees exactly the crops that produce that score.
 
\subsubsection{Unified grounding pass}
\label{appendix:metrics:pillar2:grounding}
 
For each shot $S_k$ and each scheduled entity $e \in \mathcal{E}_k$, we compute a canonical crop $c^*(k, e)$ as follows. We sample $N_{\mathrm{frame}} = 5$ frames evenly across the shot. For each frame $f_{k,i}$ we run grounding $G(f_{k,i}, \mathrm{desc}(e))$ to obtain candidate detections, and for each detection we compute three quality components:
\begin{align}
    \alpha_{\mathrm{clip}}(f_{k,i}, b) &= \mathrm{CLIPsim}\big(\mathrm{Crop}(f_{k,i}, b), \mathrm{desc}(e)\big), \\
    \alpha_{\mathrm{sharp}}(f_{k,i}, b) &= \sigma\!\left(\frac{\mathrm{LapVar}(\mathrm{Crop}(f_{k,i}, b)) - 100}{200}\right), \\
    \alpha_{\mathrm{area}}(f_{k,i}, b) &= \sigma\!\left(\frac{\mathrm{AreaPct}(b, f_{k,i}) - 2}{5}\right),
\end{align}
where $\mathrm{LapVar}$ is the variance of the Laplacian of luminance as a standard sharpness proxy, $\mathrm{AreaPct}$ is the bounding-box area as a percentage of frame area, and $\sigma(z) = (1+e^{-z})^{-1}$ is the logistic. The composite selection score is the product
\begin{equation}
    \alpha(f_{k,i}, b) = \alpha_{\mathrm{clip}}(f_{k,i}, b) \cdot \alpha_{\mathrm{sharp}}(f_{k,i}, b) \cdot \alpha_{\mathrm{area}}(f_{k,i}, b).
    \label{eq:selection-score}
\end{equation}
Among all (frame, detection) pairs for entity $e$ in shot $k$, the canonical crop is the argmax:
\begin{equation}
    c^*(k, e) = \mathrm{Crop}(f_{k,i^*}, b^*), \qquad (i^*, b^*) = \mathop{\mathrm{arg\,max}}_{(i, b) \in G_k(e)} \alpha(f_{k,i}, b),
\end{equation}
where $G_k(e) = \bigcup_{i=1}^{N_{\mathrm{frame}}} G(f_{k,i}, \mathrm{desc}(e))$ is the union of all detections for entity $e$ across the sampled frames.
 
The selection score balances three quality aspects. Crops with high CLIP score but motion blur lose on sharpness; sharp and large but wrong-entity crops lose on CLIP; right-entity sharp but tiny crops lose on area. All three components must be high for the score to be high; a crop with any one near zero is rejected.
 
\paragraph{Presence status.} Each canonical crop is assigned a tri-valued status:
\begin{equation}
    \mathrm{status}(k, e) = \begin{cases}
        \texttt{absent} & G_k(e) = \emptyset, \\
        \texttt{weak}   & G_k(e) \neq \emptyset \text{ and } \alpha_{\mathrm{clip}}(c^*(k,e), \mathrm{desc}(e)) < \tau_{\mathrm{CLIP}}, \\
        \texttt{present}& G_k(e) \neq \emptyset \text{ and } \alpha_{\mathrm{clip}}(c^*(k,e), \mathrm{desc}(e)) \geq \tau_{\mathrm{CLIP}},
    \end{cases}
\end{equation}
with $\tau_{\mathrm{CLIP}} = 0.20$. Under any model that fails to render the right entity, GroundingDINO either returns nothing (\texttt{absent}) or returns a hallucinated box rejected by CLIP (\texttt{weak}); only \texttt{present} appearances are confidently the scheduled entity.
 
\subsubsection{Presence}
\label{appendix:metrics:pillar2:presence}
 
For each entity type $\mathcal{T}$, the per-shot presence rate is the fraction of scheduled entities of that type that achieved status \texttt{present} in the shot:
\begin{equation}
    \rho^{\mathcal{T}}(S_k) = \frac{|\{e \in \mathcal{E}_k \cap \mathcal{E}^{\mathcal{T}} : \mathrm{status}(k, e) = \texttt{present}\}|}{|\mathcal{E}_k \cap \mathcal{E}^{\mathcal{T}}|},
\end{equation}
with the convention that $\rho^{\mathcal{T}}(S_k) = \mathsf{None}$ when the denominator is zero (\textit{i.e.}, the shot has no scheduled entities of this type). The episode-level metric is the mean over shots that scheduled at least one entity of the type:
\begin{equation}
    \mathrm{intra\_character\_presence}(E) = \mathrm{mean}\big(\{\rho^{\mathrm{char}}(S_k) : k \in [K], \; \rho^{\mathrm{char}}(S_k) \neq \mathsf{None}\}\big),
\end{equation}
and analogously for $\mathrm{intra\_object\_presence}$ and $\mathrm{intra\_location\_presence}$. Note that absent entities pull down the rate (\textit{e.g.}, a shot scheduling 2 characters with only 1 detected contributes $0.5$ to the mean), while shots scheduling no entities of a type are skipped rather than contributing $1.0$, which would inflate the metric.
 
\subsubsection{Per-entity fidelity}
\label{appendix:metrics:pillar2:fid}
 
For each (shot, entity) pair with $\mathrm{status}(k, e) \in \{\texttt{present}, \texttt{weak}\}$, we send the canonical crop $c^*(k, e)$ to $M_{\mathrm{LLM}}$ along with the entity's textual description:
\begin{equation}
    J_{k,e} = M_{\mathrm{LLM}}\big(\{c^*(k, e)\}, \pi_{\mathrm{fid}}(\mathrm{desc}(e), \mathcal{T}_e, \mathrm{status}(k, e))\big).
\end{equation}
Appearances with status \texttt{weak} are still scored, but the resulting fidelity values are flagged as low-confidence in the audit JSON because the underlying crop did not pass the CLIP threshold. The LLM returns an overall fidelity score $\phi(k, e) \in [0, 1]$ and four per-criterion scores $\phi^j(k, e) \in [0, 1]$ for $j \in \mathcal{J}^{\mathcal{T}_e}$, where the per-type criterion sets are:
\begin{align*}
    \mathcal{J}^{\mathrm{char}} &= \{\mathrm{face}, \; \mathrm{hair}, \; \mathrm{clothing}, \; \mathrm{build}\}, \\
    \mathcal{J}^{\mathrm{obj}}  &= \{\mathrm{shape}, \; \mathrm{color\_texture}, \; \mathrm{proportions}, \; \mathrm{details}\}, \\
    \mathcal{J}^{\mathrm{loc}}  &= \{\mathrm{layout}, \; \mathrm{color\_mood}, \; \mathrm{landmarks}, \; \mathrm{perspective}\}.
\end{align*}
The same criterion sets are reused identically in Pillar 3 (Section~\ref{appendix:metrics:pillar3:llm}), enabling direct comparison between within-shot fidelity and cross-shot consistency on the same axes.
 
\paragraph{Episode-level aggregation.} For each entity type $\mathcal{T}$ and shot $S_k$, the per-shot mean fidelity across that type's entities is
\begin{equation}
    \bar{\phi}^{\mathcal{T}}(S_k) = \mathrm{mean}\big(\{\phi(k, e) : e \in \mathcal{E}_k \cap \mathcal{E}^{\mathcal{T}}, \; \phi(k, e) \neq \mathsf{None}\}\big),
\end{equation}
and the episode-level fidelity metric is the mean over shots:
\begin{equation}
    \mathrm{intra\_face\_fidelity}(E) = \mathrm{mean}\big(\{\bar{\phi}^{\mathrm{char}}(S_k) : k \in [K], \; \bar{\phi}^{\mathrm{char}}(S_k) \neq \mathsf{None}\}\big),
\end{equation}
and analogously for $\mathrm{intra\_object\_fidelity}$ and $\mathrm{intra\_location\_fidelity}$. Per-criterion metrics are defined identically with $\phi^j$ in place of $\phi$:
\begin{equation}
    \mathrm{intra\_face\_}j(E) = \mathrm{mean}\big(\{\bar{\phi}^{\mathrm{char},j}(S_k) : k \in [K], \; \bar{\phi}^{\mathrm{char},j}(S_k) \neq \mathsf{None}\}\big),
\end{equation}
where $\bar{\phi}^{\mathrm{char},j}(S_k)$ is the per-shot mean of $\phi^j(k, e)$ over scheduled characters. This yields 5 metrics per entity type (one overall, four per-criterion), for 15 fidelity metrics in total.
 
The fidelity scores $\phi(k, e)$ from this section are reused by Pillar 3's cross-shot fidelity gate (Eq.~\ref{eq:fidelity-gate}, Section~\ref{appendix:metrics:pillar3:gate}). 
 
\subsubsection{Action fidelity}
\label{appendix:metrics:pillar2:action}
 
To evaluate whether shot $S_k$ depicts its action description $a_k$, we construct a labeled multi-frame grid that explicitly resolves the visual identity of each subject in the action.
 
\paragraph{Labeled action grid.} We sample 6 frames evenly across the shot. For each frame $f_{k,i}$ and each scheduled entity $e \in \mathcal{E}_k$ (characters and objects only; locations are omitted from the grid), we draw the bounding box of the highest-confidence detection from $G_k(e)$ on $f_{k,i}$, with a unique color assigned to entity $e$ across the entire grid. The text label is the entity name. The 6 annotated frames are tiled into a $2 \times 3$ grid image $A_k$.
The colored labeled boxes help identify the characters so that the LLM can then assess the directional language unambiguously.
 
\paragraph{LLM judgment.} The grid is sent to $M_{\mathrm{LLM}}$ with a prompt parameterized by the action description and the labeling legend:
\begin{equation}
    J_k^{\mathrm{action}} = M_{\mathrm{LLM}}\big(\{A_k\}, \pi_{\mathrm{action}}(a_k, \mathrm{legend})\big).
\end{equation}
The LLM returns six values per shot:
\begin{align*}
    \mathrm{ovr}_k &\in [0, 1] && \text{overall action-fidelity score}, \\
    \mathrm{dep}_k &\in \{0, 1\} && \text{binary verdict on whether the action is depicted}, \\
    \mathrm{ai}_k  &\in [0, 1] && \text{subject identity: are the labeled boxes the right characters?} \\
    \mathrm{aa}_k  &\in [0, 1] && \text{subject action: does the named subject perform the verb?} \\
    \mathrm{ao}_k  &\in [0, 1] \cup \{\mathsf{None}\} && \text{object interaction; \textsf{None} if no object referenced in } a_k, \\
    \mathrm{am}_k  &\in [0, 1] && \text{motion quality: is motion natural across frames?}
\end{align*}
 
\paragraph{Episode-level aggregation.} Each of the six action metrics is the mean over shots for which the corresponding value is not $\mathsf{None}$:
\begin{align}
    \mathrm{intra\_action\_overall}(E)   &= \mathrm{mean}\big(\{\mathrm{ovr}_k : k \in [K], \; \mathrm{ovr}_k \neq \mathsf{None}\}\big), \\
    \mathrm{intra\_action\_depicted}(E)  &= \mathrm{mean}\big(\{\mathrm{dep}_k : k \in [K], \; \mathrm{dep}_k \neq \mathsf{None}\}\big),
\end{align}
and similarly for $\mathrm{intra\_action\_subject\_identity}$, $\mathrm{intra\_action\_subject\_action}$, $\mathrm{intra\_action\_object\_interaction}$, and $\mathrm{intra\_action\_motion\_quality}$. The object-interaction metric in particular has a smaller denominator: only shots whose action description explicitly references an object contribute, since asking "did the action use the object correctly" is meaningless for actions like "[character 1] walks toward the door" that do not name an object. This brings the action sub-evaluation to 6 metrics, and Pillar 2 to 24 (3+15+6) metrics total.
 
\subsection{Pillar 3: Cross-Shot Consistency}
\label{appendix:metrics:pillar3}
 
Pillar 3 measures whether scheduled entities maintain identical across the shots in which they appear. It is the core of \benchname's evaluation for long-range cross-shot entity consistency. 
The pillar reuses the canonical crops $c^*(k, e)$ produced by the unified grounding pass in Pillar 2 (Section~\ref{appendix:metrics:pillar2:grounding}), and reuses Pillar 2's per-shot fidelity scores $\phi(k, e)$ to admit only well-rendered appearances into the cross-shot pool. The pillar comprises three stages: (i) an admissibility gate built from the Pillar 2 fidelity scores, (ii) DINOv2-based metrics (Section~\ref{appendix:metrics:pillar3:dino}) that score each appearance against the appearance centroid, and (iii) LLM-based metrics (Section~\ref{appendix:metrics:pillar3:llm}) that score appearances pairwise against a centroid-representative anchor.
 
\subsubsection{Cross-shot fidelity gate}
\label{appendix:metrics:pillar3:gate}

Even \texttt{present} appearances may render the entity poorly. Without further filtering, a method that produces nearly-static frames (\textit{e.g.}, the same low-quality rendering repeated) would be rewarded with high consistency. We prevent this with a fidelity gate keyed on Pillar 2's per-shot fidelity scores.

For each (shot, entity) pair, recall that $\phi(k, e) \in [0, 1] \cup \{\mathsf{None}\}$ is the intra-shot fidelity score from Pillar 2 (Section~\ref{appendix:metrics:pillar2:fid}). The cross-shot pool for entity $e$ is defined as
\begin{equation}
    \mathcal{C}(e) = \big\{ c^*(k, e) : \mathrm{status}(k, e) = \texttt{present} \;\text{and}\; (\phi(k, e) \geq \tau_{\mathrm{fid}} \;\text{or}\; \phi(k, e) = \mathsf{None}) \big\},
    \label{eq:fidelity-gate}
\end{equation}
with $\tau_{\mathrm{fid}} = 0.5$. The disjunction with $\mathsf{None}$ ensures that appearances for which Pillar 2 could not be computed (e.g., LLM call failure) are admitted by default rather than silently dropped, with the fact that they bypassed the gate logged for audit. The number of gated-out appearances is recorded per episode in the auxiliary metric \texttt{\_meta\_cross\_shot\_gate}.

\paragraph{Fidelity-gate-corrected aggregation.}
The gate filters which instances enter cross-shot computation, but a method that fails the gate on most of its outputs should not benefit from being scored only on the few it passes. We therefore aggregate per-entity metrics with an \emph{instance-weighted, gate-corrected mean} that treats gate-skipped and gate-failed instances as zero contributions.

Let $m \in \mathcal{M}^{\mathrm{ent}}$ denote a per-entity metric (any metric in Pillars 2 and 3 except presence and Pillar 1 VBench dimensions). For each episode $E$, let $v_E^m \in [0, 1] \cup \{\mathsf{None}\}$ be the episode-level value of $m$, and let $n_E^{\mathrm{eval}, m}$, $n_E^{\mathrm{skip}, m}$, $n_E^{\mathrm{fail}, m}$ count the underlying entity-instances (per-shot pairs for intra-shot metrics, per-comparison pairs for cross-shot metrics, locations for scene metrics) that respectively (i) passed the gate and were scored, (ii) were dropped by the fidelity gate, and (iii) failed at the LLM-call or grounding step. The aggregated metric for a method across the benchmark is
\begin{equation}
    \overline{m} \;=\; \frac{\sum_{E \,:\, v_E^m \neq \mathsf{None}} v_E^m \cdot n_E^{\mathrm{eval}, m}}
                            {\sum_{E} \big(n_E^{\mathrm{eval}, m} + n_E^{\mathrm{skip}, m} + n_E^{\mathrm{fail}, m}\big)}.
    \label{eq:fidelity-corrected-mean}
\end{equation}
The numerator weights each episode's score by how many gate-passing instances it contributed, so episodes with more recurring entities (which carry more cross-shot evidence) are weighted accordingly. The denominator includes \emph{all} eligible instances across all benchmark episodes, so a method failing the gate on a hard episode is correctly penalized.

Equivalently, $\overline{m}$ can be written as $\mathrm{rawmean}(m) \times \mathrm{coverage}(m)$, where
\begin{align}
    \mathrm{rawmean}(m)  &= \frac{\sum_E v_E^m \cdot n_E^{\mathrm{eval}, m}}
                                 {\sum_E n_E^{\mathrm{eval}, m}}, \\
    \mathrm{coverage}(m) &= \frac{\sum_E n_E^{\mathrm{eval}, m}}
                                 {\sum_E \big(n_E^{\mathrm{eval}, m} + n_E^{\mathrm{skip}, m} + n_E^{\mathrm{fail}, m}\big)}.
\end{align}
We report $\overline{m}$ in the main results and report $\mathrm{rawmean}(m)$ alongside $\mathrm{coverage}(m)$ in Appendix~\ref{appendix:results:coverage} for transparency. Pillar 1 VBench metrics, which are computed on every shot of every episode without gating, have $\mathrm{coverage}(m) = 1$ by construction and so $\overline{m} = \mathrm{rawmean}(m)$.
\subsubsection{DINOv2-based metrics}
\label{appendix:metrics:pillar3:dino}
 
For each entity $e$ with $|\mathcal{C}(e)| \geq 2$, we compute its appearance centroid in DINOv2 embedding space:
\begin{equation}
    \mathbf{c}_e = \mathop{\mathrm{normalize}}\!\left( \frac{1}{|\mathcal{C}(e)|} \sum_{c \in \mathcal{C}(e)} \phi_{\mathrm{DINO}}(c) \right), \qquad \mathop{\mathrm{normalize}}(\mathbf{v}) = \mathbf{v} / \|\mathbf{v}\|_2.
\end{equation}
The per-appearance similarity to the centroid is
\begin{equation}
    s(c, e) = \phi_{\mathrm{DINO}}(c)^\top \mathbf{c}_e \in [-1, 1] \quad \text{for } c \in \mathcal{C}(e).
\end{equation}
 
\paragraph{Discussion: Why centroid rather than anchor.} An anchor-based metric, comparing each appearance to a designated reference appearance, suffers from two problems. First, it depends on which appearance is chosen as anchor: if the chosen reference is a poor rendering, the entire entity is unfairly penalized as the bad anchor pulls all per-appearance similarities down. Second, in keyframe-then-animate methods, the first appearance is often generated by a different pipeline branch (\textit{e.g.}, T2I) than later appearances (\textit{e.g.}, I2V); pinning the anchor to the first appearance systematically biases the metric. The centroid is the unique reference point invariant to ordering, and an outlier crop only drags the centroid by a factor of $1/N$ rather than dominating the comparison.
 
\paragraph{Episode-level aggregation.} For entity type $\mathcal{T} \in \{\mathrm{char}, \mathrm{obj}\}$, the episode-level metric pools all per-appearance similarities across all entities of that type:
\begin{equation}
    \mathrm{cs\_face}(E) = \mathrm{mean}\!\left( \bigcup_{e \in \mathcal{E}^{\mathrm{char}}} \big\{ s(c, e) : c \in \mathcal{C}(e), \; |\mathcal{C}(e)| \geq 2 \big\} \right),
\end{equation}
and similarly for $\mathrm{cs\_object}$. This pooling means an entity that appears in $N$ shots contributes $N$ samples to the mean. This is reasonable as a character that appears 8 times usually matters more than one that appears 2 times, so is in for episode-level consistency.
 
We additionally record per-entity diagnostics in the audit JSON: the mean, minimum (worst-deviation appearance), maximum (representative appearance), and pairwise median similarities; the shot keys of the worst and most-representative appearances; and the full per-shot breakdown for failure analysis.
 
\paragraph{Cross-Shot transition boundary.} For each continuation pair $(S_k, S_{k+1})$ where $\mathrm{cut}(k+1) = \mathsf{False}$, we compute the boundary similarity
\begin{equation}
    \mathrm{btrans}(k) = \phi_{\mathrm{DINO}}(f_{k, F_k})^\top \phi_{\mathrm{DINO}}(f_{k+1, 1})
\end{equation}
between the last frame of the previous shot and the first frame of the next. The episode-level metric is
\begin{equation}
    \mathrm{cs\_transition\_boundary}(E) = \mathrm{mean}\big(\{\mathrm{btrans}(k) : \mathrm{cut}(k+1) = \mathsf{False}\}\big).
\end{equation}
This measures motion continuity at scene-internal boundaries. Hard scene cuts ($\mathrm{cut}(k+1) = \mathsf{True}$) are excluded since discontinuity at scene boundaries is intentional.
 
\paragraph{Discussion: Why no DINOv2 location metric.} A location bounding box necessarily includes the entire visible scene, including any foreground characters. Two location appearances that share the same background but with different foreground characters present will produce different DINOv2 embeddings, and the metric would penalize this as inconsistency. We therefore evaluate location consistency using only the LLM-based metrics (Section~\ref{appendix:metrics:pillar3:llm:scene}), which can be instructed to better ignore foreground.
 
\subsubsection{LLM-based metrics: characters and objects}
\label{appendix:metrics:pillar3:llm}
 
For each entity $e$ with $|\mathcal{C}(e)| \geq 2$, we select an anchor crop $c_{\mathrm{anchor}}(e) \in \mathcal{C}(e)$ and compare it pairwise against each remaining appearance using $M_{\mathrm{LLM}}$. The anchor is the centroid-representative crop:
\begin{equation}
    c_{\mathrm{anchor}}(e) = \mathop{\mathrm{arg\,max}}_{c \in \mathcal{C}(e)} \; s(c, e),
\end{equation}
i.e., the appearance whose DINOv2 embedding is closest to the entity's centroid. This anchor choice is principled in the same way the centroid metric is: it does not depend on shot order, and it does not systematically bias toward T2V outputs.
 
For each pair $(c_{\mathrm{anchor}}(e), c)$ with $c \in \mathcal{C}(e) \setminus \{c_{\mathrm{anchor}}(e)\}$, we query the LLM with both crops and the entity's textual description:
\begin{equation}
    J_{e,c} = M_{\mathrm{LLM}}\big(\{c_{\mathrm{anchor}}(e), c\}, \pi_{\mathrm{pair}}(\mathrm{desc}(e), \mathcal{T})\big),
\end{equation}
where $\pi_{\mathrm{pair}}$ is the pairwise prompt template parameterized by entity type $\mathcal{T}$. The LLM returns a JSON dictionary $J_{e,c}$ with a binary same/different verdict $\mathrm{same}_{e,c} \in \{0,1\}$, an overall similarity score $\mathrm{sim}_{e,c} \in [0,1]$, and four type-specific per-criterion scores $\mathrm{crit}_{e,c}^j \in [0,1]$ for $j \in \mathcal{J}^{\mathcal{T}}$. The per-type criterion sets $\mathcal{J}^{\mathcal{T}}$ are identical to those used in Pillar 2 (Section~\ref{appendix:metrics:pillar2:fid}), enabling direct comparison.
 
\paragraph{Discussion: Why pairwise rather than set-based.} An alternative is to send all $|\mathcal{C}(e)|$ appearances in a single LLM call and ask the model to identify outliers. We empirically found that when $|\mathcal{C}(e)|$ is large, set-based judging may produce unreliable counts. The model sometimes returns out-of-range or inaccurate indices. Pairwise judging reduces each LLM call to a clean binary decision (``are these two the same?'') which the model handles more consistently.
 
\paragraph{Episode-level aggregation.} Let $\mathcal{P}^{\mathcal{T}}(E)$ denote the multiset of all (anchor, comparison) pairs in episode $E$ for entity type $\mathcal{T}$:
\begin{equation}
    \mathcal{P}^{\mathcal{T}}(E) = \big\{ (e, c) : e \in \mathcal{E}^{\mathcal{T}}, \; |\mathcal{C}(e)| \geq 2, \; c \in \mathcal{C}(e) \setminus \{c_{\mathrm{anchor}}(e)\} \big\}.
\end{equation}
The Pillar 3 LLM metrics for characters are
\begin{align}
    \mathrm{llm\_face\_accuracy}(E)   &= \mathrm{mean}\big( \{\mathrm{same}_{e,c} : (e,c) \in \mathcal{P}^{\mathrm{char}}(E)\} \big), \\
    \mathrm{llm\_face\_mean\_score}(E) &= \mathrm{mean}\big( \{\mathrm{sim}_{e,c} : (e,c) \in \mathcal{P}^{\mathrm{char}}(E)\} \big), \\
    \mathrm{llm\_face\_}j(E)           &= \mathrm{mean}\big( \{\mathrm{crit}_{e,c}^j : (e,c) \in \mathcal{P}^{\mathrm{char}}(E),\; \mathrm{crit}_{e,c}^j \neq \mathsf{None}\} \big),
\end{align}
for each $j \in \{\mathrm{face}, \mathrm{hair}, \mathrm{clothing}, \mathrm{build}\}$. The objects suite (\texttt{llm\_object\_*}) is defined identically with $\mathcal{P}^{\mathrm{obj}}$ and $\mathcal{J}^{\mathrm{obj}}$. This yields 6 metrics per entity type, including overall accuracy, overall mean score, four per-criterion scores, for a total of 12 metrics across characters and objects.
 
\subsubsection{LLM-based metrics: locations}
\label{appendix:metrics:pillar3:llm:scene}

Locations are evaluated differently from characters and objects in two respects. First, location judging uses full frames rather than crops, with a prompt explicitly instructing the LLM to ignore foreground characters and focus on the depicted place. Second, different camera angles, distances, framings, partial views, and zoom levels of the same physical location may look completely different, and uses a chain-of-thought structure that forces the LLM to commit to a per-frame description of the location before making a similarity judgment. 
This per-frame identification step mitigates the failure mode where high cinematographic diversity (close-ups, wide shots, pans) is mistaken for location inconsistency under naive set-based judging.

Following the character pipeline, location judging is anchor-vs-each pairwise. For each location $\ell \in \mathcal{E}^{\mathrm{loc}}$ with $|\mathcal{C}(\ell)| \geq 2$, we select an anchor shot $c^{\star}_\ell \in \mathcal{C}(\ell)$ (the centroid-representative appearance, see Section~\ref{appendix:metrics:pillar3:dino}). For each non-anchor shot $c \in \mathcal{C}(\ell) \setminus \{c^{\star}_\ell\}$, we sample at most $N_{\mathrm{frames\_per\_set}} = 2$ sharpness-ranked full frames from each of the two shots, yielding $\leq 4$ images per
pairwise call. Let $X_\ell^{c}$ denote the resulting image set:
\begin{equation}
    P_\ell^{c} = M_{\mathrm{LLM}}\big(X_\ell^{c},\;
        \pi_{\mathrm{pair}}^{\mathrm{loc}}(\mathrm{desc}(\ell))\big).
\end{equation}
Each pairwise call returns a binary same-location verdict $\mathrm{same}_\ell^{c} \in \{0,1\}$, an overall similarity $\mathrm{sim}_\ell^{c} \in [0,1]$, and four per-criterion scores
$\mathrm{crit}_\ell^{c,j} \in [0,1]$ for $j \in \mathcal{J}^{\mathrm{loc}} = \{\text{layout, color\_mood,
landmarks, perspective}\}$.

\paragraph{Per-location aggregation.}
For each location, we aggregate across the $|\mathcal{C}(\ell)| - 1$
pairwise comparisons:
\begin{align}
    \mathrm{allcons}_\ell &= \prod_{c \neq c^{\star}_\ell} \mathrm{same}_\ell^{c}, \\
    \mathrm{cons}_\ell    &= \mathrm{mean}\big(\{\mathrm{sim}_\ell^{c} : c \neq c^{\star}_\ell\}\big), \\
    \mathrm{crit}_\ell^j  &= \mathrm{mean}\big(\{\mathrm{crit}_\ell^{c,j} : c \neq c^{\star}_\ell\}\big).
\end{align}

\paragraph{Episode-level aggregation.}
\begin{align}
    \mathrm{llm\_scene\_accuracy}(E)    &= \mathrm{mean}\big( \{\mathrm{allcons}_\ell : \ell \in \mathcal{E}^{\mathrm{loc}}, \; |\mathcal{C}(\ell)| \geq 2\} \big), \\
    \mathrm{llm\_scene\_mean\_score}(E) &= \mathrm{mean}\big( \{\mathrm{cons}_\ell : \ell \in \mathcal{E}^{\mathrm{loc}}, \; |\mathcal{C}(\ell)| \geq 2\} \big), \\
    \mathrm{llm\_scene\_}j(E)           &= \mathrm{mean}\big( \{\mathrm{crit}_\ell^j : \ell \in \mathcal{E}^{\mathrm{loc}}, \; |\mathcal{C}(\ell)| \geq 2, \; \mathrm{crit}_\ell^j \neq \mathsf{None}\} \big),
\end{align}
for each $j \in \mathcal{J}^{\mathrm{loc}}$. This yields 6 location metrics, bringing the Pillar 3 LLM total to 18 and the Pillar 3 overall total to 21.
 
\subsubsection{Gap-decay diagnostic}
\label{appendix:metrics:pillar3:gapdecay}
 
In addition to the 21 headline Pillar 3 metrics, we record a per-pair gap-decay dataset for diagnostic plotting. For each entity $e$ with $|\mathcal{C}(e)| \geq 2$ and each ordered pair of distinct appearances $(c_i, c_j) \in \mathcal{C}(e) \times \mathcal{C}(e)$ with $i < j$ in shot order, we record the triple
\begin{equation}
    \big( \mathrm{gap} = |k_j - k_i|, \quad \mathrm{sim} = \phi_{\mathrm{DINO}}(c_i)^\top \phi_{\mathrm{DINO}}(c_j), \quad \mathrm{type} = \mathcal{T}_e \big),
\end{equation}
where $k_i, k_j$ are the shot indices of the two appearances. The dataset enables construction of the gap-vs-similarity curve for each method, which characterizes how identity drifts as recurrence distance increases. A flat curve indicates that consistency is maintained regardless of how far apart two appearances are; a falling curve indicates degradation with distance.

\subsection{Implementation Details}
\label{appendix:metrics:strict}

\paragraph{Metric value type.} Every metric value is a structured tuple
\begin{equation}
    m = (v, n_{\mathrm{eval}}, n_{\mathrm{failed}}, n_{\mathrm{skipped}}),
\end{equation}
where $v \in [0, 1] \cup \{\mathsf{None}\}$ (or appropriate canonical range) is the headline value, $n_{\mathrm{eval}}$ counts the number of items that contributed to the aggregation, $n_{\mathrm{failed}}$ counts items where the underlying computation errored (e.g., LLM call failed), and $n_{\mathrm{skipped}}$ counts items legitimately excluded (e.g., entity appears in only one shot, so cross-shot pairing is undefined).

\paragraph{No silent-zero contract.} A metric with $n_{\mathrm{eval}} = 0$ is recorded as $v = \mathsf{None}$, never as $v = 0$. The contract distinguishes three distinct outcomes:
\begin{itemize}
    \item All items succeeded: $v \in [0, 1]$ with $n_{\mathrm{failed}} = n_{\mathrm{skipped}} = 0$.
    \item Some items failed but the rest were valid: $v \in [0, 1]$ with $n_{\mathrm{failed}} > 0$. The mean is taken only over successful items.
    \item No items contributed: $v = \mathsf{None}$ with $n_{\mathrm{eval}} = 0$. The episode is excluded from the across-episode mean.
\end{itemize}
This avoids the common failure mode where a method that produces unevaluable outputs (e.g., crashed videos) artificially looks ``perfect'' or ``terrible'' because missing values are silently substituted with extreme defaults.

\paragraph{Run manifest.} Each evaluation run produces a manifest JSON that records every model checkpoint with file fingerprints, every library version, every BENCHMARK\_CONFIG value (thresholds, sampling counts, criterion sets, etc.), and the evaluator's git revision. Two runs whose manifests differ in any non-trivial field are flagged as not directly comparable, with the differences listed in machine-readable form. The fields excluded from the comparability check are limited to: \texttt{method\_name}, \texttt{timestamp\_utc}, \texttt{platform}, \texttt{n\_llm\_keys}.

\paragraph{Hyperparameters.} The complete list of fixed hyperparameters in the canonical configuration is given in Table~\ref{tab:hyperparams}.

\begin{table}[ht]
    \centering
    \small
    \caption{Hyperparameters used in the canonical evaluation. All values are recorded in the run manifest and locked across reported numbers.}
    \scalebox{0.85}{
    \begin{tabular}{lll}
        \toprule
        Parameter & Value & Description \\
        \midrule
        $N_{\mathrm{frame}}$ & 5 & frames sampled per shot for grounding \\
        $\tau_{\mathrm{box}}$ & 0.25 & GroundingDINO box confidence threshold \\
        $\tau_{\mathrm{text}}$ & 0.20 & GroundingDINO text alignment threshold \\
        $\tau_{\mathrm{CLIP}}$ & 0.20 & CLIP threshold for \texttt{present} status \\
        $\tau_{\mathrm{fid}}$ & 0.50 & cross-shot fidelity gate threshold \\
        crop padding & 10\% & padding applied around bounding boxes \\
        crop resolution & $224 \times 224$ & input size to encoders \\
        action grid & $2 \times 3$ & frames per action evaluation \\
        $N_{\mathrm{shots}}$ (loc.\ set) & 8 & max shots sampled for set-based location LLM judging \\
        $N_{\mathrm{frames\_per\_shot}}$ (loc.\ set) & 2 & frames per shot for location judging \\
        DINOv2 model & \texttt{facebook/dinov2-base} & visual encoder for embeddings \\
        CLIP model & \texttt{openai/clip-vit-base-patch32} & for text-image matching \\
        Multimodal LLM & \texttt{gemini-2.5-pro} & for all judgment metrics \\
        \bottomrule
    \end{tabular}
    }
    \label{tab:hyperparams}
\end{table}

\section{\sysname: Entity-Aware Context Management}
\label{app:method}

\subsection{Memory Bank Design}
\label{sec:memory_bank}
 
The memory bank stores visual and textual entity references that the video generation model retrieves at each shot.
We explore a baseline for setting up an entity memory bank where we pre-generate all entity references before any video generation begins, so that each entity's visual identity is established once and reused consistently throughout the sequence.
This avoids a failure mode common in autoregressive approaches, where references are extracted from previously generated outputs: distortions in early shots quietly enter the reference pool and compound in later shots.
 
\paragraph{Per-entity references.}
The bank maintains both visual and textual references for each entity.
On the visual side, each entity receives a reference tailored to its type.
For characters, the reference is a segmented portrait showing a single character in isolation with the background removed, labeled with the character's name rendered as text at the bottom of the image.
The labeling provides an explicit name-to-appearance mapping that helps the video backbone bind textual names to visual identities, particularly when multiple characters co-occur in a shot.
For locations, a panoramic image is cropped into angle variants (left, center, right), giving the compositor a choice of camera-angle-aware backgrounds when assembling keyframes.
For objects, a Classification Agent (\S\ref{sec:agents}) first determines whether the object requires a standalone visual reference at all: mobile props such as creatures or vehicles receive segmented portraits, while wearable items and scene fixtures are parts of character or location portraits.
On the textual side, the bank extracts and stores a description of each entity at its first appearance, which can be retrieved when that entity recurs in a later shot.
 
\paragraph{Per-shot keyframes.}
For each shot, the bank also stores a keyframe composite showing the spatial arrangement of all scheduled entities against the location background.
Unlike entity portraits, which are generated once and reused, keyframes are composed per shot from the pre-generated references.
The Layout Agent (\S\ref{sec:agents}) plans each keyframe's composition: character positions on a discrete horizontal grid, camera angle selection, and (for continuation shots) reasoning about how camera panning shifts retained characters and where entering characters appear.
When characters enter or exit mid-shot, the Layout Agent decomposes the shot into multiple keyframes.
A compositor then height-normalizes the character portraits and places them at planned positions alongside any scheduled objects.
 
\paragraph{Consuming the memory bank.}
At generation time, the references for a given shot are assembled as an ordered sequence: per-character labeled portraits first, followed by the keyframe composites.
The video backbone receives this sequence alongside a text prompt that describes the shot's camera direction, entity description, and actions.
For recurring entities whose appearance descriptions do not appear in the current shot's script, the pipeline retrieves stored descriptions from the bank and injects them into the prompt, ensuring the video backbone has appearance guidance for every scheduled entity.
For continuation shots, the last frame of the previous shot is provided to the video backbone as a separate first-frame input for temporal continuity, but is excluded from the memory bank to prevent it from overriding the curated entity references.
 
\subsection{Agent-Based Context Management}
\label{sec:agents}
 
The memory bank requires high-quality, complete content.
Populating it requires a chain of context management decisions: determining what each entity needs as a reference, generating that reference, verifying its quality before it enters the bank, and arranging the bank's contents into per-shot keyframes for the video backbone.
\sysname delegates each of these decisions to a specialized agent, while deterministic operations such as image generation, segmentation, and compositing are handled by tools.
 
\paragraph{Classification Agent.}
Not every entity requires a pre-generated visual reference.
The Classification Agent examines each entity in the story and determines its reference needs based on entity type and role: characters always receive portraits, locations receive panoramic backgrounds, and objects are evaluated individually.
It distinguishes mobile props that need cross-shot visual consistency (creatures, vehicles, artifacts) from wearable items and scene fixtures that are parts of character or location references.
This filtering step keeps the memory bank focused on entities that genuinely require visual anchoring.
 
\paragraph{Portrait Agent.}
For each entity that requires a visual reference, the Portrait Agent manages its generation. It gathers the entity's description, its first-appearance context from the story script, and the story overview to infer the visual
style (e.g., anime, photorealistic).
For characters and objects, it writes a generation prompt for a text-to-image tool, which produces $N$ candidates on a chroma-key background. After a segmentation tool extracts the foreground of each candidate, the Portrait Agent evaluates the segmented results on a composite grid and selects the best one based on segmentation quality, composition, and body proportions.
For locations, it generates a panoramic image and crops it into angle variants (left, center, right) to provide camera-aware backgrounds for keyframe composition.
 
\paragraph{Verification Agent.}
Before a portrait enters the memory bank, the Verification Agent inspects it for failure modes: incorrect or missing characteristic generation, or segmentation-related failures such as missing body regions, transparent clothing, or incompletely removed backgrounds.
If verification fails, it triggers a retry with an alternative background color (e.g., magenta, blue) to improve segmentation contrast, addressing cases where character appearance blends with the original chroma-key color.
 
\paragraph{Layout Agent.}
Once the memory bank contains verified entity references, the Layout Agent translates each shot's narrative action into one or more keyframe layouts.
Given the shot's action text, the entity schedule, and (for continuation shots) the full state of the previous shot, it determines how many keyframes the shot requires, which entities appear in each, their positions, and the camera angle.
For static shots, a single keyframe captures the scene. When the action changes the spatial arrangement mid-shot, the agent produces multiple keyframes that capture the progression: for example, the first keyframe may show two characters in conversation, while the second introduces a third character arriving at a new position.
For continuation shots, the agent simulates physical camera behavior: it reasons about which direction the camera should pan to accommodate the action, shifts retained characters' positions accordingly (e.g., a character previously on the right moves to the left as the camera pans right), and selects the matching angle variant from the location's panoramic crops.
A compositor then realizes each layout by placing height-normalized portraits at the planned positions alongside any scheduled objects.
 
\paragraph{Tools.}
The agents above rely on three tools for execution: a text-to-image generator~\citep{flux2024} that produces candidate portraits from agent-written prompts, a segmentation model~\citep{ravi2024sam} that extracts foreground masks using entity-type-specific point prompt strategies, and a compositor that arranges segmented portraits onto location backgrounds at agent-specified positions.

\section{\textsc{EntityBench}: Data Examples}
\label{app:entitybench_examples}

This section showcases two stories from \textsc{EntityBench}. For each example we present three classes of figure: an \emph{overview} showing the story summary and the entity registry; an \emph{entity-persistence strip} that visualizes which entities recur in which shots; and a \emph{shot timeline} containing the verbatim per-shot \texttt{action\_descriptions} text and the per-shot \texttt{entity\_schedule} chips. 
The two representative examples are chosen.
Section~\ref{app:eb_ex1} is a compact, single-location piece with a small cast and a clear within-scene continuation chain, while Section~\ref{app:eb_ex2} is a multi-location story whose principal character carries the same wardrobe and props across four distinct locations.

\subsection{Example 1: single-location, ten-shot continuation}
\label{app:eb_ex1}

\begin{figure}[t]
\centering
\includegraphics[width=\linewidth]{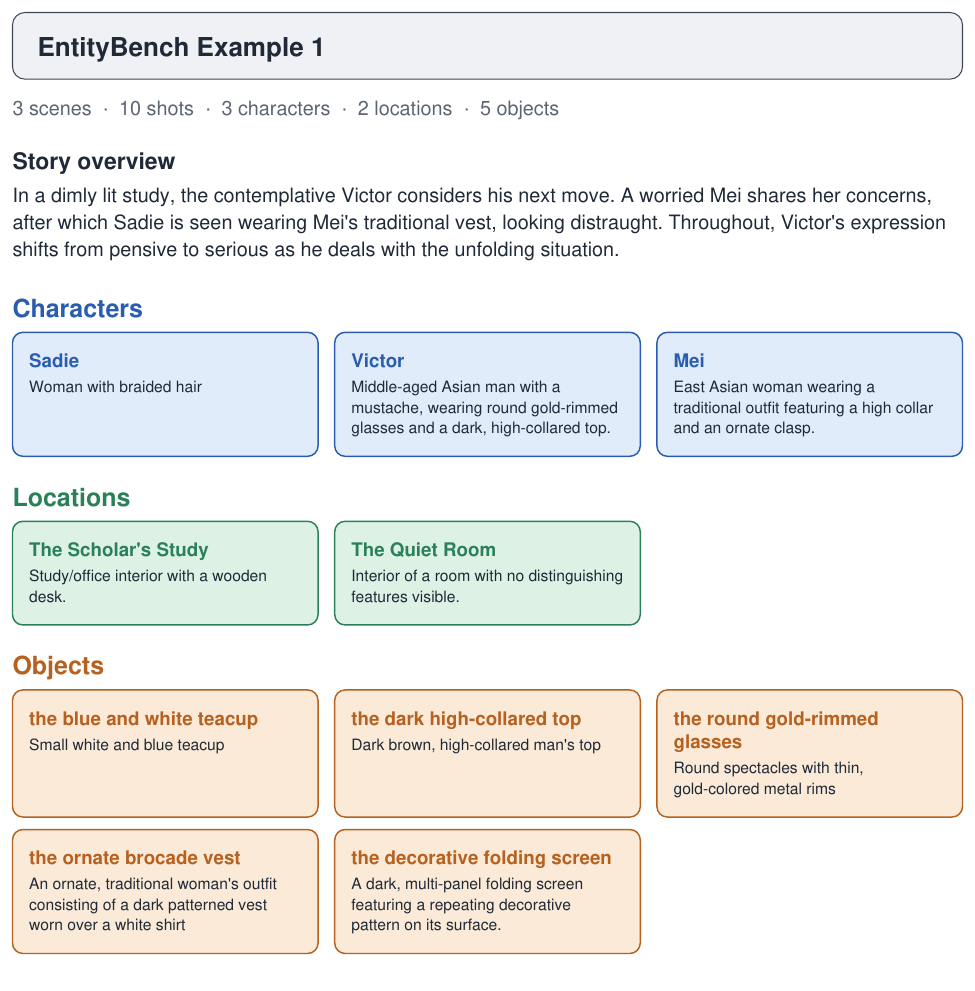}
\caption{\textsc{EntityBench} Example~1: story overview and entity registry. The header reports the structural counts (scenes, shots, characters, locations, objects). The registry below, with chip color indicating entity type, is at the bottom.}
\label{fig:eb_ex1_overview}
\end{figure}

\begin{figure}[t]
\centering
\includegraphics[width=\linewidth]{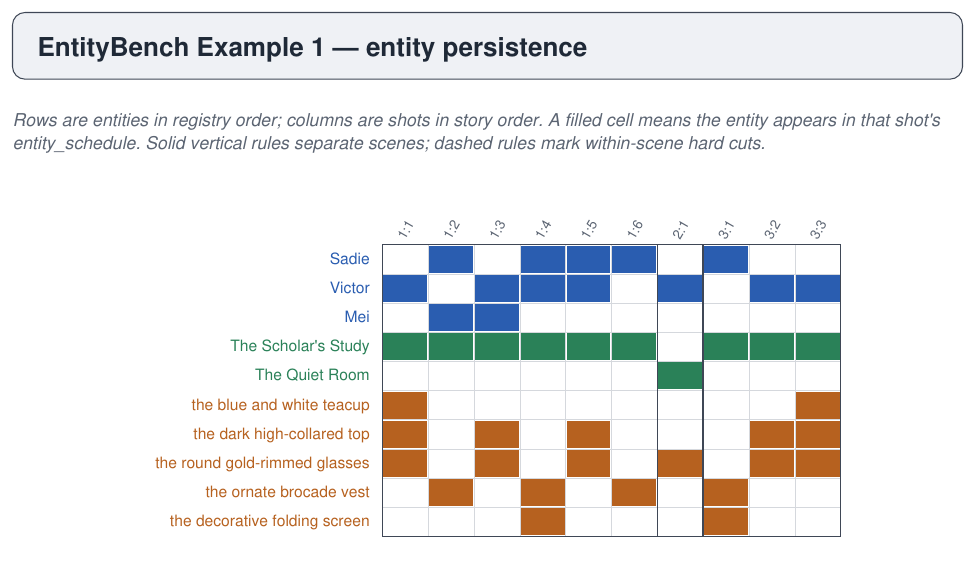}
\caption{\textsc{EntityBench} Example~1: entity-persistence strip.
Rows are entities in registry order and columns are shots in story order. A filled cell means the entity is scheduled in that shot. Solid vertical rules separate scenes; dashed rules mark within-scene hard cuts. }
\label{fig:eb_ex1_strip}
\end{figure}

\begin{figure}[t]
\centering
\includegraphics[width=0.8\linewidth]{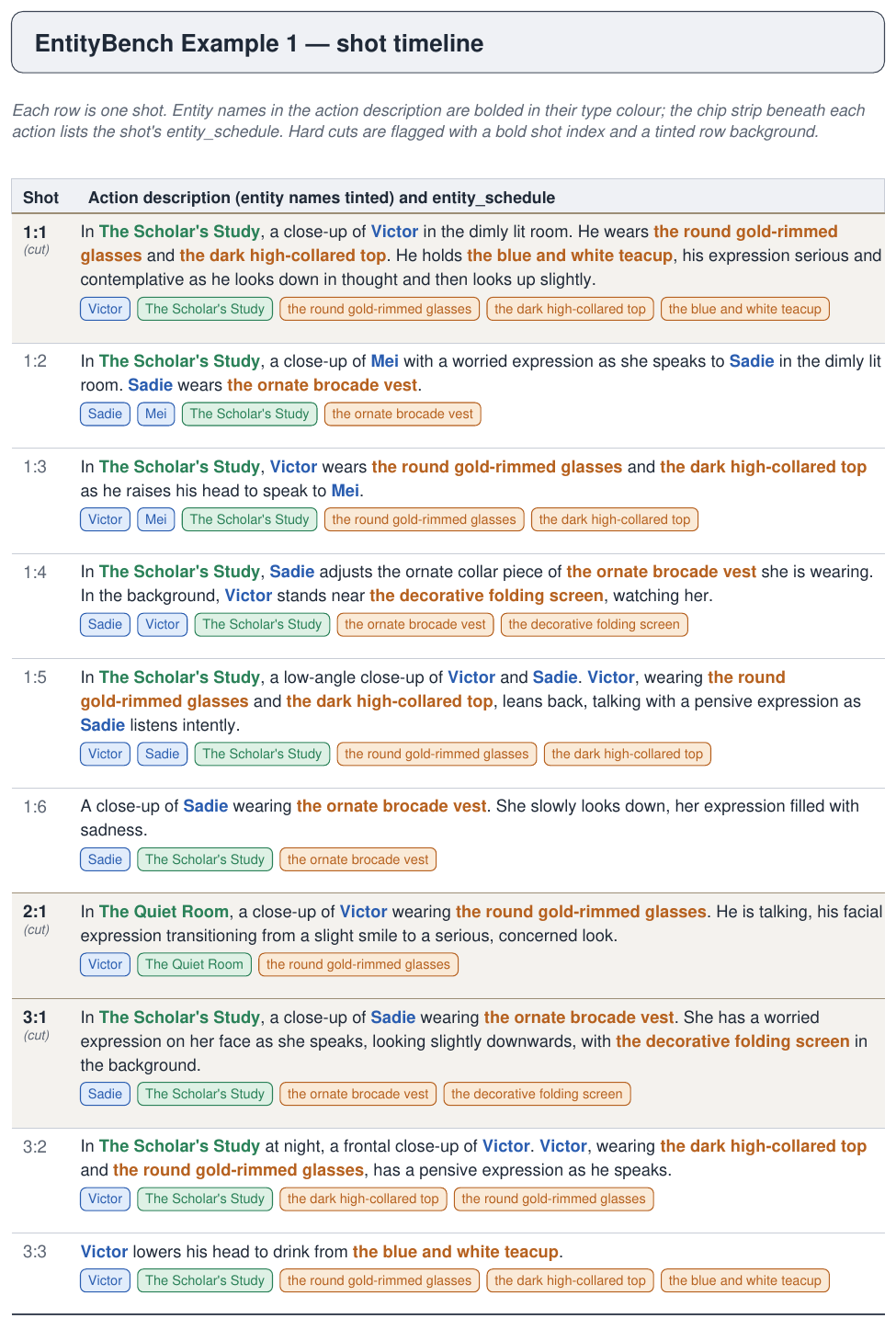}
\caption{\textsc{EntityBench} Example~1: shot timeline. Each row is
one shot. The verbatim \texttt{action\_descriptions} with every entity that the shot's \texttt{entity\_schedule} references, bolded and tinted in its type color. Hard cuts are flagged with bold shot indices and a tinted row background.}
\label{fig:eb_ex1_shots_1}
\end{figure}

The first example is a short three-scene story with three characters, two locations (\emph{The Scholar's Study} and \emph{The Quiet Room}), and five recurring objects. It is built from a single hard-cut opening shot followed by a six-shot continuation chain in \emph{The Scholar's Study}, a one-shot interlude in \emph{The Quiet Room}, and a final three-shot continuation back in the study. 
Fig.~\ref{fig:eb_ex1_overview} demonstrates the entity descriptions.
Fig.~\ref{fig:eb_ex1_strip} visualizes the persistence pattern across all ten shots. Fig.~\ref{fig:eb_ex1_shots_1} shows each shot's action descriptions and entity schedule chips.

\subsection{Example 2: multi-location, cross-scene entity persistence}
\label{app:eb_ex2}

\begin{figure}[t]
\centering
\includegraphics[width=0.8\linewidth]{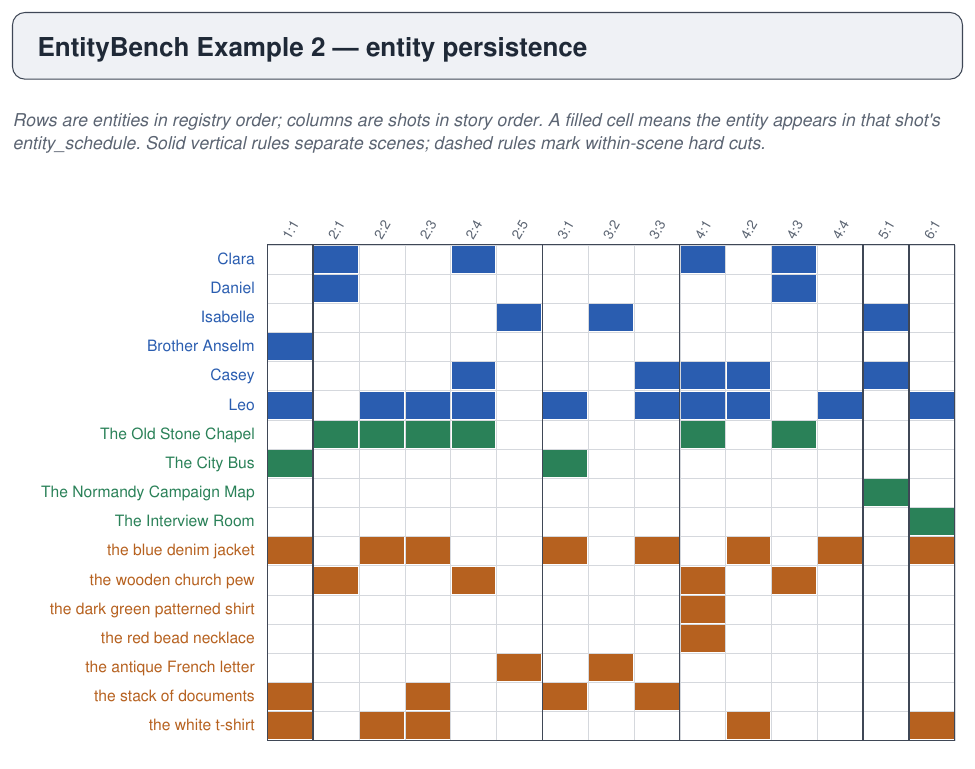}
\caption{\textsc{EntityBench} Example~2: entity-persistence strip.}
\label{fig:eb_ex2_strip}
\end{figure}

\begin{figure}[t]
\centering
\includegraphics[width=0.8\linewidth]{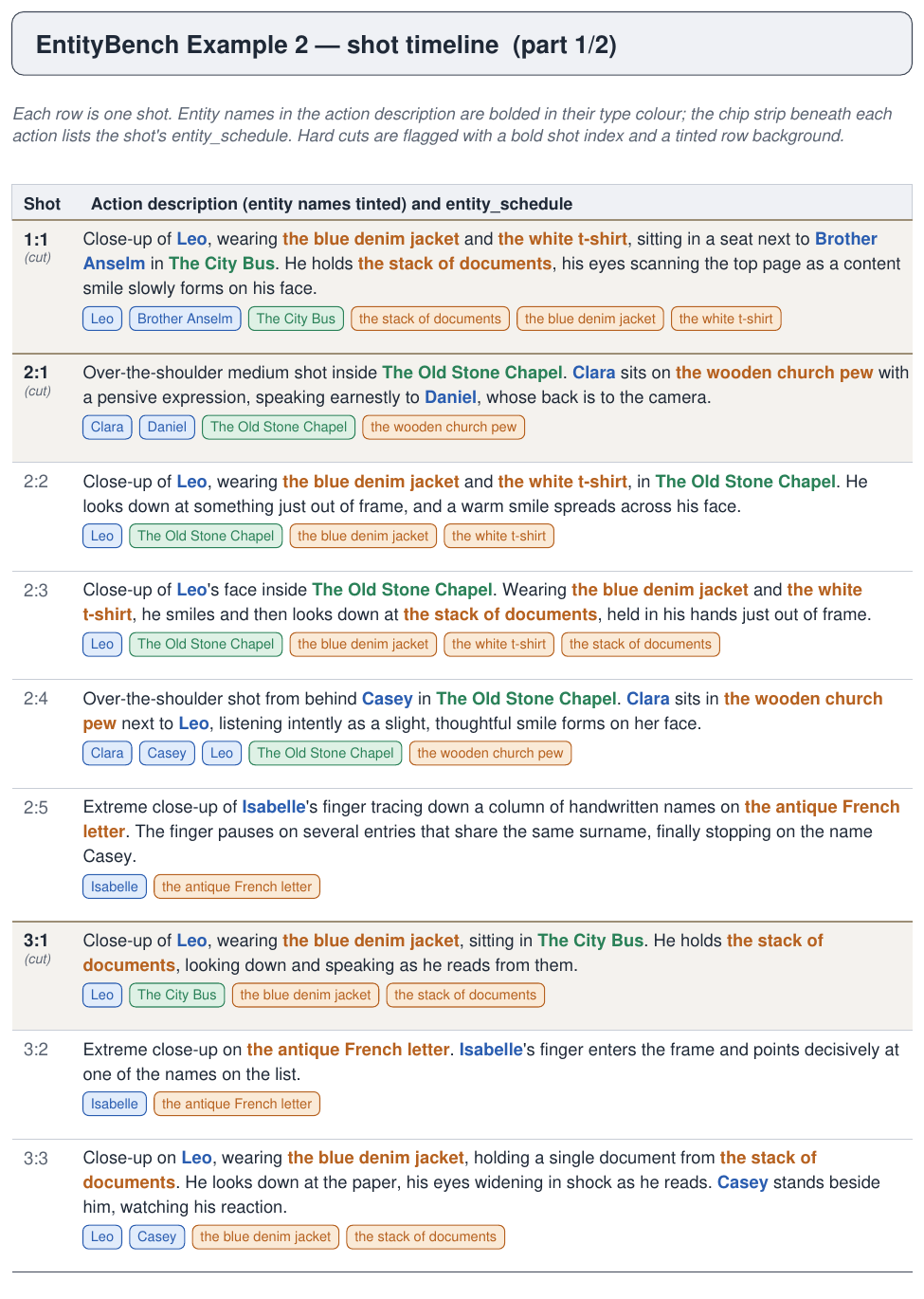}
\caption{\textsc{EntityBench} Example~2: shot timeline, part 1 of 2.
Each row is one shot; the verbatim \texttt{action\_descriptions} text appears with every entity that the shot's \texttt{entity\_schedule} references bolded and tinted in its type color. Hard cuts are flagged with bold shot indices and a tinted row background.}
\label{fig:eb_ex2_shots_1}
\end{figure}

\begin{figure}[t]
\centering
\includegraphics[width=\linewidth]{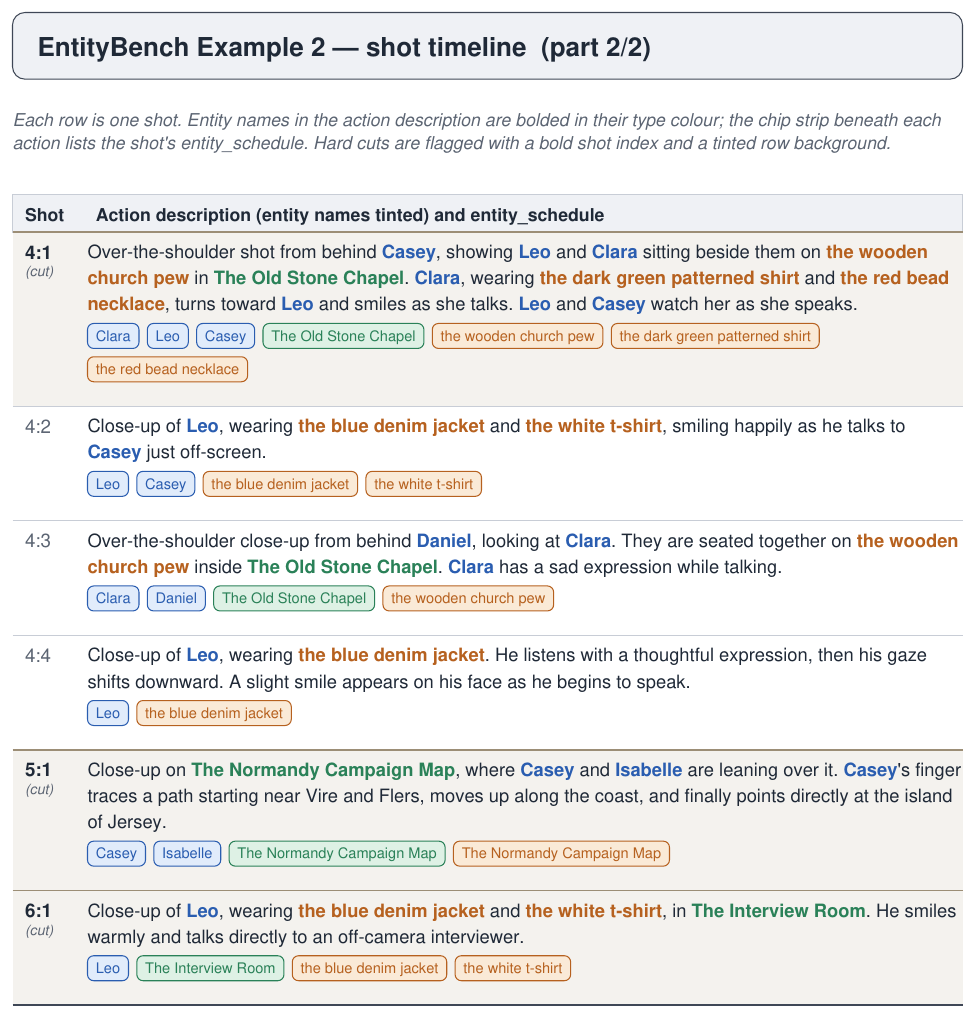}
\caption{\textsc{EntityBench} Example~2: shot timeline, part 2 of 2 (continuation of Fig.~\ref{fig:eb_ex2_shots_1}).}
\label{fig:eb_ex2_shots_2}
\end{figure}

The second example demonstrates the longer-range entity-persistence properties from \textsc{EntityBench}. The story spans six scenes and four locations (\emph{The City Bus}, \emph{The Old Stone Chapel}, \emph{The Normandy Campaign Map}, and \emph{The Interview Room}), and follows a single principal character across them. 
Two wearable objects (\emph{the blue denim jacket} and \emph{the white t-shirt}) recur in nearly every shot the principal appears in, providing a near-continuous wardrobe signal across all four locations. A location-bound prop, \emph{the wooden church pew}, is reused only within \emph{The Old Stone Chapel} (scenes
2 and 4), \emph{the antique French letter}, is referenced only in two non-consecutive shots; the persistence strip in Fig.~\ref{fig:eb_ex2_strip} makes both the dominant wardrobe-and-prop thread and the sparser narrative props visible at a glance. The two shot-timeline figures (Fig.~\ref{fig:eb_ex2_shots_1}--\ref{fig:eb_ex2_shots_2}) show fifteen shots in story order.

\begin{figure}[t]
\centering
\includegraphics[width=0.8\linewidth]{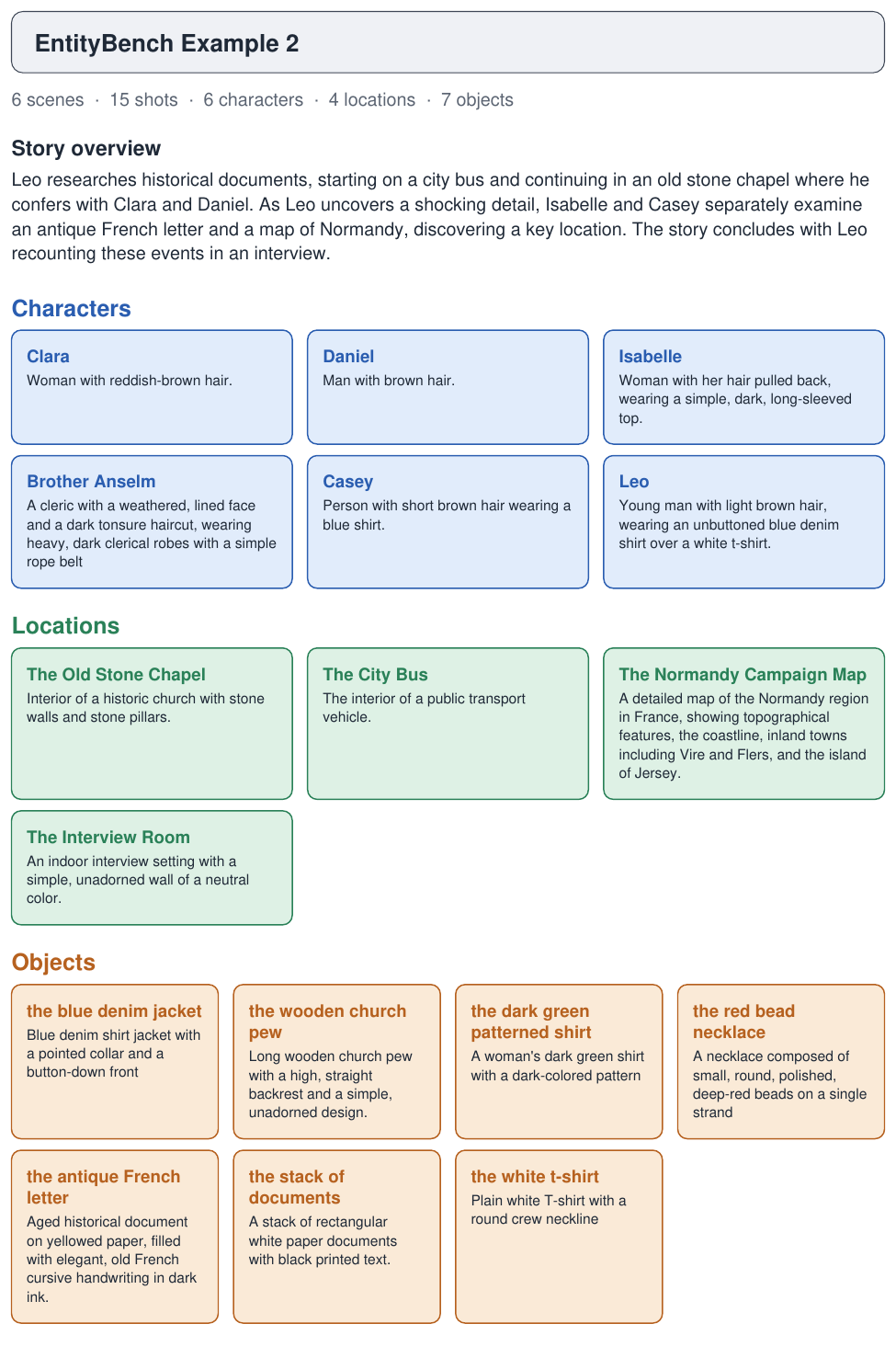}
\caption{\textsc{EntityBench} Example~2: story overview and entity
registry. }
\label{fig:eb_ex2_overview}
\end{figure}

\section{Agent Prompts}
\label{app:prompts}

This section provides the full text of every prompt used by the four \sysname agents, including Classification, Portrait, Verification, and Layout agents. Variables filled in at runtime are typeset in italic blue (e.g. \emph{\{name\}}). All other text is verbatim from our implementation.

\subsection{Classification Agent}
\label{app:prompt_classify}

The Classification Agent decides whether each object entity warrants a pre-generated visual reference (a creature, vehicle, or recurring prop) or should be handled implicitly through the character portrait or location background (a garment, a piece of furniture, a fixture). Characters and locations are unconditionally classified for portrait and panoramic generation, so they bypass this prompt.

\begin{figure}[h]
\centering
\includegraphics[width=\linewidth]{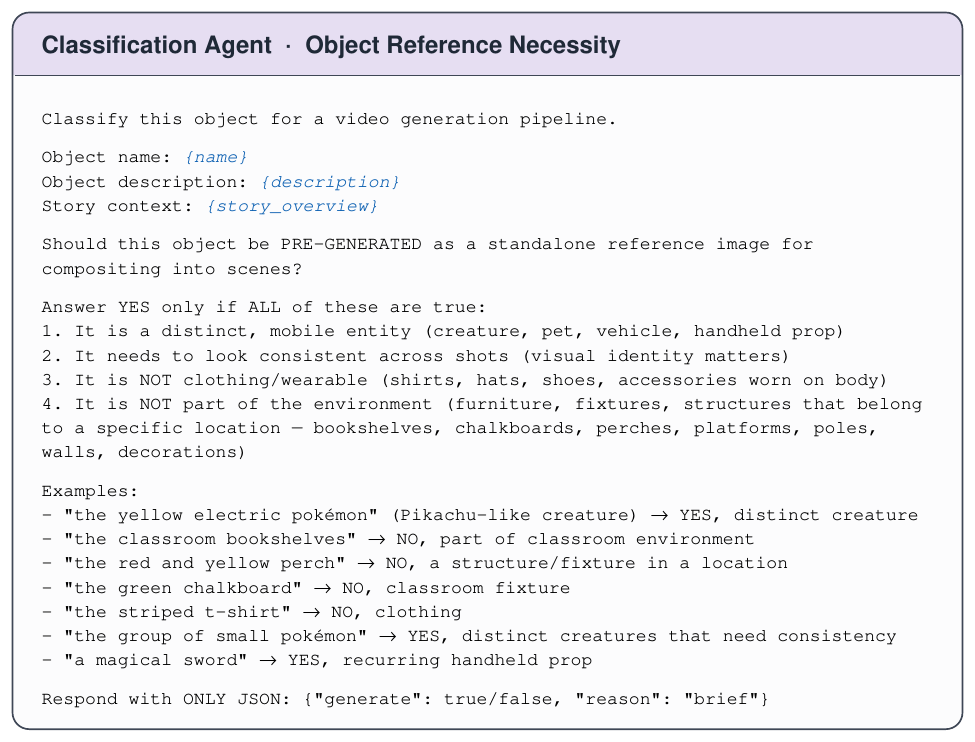}
\caption{Prompt used by the Classification Agent to decide whether an object entity requires a pre-generated standalone reference. The agent receives the object name, description, and a story overview. A negative classification routes the object's appearance into either the owning character's portrait prompt or the location background.}
\label{fig:prompt_classify}
\end{figure}

\subsection{Portrait Agent}
\label{app:prompt_portrait}

The Portrait Agent runs three distinct prompt-writing tasks, \textit{i.e.}, one per entity type, followed by a single multi-image selection call that picks the best candidate after segmentation. Style inference (anime, photorealistic, 3D rendered, etc.) is deferred to the agent in every case.

\begin{figure}[h]
\centering
\includegraphics[width=\linewidth]{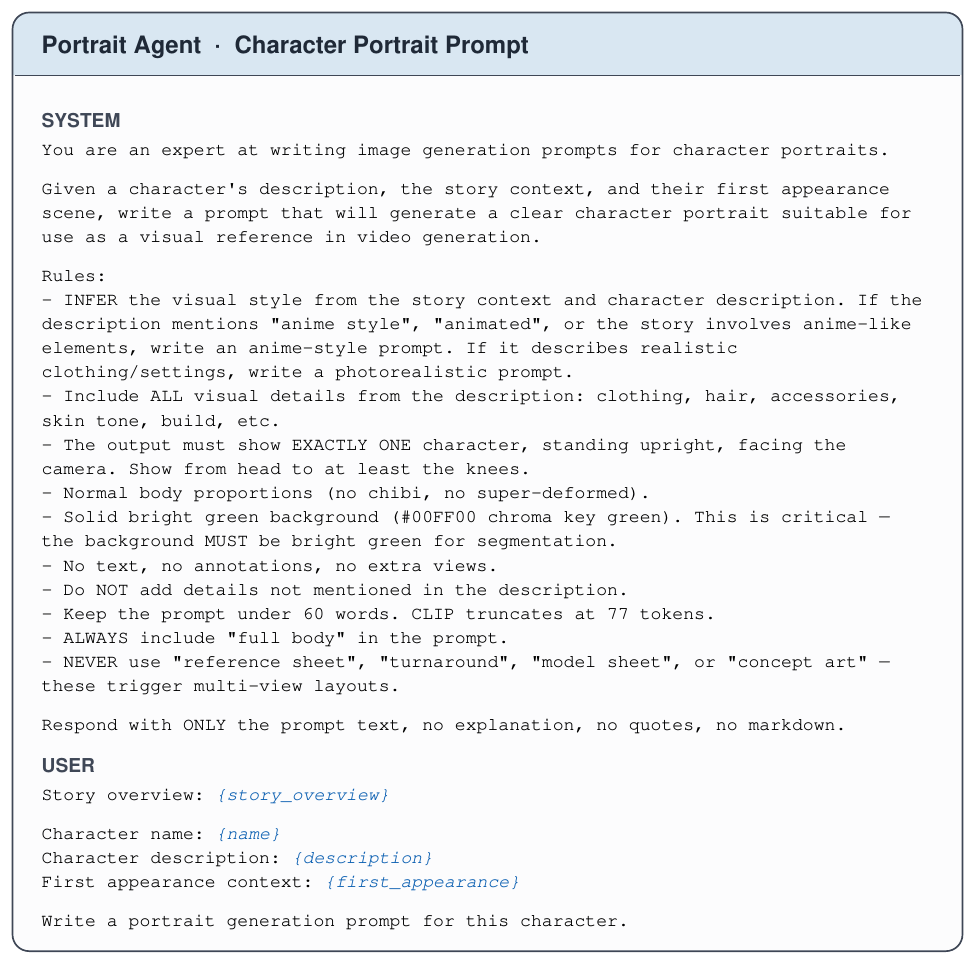}
\caption{Prompt used by the Portrait Agent to write a character-specific prompt. The first-appearance context is the registry line from the shot in which the character is introduced. On a verification failure (Fig.~\ref{fig:prompt_verify}), the chroma-key color in the agent's output is rewritten to magenta, blue, or orange before the next Flux invocation to improve segmentation contrast.}
\label{fig:prompt_portrait_char}
\end{figure}

\paragraph{Character portraits.}
For each character, the agent writes a tailored image generation prompt that preserves visual cues from the registry description while constraining the output to a single view on a chroma-key background suitable for SAM2 segmentation (Fig.~\ref{fig:prompt_portrait_char}).

\begin{figure}[h]
\centering
\includegraphics[width=\linewidth]{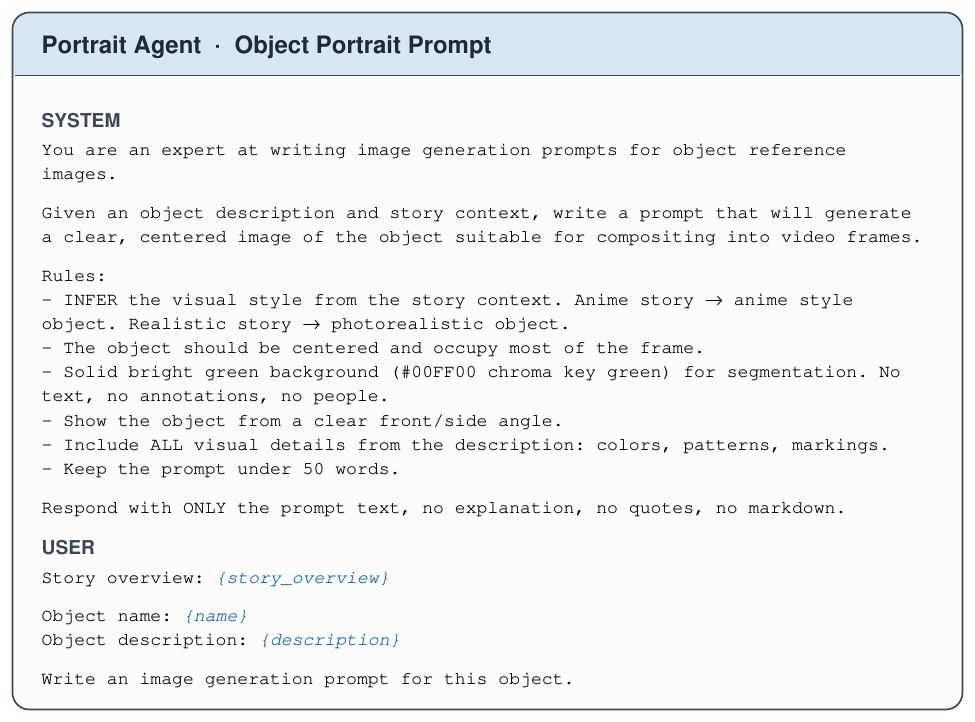}
\caption{Prompt used by the Portrait Agent for objects that the Classification Agent flagged as needing a standalone reference.}
\label{fig:prompt_portrait_obj}
\end{figure}

\paragraph{Object portraits.}
Objects that pass the Classification Agent receive their own square-format prompt, again with style inferred from the story (Fig.~\ref{fig:prompt_portrait_obj}).

\begin{figure}[h]
\centering
\includegraphics[width=\linewidth]{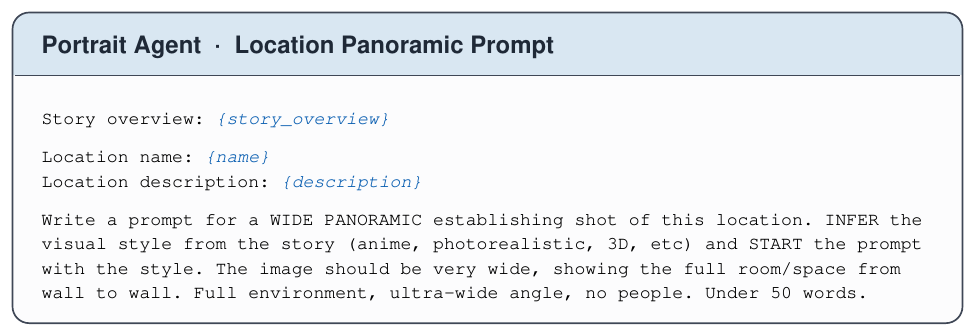}
\caption{Prompt used by the Portrait Agent to write a panoramic-shot image generation prompt for each location. The agent's single prompt drives a $1536\times512$ generation that is then cropped into left/center/right thirds by the compositor.}
\label{fig:prompt_portrait_loc}
\end{figure}

\paragraph{Location panoramas.}
Locations are generated as a single ultra-wide ($1536 \times 512$, 3:1) panorama and then deterministically cropped into left, center, and right variants by the compositor. The agent, therefore, writes one prompt specifying a wide establishing shot for panoramic view generation (Fig.~\ref{fig:prompt_portrait_loc}). This helps address the consistency issues that plague three separately generated angle variants.

\begin{figure}[h]
\centering
\includegraphics[width=\linewidth]{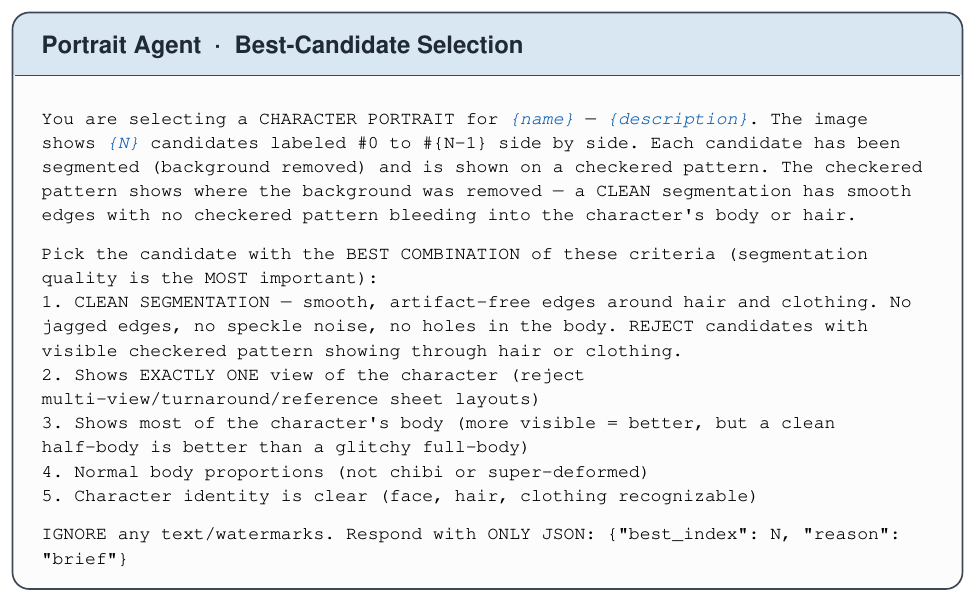}
\caption{Vision-language prompt used by the Portrait Agent to select the best of $N$ segmented candidates from a single side-by-side grid image. A single multi-image call replaces $N$ independent quality calls and compares candidates directly. The same prompt is reused (with ``CHARACTER PORTRAIT'' replaced by the relevant entity type) for object selection.}
\label{fig:prompt_select}
\end{figure}

\paragraph{Best-candidate selection.}
After the image generator~\citep{flux2024} produces $N{=}5$ candidates per entity and SAM2~\citep{ravi2024sam} segments each one, the Portrait Agent calls the LLM~\citep{comanici2025gemini} on a side-by-side grid of the segmented candidates rendered on a checkered background (Fig.~\ref{fig:prompt_select}). The checkered background makes mask artifacts visible, which are otherwise less conspicuous with a solid grey fill.

\subsection{Verification Agent}
\label{app:prompt_verify}

\begin{figure}[h]
\centering
\includegraphics[width=\linewidth]{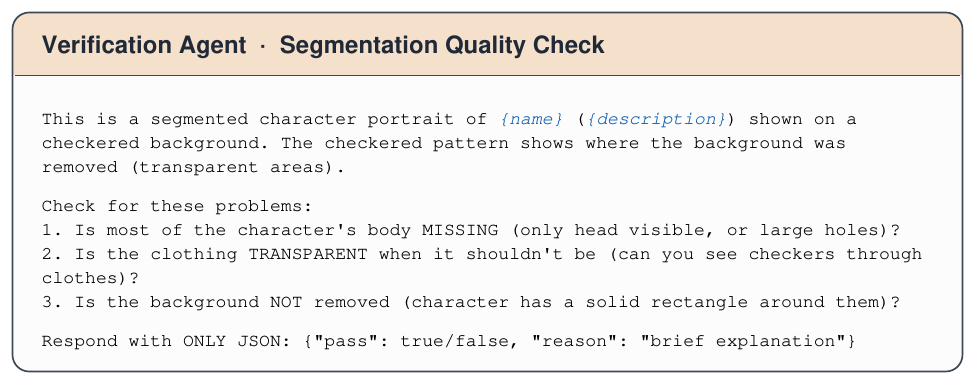}
\caption{Prompt used by the Verification Agent to gate portraits before they enter the memory bank. A failure marks the candidate's chroma-key color as ``contaminated'' and triggers regeneration with the next backup color (magenta $\to$ blue $\to$ orange) up to two retries.}
\label{fig:prompt_verify}
\end{figure}

After selection, the Verification Agent inspects the chosen segmented portrait for the failure modes that defeat downstream compositing: missing body regions, see-through clothing, etc. A failed verification triggers a retry with an alternative chroma-key color, addressing the common case where a part of the foreground matches the original green key (Fig.~\ref{fig:prompt_verify}).

\subsection{Layout Agent}
\label{app:prompt_layout}

\begin{figure}[p]
\centering
\includegraphics[width=\linewidth]{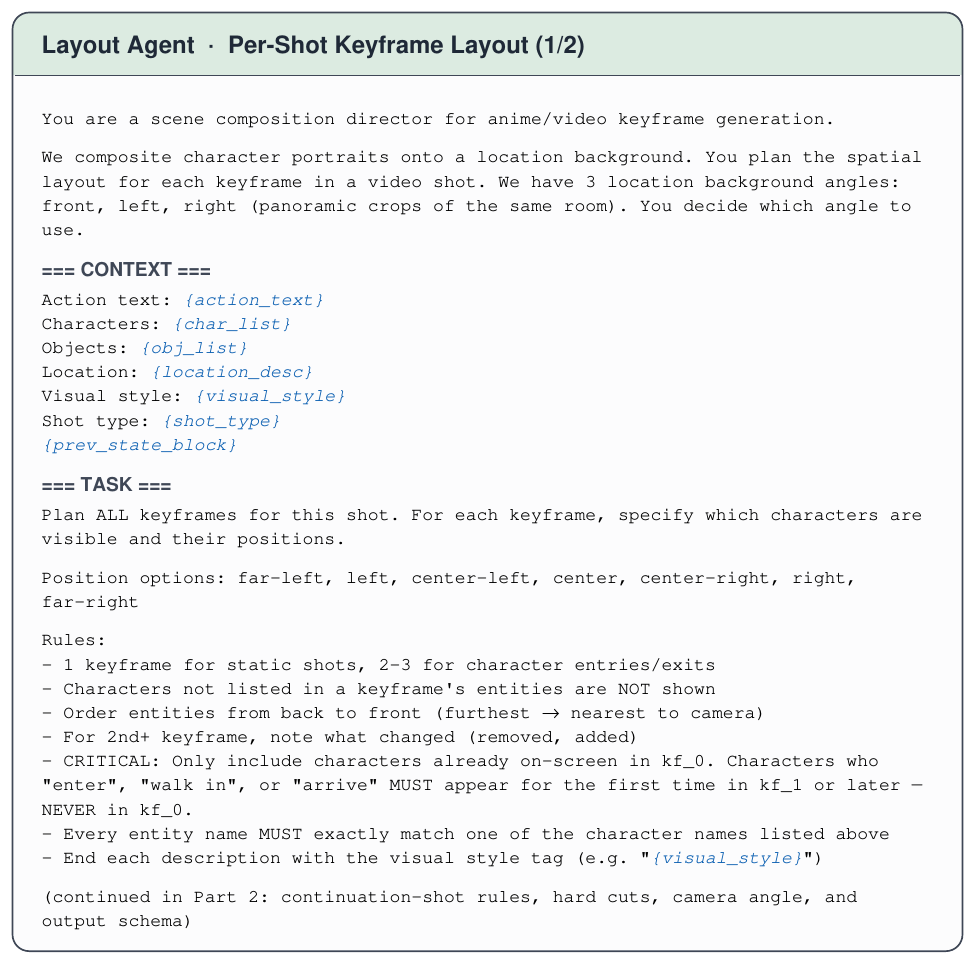}
\caption{Layout Agent prompt, part 1 of 2: input fields and the global task rules. \emph{\{prev\_state\_block\}} is empty for hard-cut shots; for continuations, it contains the previous shot's characters, their positions, camera angle, and last keyframe description, along with explicit lists of retained, entering, and leaving characters. 
The position vocabulary (\textit{far-left} $\to$ \textit{far-right}) matches a 7-cell discrete grid that the compositor maps to pixel coordinates
after height-normalizing the segmented portraits.}
\label{fig:prompt_layout_1}
\end{figure}

\begin{figure}[p]
\centering
\includegraphics[width=\linewidth]{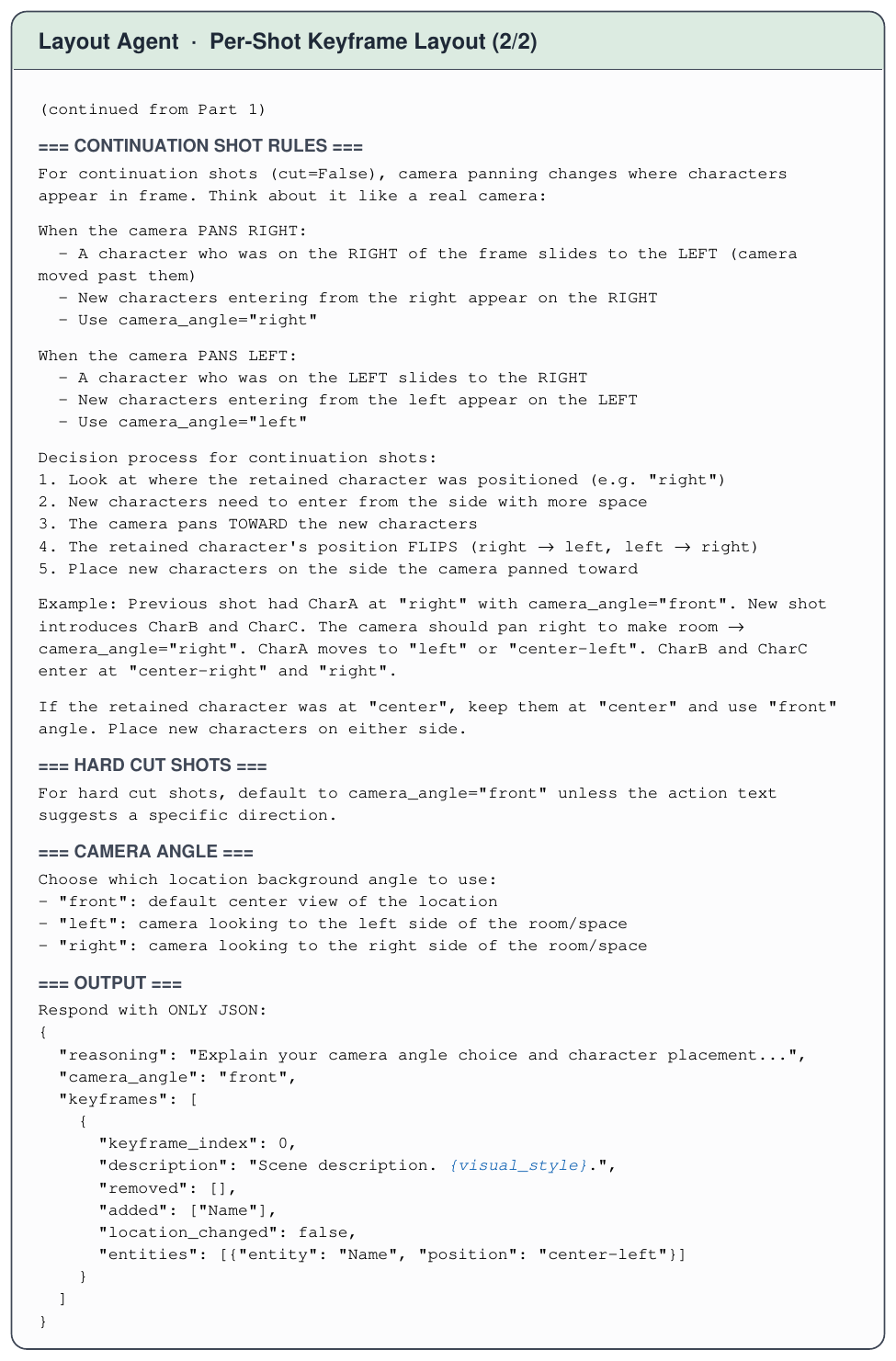}
\caption{Layout Agent prompt, part 2 of 2: continuation-shot reasoning, hard-cut defaults, the camera-angle vocabulary, and the JSON output schema. The agent's \texttt{camera\_angle} choice determines which of the location's three panoramic crops the compositor uses as the background; its \texttt{entities} list names the characters to render in each keyframe and their positions on the 7-cell horizontal grid.}
\label{fig:prompt_layout_2}
\end{figure}

The Layout Agent is context-dependent. For each shot, it receives the action text, the entity schedule, and the previous shot's character positions and camera angle if the shot is a continuation.
It returns a structured plan of one or more keyframes, each with the participating entities, their positions on a discrete 7-cell horizontal grid, and the camera angle (front/left/right) to use as background. The prompt explicitly walks the agent through camera-pan reasoning so that characters retained across a continuation translate the correct way as the camera moves. The prompts are illustrated across two figures:
Fig.~\ref{fig:prompt_layout_1} contains the inputs and the global task rules, and Fig.~\ref{fig:prompt_layout_2} contains the camera-pan reasoning, hard-cut handling, and output schema.

\section{Additional Experimental Results}
\label{appendix:results}

\subsection{\benchname Evaluation: Full 51-metric Results}
\label{appendix:results:full}

This appendix reports the complete \benchname evaluation suite across all 51 metrics for the four methods compared in the main paper. Numbers are fidelity-gate-corrected means following the convention defined in \S\ref{sec:gate} (formal definition in Appendix~\ref{appendix:metrics:pillar3:gate}). Tables~\ref{tab:full_p1}, \ref{tab:full_p2}, and \ref{tab:full_p3} report Pillars 1, 2, and 3 respectively. \textbf{Bold} marks the column winner per row. The 12-metric subset highlighted in main-paper Table~\ref{tab:main} is identified by an asterisk ($^\ast$).

\paragraph{Reading the tables.}
For Pillar~1 (VBench~\citep{huang2024vbench}), \texttt{imaging\_quality} is reported on its native MUSIQ scale of $[0, 100]$; all other Pillar~1 metrics and all metrics in Pillars 2 and 3 are bounded in $[0, 1]$. Pillar~2 organizes per-entity scores by entity type (characters, objects, locations) and includes action correctness as a separate sub-pillar. Pillar~3 organizes cross-shot consistency scores by signal source: DINOv2 embedding similarity, then LLM-judged identity for characters, objects, and scenes (the latter using the camera-invariant pairwise prompt described in Appendix~\ref{appendix:metrics:pillar3:llm:scene}).

\paragraph{Per-method coverage.}
Methods produce evaluable outputs at different rates. Per-metric coverage fractions are reported in Appendix~\ref{appendix:results:coverage}. The fidelity-gate-corrected means in this appendix already incorporate coverage by treating gate-skipped instances as zero contributions; raw means without gate correction are also tabulated in Appendix~\ref{appendix:results:coverage}.

\begin{table}[t]
\centering
\caption{\textbf{Pillar 1: Intra-shot quality} (6 VBench dimensions). $\texttt{imaging\_quality}$ on $[0, 100]$; others on $[0, 1]$.}
\label{tab:full_p1}
\small
\begin{tabular}{lcccc}
\toprule
Metric & Ours & StoryMem & HoloCine & CineTrans \\
\midrule
subject\_consistency$^\ast$  & 0.881 & 0.759 & 0.860 & \textbf{0.968} \\
temporal\_flickering         & 0.976 & 0.838 & 0.957 & \textbf{0.979} \\
motion\_smoothness$^\ast$    & 0.988 & 0.849 & 0.964 & \textbf{0.990} \\
dynamic\_degree              & 0.657 & 0.562 & \textbf{0.721} & 0.688 \\
aesthetic\_quality$^\ast$    & 0.593 & 0.475 & 0.518 & \textbf{0.596} \\
imaging\_quality$^\ast$      & 66.00 & 56.41 & 49.97 & \textbf{68.57} \\
\bottomrule
\end{tabular}
\end{table}

\begin{table}[t]
\centering
\caption{\textbf{Pillar 2: Intra-shot prompt-following alignment} (24 metrics). Per-entity scores aggregate over all (shot, entity) instances passing the fidelity gate.}
\label{tab:full_p2}
\small
\begin{tabular}{lcccc}
\toprule
Metric & Ours & StoryMem & HoloCine & CineTrans \\
\midrule
\multicolumn{5}{l}{\textit{Presence (3 metrics)}} \\
intra\_character\_presence$^\ast$  & \textbf{0.967} & 0.849          & 0.882 & 0.796 \\
intra\_object\_presence$^\ast$     & 0.888          & \textbf{0.893} & 0.723 & 0.776 \\
intra\_location\_presence          & \textbf{0.687} & 0.681          & 0.624 & 0.651 \\
\midrule
\multicolumn{5}{l}{\textit{Character fidelity (5 metrics)}} \\
intra\_face\_fidelity$^\ast$  & \textbf{0.740} & 0.452 & 0.349 & 0.327 \\
intra\_face\_face             & \textbf{0.607} & 0.424 & 0.369 & 0.366 \\
intra\_face\_hair             & \textbf{0.684} & 0.485 & 0.482 & 0.413 \\
intra\_face\_clothing         & \textbf{0.802} & 0.504 & 0.339 & 0.378 \\
intra\_face\_build            & \textbf{0.726} & 0.539 & 0.449 & 0.521 \\
\midrule
\multicolumn{5}{l}{\textit{Object fidelity (5 metrics)}} \\
intra\_object\_fidelity$^\ast$  & 0.601 & \textbf{0.618} & 0.267 & 0.384 \\
intra\_object\_shape            & \textbf{0.712} & 0.701 & 0.373 & 0.508 \\
intra\_object\_color\_texture   & 0.691 & \textbf{0.709} & 0.331 & 0.480 \\
intra\_object\_proportions      & \textbf{0.728} & 0.715 & 0.383 & 0.539 \\
intra\_object\_details          & 0.573 & \textbf{0.598} & 0.256 & 0.371 \\
\midrule
\multicolumn{5}{l}{\textit{Location fidelity (5 metrics)}} \\
intra\_location\_fidelity$^\ast$  & \textbf{0.555} & 0.504 & 0.306 & 0.428 \\
intra\_location\_layout           & \textbf{0.603} & 0.529 & 0.354 & 0.474 \\
intra\_location\_color\_mood      & \textbf{0.706} & 0.627 & 0.474 & 0.588 \\
intra\_location\_landmarks        & \textbf{0.562} & 0.522 & 0.305 & 0.429 \\
intra\_location\_perspective      & \textbf{0.557} & 0.520 & 0.346 & 0.488 \\
\midrule
\multicolumn{5}{l}{\textit{Action correctness (6 metrics)}} \\
intra\_action\_overall$^\ast$         & \textbf{0.618} & 0.547 & 0.569 & 0.273 \\
intra\_action\_depicted               & \textbf{0.519} & 0.446 & 0.458 & 0.124 \\
intra\_action\_subject\_identity      & \textbf{0.706} & 0.595 & 0.606 & 0.478 \\
intra\_action\_subject\_action        & \textbf{0.697} & 0.626 & 0.695 & 0.323 \\
intra\_action\_object\_interaction    & \textbf{0.781} & 0.712 & 0.616 & 0.346 \\
intra\_action\_motion\_quality        & 0.716          & 0.723 & \textbf{0.772} & 0.528 \\
\bottomrule
\end{tabular}
\end{table}

\begin{table}[t]
\centering
\caption{\textbf{Pillar 3: Cross-shot consistency} (21 metrics). DINOv2 metrics use centroid-anchor cosine similarity; LLM metrics use anchor-vs-each pairwise judgment with type-specific criteria. Scenes use the camera-invariant pairwise prompt (Appendix~\ref{appendix:metrics:pillar3:llm:scene}).}
\label{tab:full_p3}
\small
\begin{tabular}{lcccc}
\toprule
Metric & Ours & StoryMem & HoloCine & CineTrans \\
\midrule
\multicolumn{5}{l}{\textit{DINOv2 embedding similarity (3 metrics)}} \\
cs\_face$^\ast$                & 0.737 & \textbf{0.792} & 0.751 & 0.772 \\
cs\_object$^\ast$              & 0.798 & \textbf{0.839} & 0.803 & 0.794 \\
cs\_transition\_boundary$^\ast$  & \textbf{0.738} & 0.663 & 0.498 & 0.508 \\
\midrule
\multicolumn{5}{l}{\textit{LLM characters (6 metrics)}} \\
llm\_face\_accuracy$^\ast$    & \textbf{0.406} & 0.226 & 0.228 & 0.091 \\
llm\_face\_mean\_score$^\ast$ & \textbf{0.426} & 0.234 & 0.242 & 0.145 \\
llm\_face\_face               & \textbf{0.381} & 0.216 & 0.223 & 0.145 \\
llm\_face\_hair               & \textbf{0.447} & 0.248 & 0.282 & 0.175 \\
llm\_face\_clothing           & \textbf{0.464} & 0.241 & 0.242 & 0.143 \\
llm\_face\_build              & \textbf{0.489} & 0.260 & 0.285 & 0.217 \\
\midrule
\multicolumn{5}{l}{\textit{LLM objects (6 metrics)}} \\
llm\_object\_accuracy$^\ast$       & 0.164 & \textbf{0.203} & 0.088 & 0.092 \\
llm\_object\_mean\_score$^\ast$    & 0.202 & \textbf{0.222} & 0.094 & 0.145 \\
llm\_object\_shape                 & 0.232 & \textbf{0.239} & 0.104 & 0.180 \\
llm\_object\_color\_texture        & 0.235 & \textbf{0.243} & 0.104 & 0.190 \\
llm\_object\_proportions           & 0.238 & \textbf{0.244} & 0.105 & 0.195 \\
llm\_object\_details               & 0.184 & \textbf{0.209} & 0.087 & 0.124 \\
\midrule
\multicolumn{5}{l}{\textit{LLM scenes — camera-invariant pairwise (6 metrics)}} \\
llm\_scene\_accuracy        & 0.309 & \textbf{0.398} & 0.304 & 0.119 \\
llm\_scene\_mean\_score$^\ast$  & 0.659 & \textbf{0.671} & 0.616 & 0.432 \\
llm\_scene\_layout          & \textbf{0.697} & 0.684 & 0.641 & 0.449 \\
llm\_scene\_color\_mood     & 0.716 & \textbf{0.724} & 0.669 & 0.619 \\
llm\_scene\_landmarks       & 0.603 & \textbf{0.637} & 0.563 & 0.346 \\
llm\_scene\_perspective     & \textbf{0.727} & 0.696 & 0.713 & 0.467 \\
\bottomrule
\end{tabular}
\end{table}

\subsection{Per-method Coverage and Raw Means}
\label{appendix:results:coverage}

The fidelity-gate-corrected means in the main paper (Table~\ref{tab:main}) and Appendix~\ref{appendix:results:full} aggregate as $\overline{m} = \mathrm{rawmean}(m) \times \mathrm{coverage}(m)$, where coverage is the fraction of eligible (shot, entity) instances that pass the fidelity gate (Equation~\ref{eq:fidelity-gate}). This appendix decomposes the corrected means into their two components for transparency.

\paragraph{What coverage measures.}
For each per-entity metric, coverage answers a different question:
For Pillar~2 fidelity (intra-shot) and Pillar~3 DINOv2, the fidelity gate is applied at the embedding-similarity level rather than as a hard rejection, so all eligible instances enter the pool with $\mathrm{coverage} = 1$ for these metrics.
However, for Pillar~3 cross-shot LLM, coverage is the fraction of (anchor, comparison) pairs where the gate admits both appearances and the LLM call completes successfully. Low coverage indicates a method whose entity renderings often fail intra-shot fidelity, leaving few admissible appearances for cross-shot comparison.
A method's coverage on Pillar~3 LLM metrics thus principally reflects \emph{intra-shot rendering fidelity}, because a method that fails the gate frequently has fewer pairs available for cross-shot judgment, and the corrected mean correctly penalizes this because gate failure is a method failure, not an evaluation artifact.

\begin{table}[t]
\centering
\caption{\textbf{Per-method raw means and coverage fractions} for representative per-entity metrics. Each cell shows raw mean / coverage. }
\label{tab:coverage_raw}
\small
\begin{tabular}{lcccc}
\toprule
Metric & Ours & StoryMem & HoloCine & CineTrans \\
\midrule
\multicolumn{5}{l}{\textit{Pillar 2: Intra-shot fidelity (coverage = 1.00 by design)}} \\
intra\_face\_fidelity     & \textbf{0.740 / 1.00} & 0.452 / 1.00 & 0.349 / 1.00 & 0.327 / 1.00 \\
intra\_object\_fidelity   & 0.601 / 1.00 & \textbf{0.618 / 1.00} & 0.267 / 1.00 & 0.384 / 1.00 \\
intra\_location\_fidelity & \textbf{0.555 / 1.00} & 0.504 / 1.00 & 0.306 / 1.00 & 0.428 / 1.00 \\
\midrule
\multicolumn{5}{l}{\textit{Pillar 3: Cross-shot DINOv2 (coverage $\approx$ 1.00; transition is gated)}} \\
cs\_face                  & 0.737 / 1.00 & \textbf{0.792 / 1.00} & 0.751 / 1.00 & 0.771 / 1.00 \\
cs\_object                & 0.798 / 1.00 & \textbf{0.839 / 1.00} & 0.802 / 1.00 & 0.794 / 1.00 \\
cs\_transition\_boundary  & \textbf{0.738 / 1.00} & 0.795 / 0.83 & 0.509 / 0.98 & 0.508 / 1.00 \\
\midrule
\multicolumn{5}{l}{\textit{Pillar 3: Cross-shot LLM (gate-corrected)}} \\
llm\_face\_accuracy   & \textbf{0.678 / 0.60} & 0.718 / 0.31 & 0.536 / 0.43 & 0.266 / 0.34 \\
llm\_object\_accuracy & 0.522 / 0.31 & \textbf{0.699 / 0.29} & 0.592 / 0.15 & 0.296 / 0.31 \\
\bottomrule
\end{tabular}
\end{table}

\paragraph{Why \sysname wins the corrected LLM metrics despite a slightly lower raw score.}
The most informative entries in Table~\ref{tab:coverage_raw} are the Pillar~3 LLM rows. On \texttt{llm\_face\_accuracy}, StoryMem's raw mean (0.718) is slightly higher than \sysname's (0.678). It indicates when StoryMem manages to produce two gate-passing appearances of the same character, the LLM judges them roughly correctly. But StoryMem's coverage is only 0.31, vs. 0.60 for \sysname: nearly half as many appearances pass the gate, leaving correspondingly fewer pairs to evaluate. The fidelity-gate-corrected means are therefore $0.678 \times 0.60 = 0.407$ for \sysname vs. $0.718 \times 0.31 = 0.222$ for StoryMem---a 1.83$\times$ advantage that derives entirely from \sysname's better intra-shot rendering rate. The same pattern, slightly muted, applies to \texttt{llm\_object\_accuracy}: StoryMem's raw mean is higher (0.699 vs.\ 0.522), but neither method has high coverage on objects (0.29 vs.\ 0.31), so the corrected means are closer (0.203 vs.\ 0.162).
This decomposition validates the fidelity-gate-corrected aggregation as the appropriate metric for evaluating cross-shot generators. 

\paragraph{Coverage on cs\_transition\_boundary.}
The \texttt{cs\_transition\_boundary} metric measures continuity at scene-internal cuts and is computed only for shot pairs with detectable matched content. Coverage is near-1 for all methods except StoryMem (0.83), which fails to produce detectable continuity content on roughly 17\% of in-scene boundaries; the corrected mean penalizes this gap, dropping StoryMem's raw 0.795 to a corrected 0.660 and reversing the ranking against \sysname.

\paragraph{Pillar~1 and action metrics.}
Pillar~1 VBench dimensions and Pillar~2 action metrics are computed on every shot of every episode without an admission gate, so coverage is uniformly 1.00 and raw equals corrected. We omit those rows from this appendix; their values appear in Tables~\ref{tab:full_p1} and \ref{tab:full_p2}.

\subsection{Per-metric Effect Sizes (\sysname vs.\ StoryMem)}
\label{appendix:results:cohens_d_full}

This appendix reports Cohen's $d$ for each of the 51 metrics in the head-to-head between \sysname and its backbone StoryMem. $d$ is reported with pooled-variance (the more common Cohen's $d$, used for between-groups comparison) and as $d_z$ (the paired-samples variant, $d_z = \mathrm{mean}(\Delta) / \mathrm{sd}(\Delta)$, more appropriate when the same episodes are evaluated under both methods). Both are reported because their values diverge slightly under our pairing structure. Per the convention in Table~\ref{tab:cohens_d}, we lead with pooled $d$ in the main paper. Positive values are where \sysname contributes.
$\Delta$ is the raw mean difference (\sysname minus StoryMem) computed on episodes where both methods produced an evaluable score. $n_{\mathrm{paired}}$ is the number of such episodes.

\begin{table}[h]
\centering
\caption{\textbf{Pillar 1: Intra-shot quality.} VBench dimensions are computed on every shot of every episode, so $n_{\mathrm{paired}} = 140$ for all rows. \texttt{imaging\_quality} $\Delta$ and $d$ are computed on the native MUSIQ scale of $[0, 100]$.}
\label{tab:cohens_d_p1}
\small
\begin{tabular}{lrrrr}
\toprule
Metric & $\Delta$ & $d$ & $d_z$ & $n_{\mathrm{paired}}$ \\
\midrule
subject\_consistency  & $+0.122$ & $+1.18$ & $+1.13$ & 140 \\
temporal\_flickering  & $+0.137$ & $+1.27$ & $+1.21$ & 140 \\
motion\_smoothness    & $+0.139$ & $+1.30$ & $+1.24$ & 140 \\
dynamic\_degree       & $+0.095$ & $+0.43$ & $+0.39$ & 140 \\
aesthetic\_quality    & $+0.118$ & $+1.04$ & $+0.97$ & 140 \\
imaging\_quality      & $+9.59$  & $+1.21$ & $+1.13$ & 140 \\
\bottomrule
\end{tabular}
\end{table}

\begin{table}[h]
\centering
\caption{\textbf{Pillar 2: Intra-shot prompt-following alignment.} Effect sizes computed at the episode level. Per-entity sub-metrics inherit the fidelity-gate convention from Pillar 2's overall metric.}
\label{tab:cohens_d_p2}
\small
\begin{tabular}{lrrrr}
\toprule
Metric & $\Delta$ & $d$ & $d_z$ & $n_{\mathrm{paired}}$ \\
\midrule
\multicolumn{5}{l}{\textit{Presence}} \\
intra\_character\_presence  & $+0.097$ & $+1.23$ & $+0.97$ & 139 \\
intra\_object\_presence     & $-0.021$ & $-0.24$ & $-0.18$ & 138 \\
intra\_location\_presence   & $-0.012$ & $-0.05$ & $-0.04$ & 139 \\
\midrule
\multicolumn{5}{l}{\textit{Character fidelity}} \\
intra\_face\_fidelity   & $\mathbf{+0.262}$ & $\mathbf{+2.33}$ & $+2.06$ & 139 \\
intra\_face\_face       & $+0.183$ & $+1.66$ & $+1.49$ & 139 \\
intra\_face\_hair       & $+0.199$ & $+1.81$ & $+1.62$ & 139 \\
intra\_face\_clothing   & $+0.298$ & $+1.94$ & $+1.74$ & 139 \\
intra\_face\_build      & $+0.187$ & $+1.51$ & $+1.36$ & 139 \\
\midrule
\multicolumn{5}{l}{\textit{Object fidelity}} \\
intra\_object\_fidelity        & $-0.053$ & $-0.41$ & $-0.37$ & 138 \\
intra\_object\_shape           & $+0.012$ & $+0.10$ & $+0.09$ & 138 \\
intra\_object\_color\_texture  & $-0.018$ & $-0.13$ & $-0.12$ & 138 \\
intra\_object\_proportions     & $+0.013$ & $+0.09$ & $+0.08$ & 138 \\
intra\_object\_details         & $-0.025$ & $-0.20$ & $-0.18$ & 138 \\
\midrule
\multicolumn{5}{l}{\textit{Location fidelity}} \\
intra\_location\_fidelity      & $+0.017$ & $+0.11$ & $+0.10$ & 140 \\
intra\_location\_layout        & $+0.074$ & $+0.49$ & $+0.45$ & 140 \\
intra\_location\_color\_mood   & $+0.078$ & $+0.65$ & $+0.59$ & 140 \\
intra\_location\_landmarks     & $+0.039$ & $+0.27$ & $+0.24$ & 140 \\
intra\_location\_perspective   & $+0.037$ & $+0.31$ & $+0.28$ & 140 \\
\midrule
\multicolumn{5}{l}{\textit{Action correctness}} \\
intra\_action\_overall              & $+0.043$ & $+0.33$ & $+0.30$ & 139 \\
intra\_action\_depicted             & $+0.074$ & $+0.42$ & $+0.38$ & 139 \\
intra\_action\_subject\_identity    & $+0.110$ & $+0.71$ & $+0.64$ & 139 \\
intra\_action\_subject\_action      & $+0.040$ & $+0.33$ & $+0.30$ & 139 \\
intra\_action\_object\_interaction  & $+0.069$ & $+0.50$ & $+0.45$ & 138 \\
intra\_action\_motion\_quality      & $-0.008$ & $-0.07$ & $-0.06$ & 139 \\
\bottomrule
\end{tabular}
\end{table}

\begin{table}[h]
\centering
\caption{\textbf{Pillar 3: Cross-shot consistency.} Effect sizes computed at the episode level on the paired subset. }
\label{tab:cohens_d_p3}
\small
\begin{tabular}{lrrrr}
\toprule
Metric & $\Delta$ & $d$ & $d_z$ & $n_{\mathrm{paired}}$ \\
\midrule
\multicolumn{5}{l}{\textit{DINOv2 embedding similarity}} \\
cs\_face                  & $-0.043$ & $-0.66$ & $-0.62$ & 129 \\
cs\_object                & $-0.031$ & $-0.43$ & $-0.39$ & 121 \\
cs\_transition\_boundary  & $-0.052$ & $-0.40$ & $-0.36$ & 130 \\
\midrule
\multicolumn{5}{l}{\textit{LLM characters}} \\
llm\_face\_accuracy    & $-0.026$ & $-0.11$ & $-0.10$ & 129 \\
llm\_face\_mean\_score & $-0.017$ & $-0.10$ & $-0.09$ & 129 \\
llm\_face\_face        & $-0.037$ & $-0.19$ & $-0.17$ & 129 \\
llm\_face\_hair        & $-0.014$ & $-0.08$ & $-0.07$ & 129 \\
llm\_face\_clothing    & $+0.006$ & $+0.03$ & $+0.03$ & 129 \\
llm\_face\_build       & $-0.001$ & $-0.00$ & $-0.00$ & 129 \\
\midrule
\multicolumn{5}{l}{\textit{LLM objects}} \\
llm\_object\_accuracy       & $-0.208$ & $-0.68$ & $-0.62$ & 121 \\
llm\_object\_mean\_score    & $-0.142$ & $-0.64$ & $-0.58$ & 121 \\
llm\_object\_shape          & $-0.112$ & $-0.53$ & $-0.48$ & 121 \\
llm\_object\_color\_texture & $-0.100$ & $-0.53$ & $-0.48$ & 121 \\
llm\_object\_proportions    & $-0.112$ & $-0.57$ & $-0.51$ & 121 \\
llm\_object\_details        & $-0.152$ & $-0.63$ & $-0.56$ & 121 \\
\midrule
\multicolumn{5}{l}{\textit{LLM scenes (camera-invariant pairwise)}} \\
llm\_scene\_accuracy     & $-0.097$ & $-0.29$ & $-0.26$ & 140 \\
llm\_scene\_mean\_score  & $-0.032$ & $-0.16$ & $-0.13$ & 140 \\
llm\_scene\_layout       & $+0.006$ & $+0.03$ & $+0.02$ & 140 \\
llm\_scene\_color\_mood  & $-0.031$ & $-0.17$ & $-0.13$ & 140 \\
llm\_scene\_landmarks    & $-0.052$ & $-0.24$ & $-0.20$ & 140 \\
llm\_scene\_perspective  & $+0.013$ & $+0.07$ & $+0.05$ & 140 \\
\bottomrule
\end{tabular}
\end{table}

\paragraph{Largest effects.}
The largest single-metric advantage for \sysname is \texttt{intra\_face\_fidelity} at $d = +2.33$ ($\Delta = +0.262$, $n = 139$). Five additional metrics exceed $d > +1.0$, all in character-related categories: \texttt{intra\_face\_clothing} ($d = +1.94$), \texttt{intra\_face\_hair} ($d = +1.81$), \texttt{intra\_face\_face} ($d = +1.66$), \texttt{intra\_face\_build} ($d = +1.51$), and \texttt{intra\_character\_presence} ($d = +1.23$). The largest deficit is \texttt{llm\_object\_accuracy} at $d = -0.68$, with five additional cross-shot object metrics in the range $d \in [-0.68, -0.53]$. The DINOv2 cross-shot metrics show $d \in [-0.66, -0.40]$, but as noted in Table~\ref{tab:cohens_d_p3}'s caption, this disagrees with LLM identity judgment on the same episodes; we discuss this disagreement in Appendix~\ref{app:results:tradeoffs}.

\subsection{Trade-offs and Limitations}
\label{app:results:tradeoffs}

\begin{figure}[t]
    \centering
    \includegraphics[width=\linewidth]{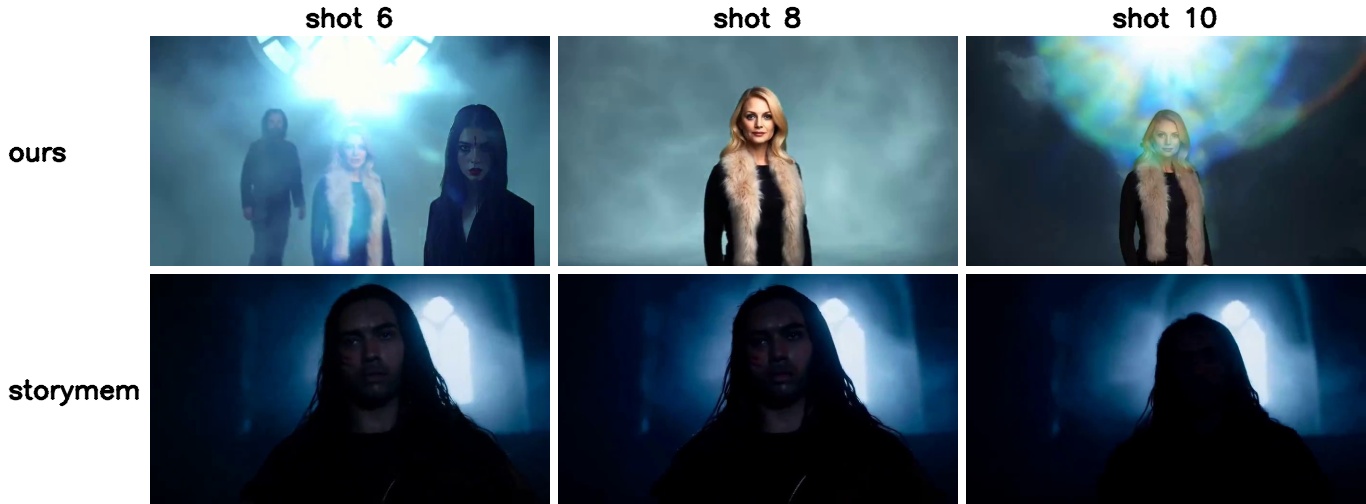}
    \caption{DINOv2 similarity measures consistency in a different way from LLM.}
    \label{fig:trade_off}
\end{figure}

\paragraph{Embedding similarity rewards uniformity, not identity.}
\sysname's largest negative effects concentrate on DINOv2 cross-shot similarity ($d = -0.50$) and on LLM-pairwise object metrics ($d = -0.60$). These metrics measure different things, and their disagreement on character faces is diagnostic of \emph{what each metric actually rewards}. For example, the episode script describes \emph{long-haired blonde girl in a fur-collared coat} in Figure~\ref{fig:trade_off}. \sysname and StoryMem reach near-identical DINOv2 face similarity ($\texttt{cs\_face} = 0.883$ vs.\ $0.8 75$), yet the LLM-pairwise rater identifies zero of StoryMem's face pairs as the same character ($\texttt{llm\_face\_accuracy} = 0.000$) vs.\ 71\% for \sysname ($0.714$).  DINOv2 cs\_face rewards low-fidelity, generic facial renderings that cluster tightly in embedding space without depicting the named character. The LLM, asked an identity-verification question, sees through the mode collapse. Where the two metrics disagree, the LLM is the more honest signal of what a downstream user would notice. This disagreement is not asymptotic. Across all 4 methods, character cs\_face decays as a function of recurrence gap (number of shots between appearances), but the rate of decay differs sharply. 

\paragraph{Per-entity bank limitations on objects.}
Object presence is comparable across methods ($\texttt{intra\_object\_presence} = 0.884$ for \sysname vs.\ $0.906$ for StoryMem), both methods detect roughly the same number of objects, with StoryMem slightly ahead. The cross-shot object gap ($d = -0.60$) is therefore not a denominator artifact: the LLM identifies more genuine inconsistencies in \sysname's object renderings. We attribute this to the ``sticker look'' produced by per-entity object compositing and potential less focus in the pretrain data. Even when objects are provided as conditions, the video generation models tend to fail include or generate it consistently, indicating the challenge in object consistency.

\paragraph{Visual quality is not entity consistency.}
CineTrans wins all four highlighted VBench dimensions (\texttt{subject\_consistency} 0.97, \texttt{imaging\_quality} 68.6) yet has the lowest character presence (0.80) and the worst LLM character accuracy (0.09) of all four methods. A method can render beautifully and still fail to depict the right characters. The two are orthogonal axes. Pixel-quality benchmarks do not predict entity consistency, and \benchname's three-pillar structure is designed precisely to surface this distinction.

\subsection{Long-range Identity Stability: Gap-Decay Analysis}
\label{appendix:results:gap_decay}

\sysname explores whether per-entity memory bank can maintain character identity over long-range recurrence. We test this by binning adjacent-appearance pairs of the same character by their \emph{gap}. The number of intervening shots are between two appearances. We report per-bin mean similarity for each method. 

We compute gap-decay for two cross-shot signals: DINOv2 cosine similarity (face embeddings, the same signal as \texttt{cs\_face}) and LLM identity similarity (per-pair scores from the anchor-vs-each pairwise judge, the same signal as \texttt{llm\_face\_mean\_score}).

We compare \sysname against the two holistic baselines (HoloCine, CineTrans). StoryMem is excluded from this analysis: its strict fidelity gate admits only $\sim$1.3 LLM pairs per episode (180 pairs total across the benchmark, vs.\ \sysname's 502 and HoloCine's 155), and these surviving pairs are systematically the easy cases where character identity is unambiguous. Per-bin estimates from such a heavily-gated, selection-biased subset are unstable. The aggregate \sysname-vs-StoryMem comparison, which folds together coverage and per-pair quality, is reported in Table~\ref{tab:main} (corrected) and Table~\ref{tab:coverage_raw} (decomposed). Per-bin numbers for StoryMem are tabulated in Table~\ref{tab:gap_decay_full} for completeness but should not be interpreted as a directional comparison.

\begin{table}[t]
\centering
\caption{\textbf{LLM face identity similarity by gap distance}, comparing \sysname against the two holistic baselines. }
\label{tab:gap_decay_llm}
\small
\begin{tabular}{lccccc}
\toprule
Method & Gap 1--2 & Gap 3--5 & Gap 6--10 & Gap 11--20 & Gap 21--50 \\
\midrule
\sysname     & \textbf{0.744} (250) & \textbf{0.698} (126) & \textbf{0.646} (76) & \textbf{0.669} (36) & \textbf{0.657} (14) \\
HoloCine     & 0.765 (82)           & 0.517 (36)           & 0.614 (22)          & 0.420 (15)          & ---                 \\
CineTrans    & 0.371 (51)           & 0.408 (25)           & 0.333 (6)           & 0.600 (10)          & 0.457 (7)          \\
\bottomrule
\end{tabular}
\end{table}

On the LLM identity metric (Table~\ref{tab:gap_decay_llm}), \sysname's score declines by only 0.075 from gap 1-2 to gap 11-20, and remains essentially flat (0.66-0.67) thereafter. HoloCine declines by 0.345 over the same range and falls below \sysname at every gap distance beyond 1-2 shots, with the gap widening to $+0.249$ at gap 11-20. The DINOv2 measurement (Table~\ref{tab:gap_decay_dinov2}) shows little gap effect for any method, as DINOv2 cosine similarity reflects identity differently, discussed in Appendix~\ref{app:results:tradeoffs}.

\begin{table}[t]
\centering
\caption{\textbf{DINOv2 face similarity by gap distance.} DINOv2 cosine similarity (mean of adjacent-pair sims to centroid) shows little gap effect across methods, consistent with embedding-similarity rewarding visual self-similarity rather than identity preservation.}
\label{tab:gap_decay_dinov2}
\small
\begin{tabular}{lccccc}
\toprule
Method & Gap 1--2 & Gap 3--5 & Gap 6--10 & Gap 11--20 & Gap 21--50 \\
\midrule
\sysname     & 0.729 (376) & 0.712 (80) & 0.752 (27) & 0.708 (6)  & 0.761 (1) \\
HoloCine     & 0.760 (124) & 0.720 (26) & 0.762 (11) & 0.795 (5)  & ---       \\
CineTrans    & 0.760 (61)  & 0.751 (41) & 0.761 (7)  & 0.773 (2)  & 0.799 (3) \\
\bottomrule
\end{tabular}
\end{table}

For completeness we include StoryMem's per-bin numbers in Table~\ref{tab:gap_decay_full}, with the caveat that these are computed on a heavily-gated subset (StoryMem admits only 22\% of the comparison pairs that \sysname admits) and are not directly comparable to \sysname's broader pool. The \sysname-vs-StoryMem comparison should be read at the aggregate level, where the corrected metric in Table~\ref{tab:main} accounts for coverage.

\begin{table}[t]
\centering
\caption{\textbf{Full gap-decay table including StoryMem,} for completeness. StoryMem's per-bin estimates are based on a small, gate-selected subset and are not directly comparable across methods at the bin level.}
\label{tab:gap_decay_full}
\small
\begin{tabular}{llccccc}
\toprule
Signal & Method & Gap 1-2 & Gap 3-5 & Gap 6-10 & Gap 11-20 & Gap 21-50 \\
\midrule
\multirow{4}{*}{LLM} & \sysname     & 0.744 (250) & 0.698 (126) & 0.646 (76) & 0.669 (36) & 0.657 (14) \\
                     & StoryMem     & 0.830 (105) & 0.759 (51)  & 0.763 (16) & 0.950 (6)  & 0.650 (2)  \\
                     & HoloCine     & 0.765 (82)  & 0.517 (36)  & 0.614 (22) & 0.420 (15) & ---        \\
                     & CineTrans    & 0.371 (51)  & 0.408 (25)  & 0.333 (6)  & 0.600 (10) & 0.457 (7)  \\
\midrule
\multirow{4}{*}{DINOv2} & \sysname  & 0.729 (376) & 0.712 (80)  & 0.752 (27) & 0.708 (6)  & 0.761 (1)  \\
                        & StoryMem  & 0.807 (113) & 0.782 (26)  & 0.733 (6)  & ---        & 0.808 (1)  \\
                        & HoloCine  & 0.760 (124) & 0.720 (26)  & 0.762 (11) & 0.795 (5)  & ---        \\
                        & CineTrans & 0.760 (61)  & 0.751 (41)  & 0.761 (7)  & 0.773 (2)  & 0.799 (3)  \\
\bottomrule
\end{tabular}
\end{table}

\subsection{Per-tier Performance: Long-range Robustness}
\label{appendix:results:per_tier}

\benchname is structured into three difficulty tiers based on episode length: easy (80 episodes, $\le 14$ shots each), medium (40 episodes, $14$--$30$ shots), and hard (20 episodes, 50 shots each). Hard-tier episodes test long-range character recurrence specifically. We use this tier structure to ask whether \sysname's advantage at the aggregate level reflects short-range performance or scales to long sequences.

\begin{table}[t]
\centering
\caption{\textbf{Per-tier breakdown of headline metrics}. \sysname's advantage on character-related and action metrics is robust across all tiers, with several metrics showing the gap \emph{widening} at hard tier (50-shot episodes). DINOv2 cross-shot face shows a flat gap across tiers, consistent with the embedding-vs-identity disagreement (Appendix~\ref{app:results:tradeoffs}).}
\label{tab:per_tier}
\small
\begin{tabular}{llcccc}
\toprule
Metric & Tier & Ours & StoryMem & HoloCine & CineTrans \\
\midrule
\multirow{3}{*}{intra\_character\_presence}
  & Easy   & \textbf{0.961} & 0.884 & 0.896 & 0.818 \\
  & Medium & \textbf{0.974} & 0.836 & 0.845 & 0.788 \\
  & Hard   & \textbf{0.968} & 0.813 & 0.894 & 0.781 \\
\midrule
\multirow{3}{*}{intra\_face\_fidelity}
  & Easy   & \textbf{0.738} & 0.483 & 0.373 & 0.332 \\
  & Medium & \textbf{0.748} & 0.438 & 0.315 & 0.314 \\
  & Hard   & \textbf{0.736} & 0.420 & 0.349 & 0.330 \\
\midrule
\multirow{3}{*}{intra\_action\_overall}
  & Easy   & \textbf{0.618} & 0.590 & 0.588 & 0.262 \\
  & Medium & \textbf{0.626} & 0.563 & 0.585 & 0.256 \\
  & Hard   & \textbf{0.614} & 0.468 & 0.543 & 0.293 \\
\midrule
\multirow{3}{*}{cs\_face (DINOv2)}
  & Easy   & 0.787 & \textbf{0.822} & 0.796 & 0.812 \\
  & Medium & 0.744 & \textbf{0.789} & 0.765 & 0.792 \\
  & Hard   & 0.693 & \textbf{0.746} & 0.713 & 0.735 \\
\midrule
\multirow{3}{*}{llm\_face\_accuracy}
  & Easy   & \textbf{0.344} & 0.197 & 0.188 & 0.058 \\
  & Medium & \textbf{0.393} & 0.220 & 0.223 & 0.042 \\
  & Hard   & \textbf{0.476} & 0.306 & 0.272 & 0.169 \\
\midrule
\multirow{3}{*}{llm\_object\_accuracy}
  & Easy   & 0.101 & \textbf{0.174} & 0.055 & 0.051 \\
  & Medium & 0.169 & \textbf{0.210} & 0.074 & 0.119 \\
  & Hard   & \textbf{0.244} & 0.267 & 0.144 & 0.123 \\
\bottomrule
\end{tabular}
\end{table}

\paragraph{Character-related metrics: robustness improves with sequence length.}
Across the three character-centric metrics in Table~\ref{tab:per_tier}, the head-to-head gap between \sysname and StoryMem (the strongest baseline) is stable or grows as episodes get longer.

\begin{table}[t]
\centering
\caption{\textbf{\sysname-vs-StoryMem head-to-head gap by tier} on selected metrics. Positive values favor \sysname.}
\label{tab:per_tier_gap}
\small
\begin{tabular}{lrrrl}
\toprule
Metric & Easy & Medium & Hard & Direction \\
\midrule
intra\_character\_presence  & $+0.077$ & $+0.138$ & $+0.155$ & grows with tier \\
intra\_face\_fidelity       & $+0.255$ & $+0.310$ & $+0.316$ & grows slightly \\
intra\_action\_overall      & $+0.028$ & $+0.063$ & $+0.146$ & grows substantially \\
llm\_face\_accuracy         & $+0.147$ & $+0.173$ & $+0.170$ & grows then plateaus \\
\midrule
llm\_object\_accuracy       & $-0.073$ & $-0.041$ & $-0.022$ & deficit shrinks \\
cs\_face (DINOv2)           & $-0.035$ & $-0.045$ & $-0.053$ & flat \\
\bottomrule
\end{tabular}
\end{table}

The pattern is sharpest on \texttt{intra\_action\_overall}: at easy tier the gap is $+0.028$ (essentially tied), but at hard tier it grows to $+0.146$ ($5\times$ larger). Inspection of the underlying numbers shows that \sysname's score is roughly flat across tiers ($0.618 \rightarrow 0.626 \rightarrow 0.614$) while StoryMem drops by 21\% from easy to hard ($0.590 \rightarrow 0.468$). On 50-shot episodes, where actions span longer, more diverse sequences with more potential for character drift, \sysname's per-entity bank actively prevents the action-correctness collapse that other methods suffer.

A similar but smaller-magnitude pattern holds on \texttt{intra\_character\_presence}: the gap grows from $+0.077$ to $+0.155$, with \sysname remaining essentially flat ($0.96$--$0.97$) while StoryMem drops from $0.88$ to $0.81$. On longer episodes, baselines start failing to render scheduled characters at all in some shots; \sysname continues to render them.

\paragraph{LLM identity advantage compounds with longer sequences.}
On \texttt{llm\_face\_accuracy}, \sysname's score itself \emph{grows} with tier difficulty: $0.344 \rightarrow 0.393 \rightarrow 0.476$ (a 38\% improvement from easy to hard). StoryMem also improves ($0.197 \rightarrow 0.306$), but \sysname's growth is larger in absolute terms. We interpret this as a measurement effect: hard-tier (50-shot) episodes provide more pairs for the LLM to judge, and \sysname's higher gate pass-rate (Appendix~\ref{appendix:results:coverage}) translates into a larger admitted pool where the LLM can identify successfully-preserved character identity. 

\paragraph{Object trade-off shrinks at hard tier.}
The aggregate \texttt{llm\_object\_accuracy} loss to StoryMem ($d = -0.68$, Appendix~\ref{appendix:results:cohens_d_full}) is concentrated at easy tier ($-0.073$) and shrinks substantially at hard tier ($-0.022$, essentially tied). \sysname's object accuracy itself grows with tier ($0.101 \rightarrow 0.169 \rightarrow 0.244$), suggesting that directly applying object entity bank is challenging to improve object consistency, but may help more when backbone's performance weakens.

\paragraph{DINOv2 deficit stays flat across tiers.}
In contrast, the \texttt{cs\_face} gap to StoryMem is uniformly $-0.04$ to $-0.05$ across all three tiers, with \emph{no} sequence-length effect. The flat DINOv2 gap, combined with a tier-dependent LLM gap on the same characters, supports the §4.3 / Appendix~\ref{app:results:tradeoffs} interpretation that DINOv2 cross-shot measures generic visual similarity rather than character identity. This is the gap-decay observation (Appendix~\ref{appendix:results:gap_decay}) reproduced at the tier level: same finding, different aggregation.

The hard tier contains only 20 episode, so per-tier estimates have wider confidence intervals than the aggregate. The directional claims above (gap-grows-with-tier, deficit-shrinks-at-hard) are robust to this in the sense that the differences exceed plausible standard errors at $n=20$, but precise hard-tier values should be treated as approximate. The tier breakdown is intended primarily as a sanity check on the aggregate story, not as the basis for new claims.

\section{Additional Related Work}
\label{sec:app_related_work}

\paragraph{Identity-Preserving Video Generation.}
Maintaining consistent character appearance within generated video has been approached through frequency-domain identity decomposition~\citep{yuan2025identity}, multi-subject reference conditioning~\citep{liu2025phantom,jiang2025vace}, reinforcement learning with identity-aware rewards~\citep{meng2025identity}, and training-free cross-shot feature sharing~\citep{singh2025storybooth}.
Other works explore zero-shot identity animation~\citep{he2024id} and universal identity-preserving synthesis~\citep{zhong2025concat}.
However, these methods focus primarily on human facial identity for one or two subjects, leaving broader entity types, such as objects, locations, and character ensembles, largely unaddressed.
 
\paragraph{LLM-Directed Video Generation.}
LLMs have been used as video planners to produce scene descriptions with entity layouts and consistency groupings~\citep{lin2023videodirectorgpt}, decompose prompts into structured shot instructions~\citep{chen2025skyreels}, and combine coarse scene planning with fine-grained object-level layout control~\citep{wang2026dreamrunner}.
Multi-agent frameworks further coordinate specialized modules for long video planning~\citep{xie2024dreamfactory,huang2025filmaster}.
These methods demonstrate the value of LLM-guided planning for structural coherence but treat entity consistency as a byproduct of shared embeddings or layout constraints.
Our multi-agent system differs by maintaining persistent per-entity memory banks for both visual and textual information and injecting the retrieved entity memory as context for consistent cross-shot video generation.

\section{Broader Impact}
\label{sec:broader_impact}

\benchname and \sysname operate in cross-shot long video generation with both clear creative-tool applications and well-documented misuse risks. We discuss both, along with limitations of the proposed evaluation framework.

\paragraph{Beneficial applications.}
Reliable character consistency in multi-shot generation lowers the barrier for creators (independent animators, educators, accessibility advocates) to produce longer-form visual narratives without large production teams. Narrative video that preserves character identity across shots is a precondition for accessible storytelling, educational content with recurring characters, and rapid prototyping in animation and storyboarding.

\paragraph{Misuse risks.}
The same capabilities enable the generation of synthetic videos depicting real people in fabricated scenarios, with applications including non-consensual deepfakes, defamation, and political disinformation. \sysname is built on pretrained text-to-video backbone, LLM, text-to-image generation model and inherits any safety properties of that backbone.

\end{document}